\documentclass[twoside]{article}
\usepackage{ecj,palatino,,latexsym,natbib} 
   \usepackage[pdftex]{graphicx}
\graphicspath{{figures/}}
\DeclareGraphicsExtensions{.pdf,.jpeg,.png}

\usepackage{amsmath}

\usepackage{textcomp}
\newcommand{\textapprox}{\raisebox{0.5ex}{\texttildelow}}

\usepackage[ruled,linesnumbered]{algorithm2e}
\makeatletter
\newcommand{\removelatexerror}{\let\@latex@error\@gobble}
\makeatother

  \usepackage[caption=false,font=footnotesize]{subfig}

\usepackage{tabularx}
\usepackage{booktabs}
\usepackage{framed}
\usepackage{wrapfig}
\usepackage{xcolor}
 \usepackage[tight-spacing=true]{siunitx} 
\usepackage{setspace}

\newlength\hMp

\newcommand{\revisionOne}[1]{#1}
\newcommand{\revisionOnePar}[1]{#1}
\newcommand{\revisionTwo}[1]{#1}

\newcommand{\revisionThree}[1]{#1}
\newcommand{\revisionThreePar}[1]{#1}
\newcommand{\baselines}[2]{
		\setlength{\tabcolsep}{.6em}
	\subfloat[#1\vspace{-.25em}]{%
		\centering
		\footnotesize
		\begin{tabularx}{.485\textwidth}{@{}XS[table-format=2.1,round-precision=1]S[table-format=2.1,round-precision=1]S[table-format=1.3,round-precision=3]S[table-format=1.3,round-precision=3]S[table-format=1.3,round-precision=3]@{}}
			\toprule
			Method & \text{K} & \text{Conn} & \text{Spar} & \text{Sep} & \revisionOne{ARI} \\
			\midrule
			#2
			\bottomrule
		\end{tabularx}
	}
	\vspace{-.5em}
}

\newcommand{\multitree}[2]{
	\setlength{\tabcolsep}{.4em}
	\subfloat[#1\vspace{-.25em}]{%
		\centering
		\footnotesize
		\begin{tabularx}{.485\textwidth}{@{}XS[table-format=2.2,round-precision=2]S[table-format=2.1,round-precision=1]S[table-format=2.1,round-precision=1]S[table-format=1.3,round-precision=3]S[table-format=1.3,round-precision=3]S[table-format=1.3,round-precision=3]@{}}
			\toprule
			Method & \text{Fitness} & \text{K} & \text{Conn.} & \text{Spar.} & \text{Sep.} & \revisionOne{ARI} \\
			\midrule
			#2
			\bottomrule
		\end{tabularx}
	}
	\vspace{-.5em}
}

\newcommand{\mtGraph}[2]{
	\subfloat[#2]{
		\includegraphics[width=.32\textwidth]{#2-mtR}
	}
}

\newcount\tmpnum
\def\tallymarks#1{\leavevmode \lower1bp\vbox to9bp{}%
	\tmpnum=#1
	\loop \ifnum\tmpnum<5 \kern1bp \tallynum\tmpnum \else \tallyV \fi
	\advance\tmpnum by-5
	\ifnum\tmpnum>0 \repeat
}
\def\tallynum#1{\bgroup\tmpnum=#1\relax
	\loop \ifnum\tmpnum>0
	\kern1bp \tallyI \kern1bp
	\advance\tmpnum by-1
	\repeat
	\egroup
}
\def\tallyI{\pdfliteral{q .5 w 0 -1 m 0 8 l S Q}}
\def\tallyV{\kern1bp\pdfliteral{q .5 w -1 0 m 9 7 l S Q}\tallynum4\kern1bp }

\begin{document}

\ecjHeader{x}{x}{xxx-xxx}{201X}{GP for Evolving Cluster Similarity Functions}{A. Lensen, B. Xue, and M. Zhang}
\title{\bf Genetic Programming for Evolving\\ Similarity Functions for Clustering: Representations and Analysis}  

\author{\name{\bf Andrew Lensen} \hfill \addr{Andrew.Lensen@ecs.vuw.ac.nz}\\ 
        \addr{Evolutionary Computation Research Group, Victoria University of Wellington, Wellington 6140, New Zealand}
\AND
       \name{\bf Bing Xue} \hfill \addr{Bing.Xue@ecs.vuw.ac.nz}\\
       \addr{Evolutionary Computation Research Group, Victoria University of Wellington, Wellington 6140, New Zealand}
\AND
		\name{\bf Mengjie Zhang} \hfill \addr{Mengjie.Zhang@ecs.vuw.ac.nz}\\
		\addr{Evolutionary Computation Research Group, Victoria University of Wellington, Wellington 6140, New Zealand}\\
}

\maketitle

\begin{abstract}
	Clustering is a difficult and widely-studied data mining task, with many varieties of clustering algorithms proposed in the literature. Nearly all algorithms use a similarity measure such as a distance metric (e.g.\ Euclidean distance) to decide which instances to assign to the same cluster. These similarity measures are generally pre-defined and cannot be easily tailored to the properties of a particular dataset, which leads to limitations in the quality and the interpretability of the clusters produced. In this paper, we propose a new approach to automatically evolving similarity functions for \revisionTwo{a given clustering algorithm by} using genetic programming. We introduce a new genetic programming-based method which automatically selects a small subset of features (feature selection) and then combines them using a variety of functions (feature construction) to produce dynamic and flexible similarity functions that are specifically designed for a given dataset. We demonstrate how the evolved similarity functions can be used to perform clustering using a graph-based representation. The results of a variety of experiments across a range of large, high-dimensional datasets show that the proposed approach can achieve higher and more consistent performance than the benchmark methods. We further extend the proposed approach to automatically produce multiple complementary similarity functions by using a multi-tree approach, which gives further performance improvements. We also analyse the interpretability and structure of the automatically evolved similarity functions to provide insight into how and why they are superior to standard distance metrics.
\end{abstract}


\section{Introduction}
Clustering  is a fundamental data mining task \citep{fayyad1996data}, which aims to group related/similar instances into a number of clusters where the data is unlabelled. It is one of the key tasks in exploratory data analysis, as it enables data scientists to reveal the underlying structure of unfamiliar data, which can then be used for further analysis \citep{jain2010data}.

Nearly all clustering algorithms utilise a similarity measure, usually a distance function, to perform clustering as close instances are similar to each other, and expected to be in the same cluster. The most common distance functions, such as Manhattan or Euclidean distance, are quite inflexible: they consider all features equally despite features often varying significantly in their usefulness. Consider a weather dataset of daily records which contains the two following features: \textit{day of week} and \textit{rainfall (mm)}. Clusters generated using the rainfall feature will give insight into what days are likely to be rainy, which may allow better prediction of whether we should take an umbrella in the future. Clusters generated with the day of week feature however are likely to give little insight or be misleading --- intuitively, we know that the day of the week has no effect on long-term weather patterns and so any clusters produced could mislead us. Ideally, we would like to perform \textit{feature selection} to select only the most useful features in a dataset. These distance functions also have uniform behaviour across a whole dataset, which makes them struggle with common problems such as clusters of varying density or separation, and noisy data. Indeed, trialling a range of similarity measures is commonly a tedious but necessary parameter tuning step when performing cluster analysis. \textit{Feature construction} is the technique of automatically combining existing low-level features into more powerful high-level features. Feature construction could produce similarity measures which are better fitted to a given dataset, by combining features in a non-uniform and flexible manner. These \textit{feature reduction} techniques have been shown to be widely effective in both supervised and unsupervised domains \citep{xue2015survey,aggarwal2014data}.

A variety of representations have been proposed for modelling clustering solutions. The graph representation models the data in an intuitive way, where instances (represented as nodes) are connected by an edge if they are \textit{similar enough} \citep{luxburg2007spectral}. This is a powerful representation that allows modelling a variety of cluster shapes, sizes, and densities, unlike the more common prototype-based representations such as $k$-means. However, algorithms using graph representations are very dependent on the criterion used to select edges \citep{luxburg2007spectral}. One of the most common criteria is to simply use a fixed threshold \citep{luxburg2007spectral}, which indicates the distance at which two instances are considered too dissimilar to share an edge. Such a threshold must be determined independently for every dataset, and this approach typically does not allow varying thresholds to be used in different clusters. Another popular criterion is to put an edge between each instance and its $N$-nearest neighbours \citep{luxburg2007spectral}, where N is a small integer value such as 2, 3, or 4. $N$ must also be determined before running the algorithm, with results being very sensitive to the $N$ value chosen. Again, this method does not allow for varied cluster structure.

Many of the above issues can be tackled by using a similarity function which is able to be \textit{automatically} tailored to a specific dataset, and which can treat different clusters within a dataset differently. Genetic programming (GP) \citep{koza1992genetic} is an evolutionary computation (EC) \citep{eiben2015introduction} method that automatically evolves \textit{programs}. The most common form of GP is \textit{tree-based} GP, which models solutions in the form of a tree which takes input (such as a feature set) and produces an output, based on the functions performed within the tree. We hypothesise that this approach can be used to automatically evolve similarity functions that are represented as GP trees, where a tree takes two instances as input and produces an output that corresponds to how similar those two instances are. By automatically combining only the most relevant features (i.e.\ performing feature selection and construction), more powerful and specific similarity functions can be generated to improve clustering performance on a range of datasets. GP can also use multiple trees to represent each individual/solution. Multi-tree GP has the  potential to automatically generate multiple complementary similarity functions, which are able to \textit{specialise} on different clusters in the dataset. To our best knowledge, such an approach has not been investigated to date.

\subsection{Goals}
This work aims to propose the first approach to using GP for automatically evolving similarity functions with a graph representation for clustering (GPGC). This work is expected to improve clustering performance while also producing more interpretable clusters which use only a small subset of the full feature set. We will investigate:
\begin{itemize}
\itemsep 1pt
	\item how the output of an evolved similarity function can be used to create edges in clustering with a graph representation;
	\item what fitness function should be used to give high-quality clustering results;
	\item whether using multiple similarity functions to make a consensus decision can further improve clustering results; and 
	\item whether the evolved similarity functions are more interpretable and produce simpler clusters than standard distance functions.
\end{itemize}

A piece of preliminary work was presented in our previous research \citep{lensen2017gpgc}, which proposed evolving a single similarity function with a single-tree approach for clustering. This work extends the preliminary work significantly  by providing more detailed and systematic description and justification and introducing a multi-tree approach, as well as much more rigorous comparisons to existing techniques and more detailed analysis of the proposed method.

\section{Background}
\subsection{Clustering}
A huge variety of approaches have been proposed for performing clustering \citep{xu2005survey,aggarwal2014data}, which can be generally categorised into hard, soft (fuzzy), or hierarchical clustering methods. In hard and soft clustering, each instance belongs to exactly one or to at least one cluster respectively. In contrast, hierarchical clustering methods build a hierarchy of clusters, where a parent cluster contains the union of its child clusters. The majority of work has focused on hard clustering, as partitions where each instance is in exactly one cluster tend to be easier to interpret and analyse. A number of distinct models have been proposed for performing hard clustering: prototype-based models (including the most famous clustering method $k$-means \citep{hartigan1979}, and its successor $k$-means++ \citep{arthur2007kmeansplusplus}), density-based models (e.g.\ DBSCAN \citep{ester1996density} and OPTICS \citep{ankerst1999optics}), graph-based models (e.g.\  the Highly Connected Subgraph (HCS) algorithm \citep{shamir2000clustering}), and statistical approaches such as distribution-based (e.g.\ EM clustering) and kernel-based models. Prototype-based models produce a number of prototypes, each of which corresponds to a unique cluster, and then assigns each instance to its nearest prototype using a distance function, such as Euclidean distance. While these models are the most popular, they are inherently limited by their use of prototypes to define clusters: when there are naturally clusters that are non-hyper-spherically shaped, prototype-based models will tend to perform poorly as minimising the distance of instances to the prototype encourages spherical clusters. This problem is further exemplified when a cluster is non-convex. 

Graph-based clustering algorithms \citep{luxburg2007spectral} represent clusters as distinct graphs, where there is a path between every pair of instances in a cluster graph. This representation means that graph-based measures are not restricted to clusters with hyper-spherical or convex shapes. The HCS algorithm \citep{shamir2000clustering} uses a similarity graph which connects instances sharing a similarity value (e.g.\ distance) above a certain threshold, and then iteratively splits graphs which are not \textit{highly connected} by finding the minimum cut, until all graphs are highly connected. Choosing a good threshold value in HCS can be difficult when there is no prior knowledge of the data.

EC techniques have also been applied to clustering successfully \citep{lorena2001constructive,picarougne2007new,nanda2014survey,garcia2016automatic,sheng2016adaptive} with many genetic algorithms (GA) and particle swarm optimisation (PSO) techniques used to automatically evolve clusters. Again, the majority of the literature tends to use prototype-based models, and little work uses feature reduction techniques to improve the performance of clustering methods and to produce more interpretable clusters. There is notably a deficit of methods using GP for clustering, and no current methods, asides from our preliminary work \citep{lensen2017gpgc}, that use GP to automatically evolve similarity functions. Relevant EC clustering methods will be discussed further in the related work section.

\subsection{Feature Reduction}
\textit{Feature reduction} is a common strategy used to improve the performance of data mining algorithms and interpretability of the models or solutions produced \citep{liu2012feature}. The most common feature reduction strategy is \textit{feature selection}, where a subset of the original feature set is selected for use in the data mining algorithm. Using fewer features can decrease training time, produce more concise and understandable results, and even improve performance by removing irrelevant/misleading features or reducing over-fitting (in supervised learning). Feature selection has been extensively studied, on a range of problems, such as classification \citep{tang2014feature} and clustering \citep{alelyani2013feature}. \textit{Feature construction}, another feature reduction strategy, focuses on automatically producing new high-level features, which combine multiple features from the original feature set in order to produce more powerful constructed features (CFs). As with feature selection, the use of feature construction can improve performance and interpretability by automatically combining useful features.

Research into the use of EC techniques for performing feature reduction has become much more popular during the last decade, due to the ability of EC techniques to efficiently search a large feature set space. Feature selection has been widely studied using PSO and GAs \citep{pedrajas2014scalable,xue2015survey}, and GP has emerged as a powerful feature construction technique due to its tree representation allowing features to be combined in a hierarchical manner using a variety of functions \citep{espejo2010survey,neshatian2012filter}. Despite this, the use of EC for feature reduction in clustering tasks has thus far been relatively unexplored. Given that clustering is an unsupervised learning task with a huge search space, especially when there are many instances, features, or clusters, good feature reduction methods for clustering are needed.

\revisionOnePar{
\subsection{Subspace Clustering}
Another approach for performing feature reduction in clustering tasks is \textit{subspace clustering} \citep{liu2005toward,muller2009evaluating}, where each cluster is located in a subspace of the data, i.e. it uses only a subset of the features. In this regard, each cluster is able to correspond to a specific set of features that are used to characterise that cluster, which has the potential to produce better-fitted and more interpretable clusters. Several EC methods have been proposed for performing subspace clustering \citep{vahdat2014on,peignier2015subspace}. However, subspace clustering intrinsically has an even larger search space than normal clustering, as the quantity and choice of features must be made for every cluster, rather than only once for the dataset \citep{parsons2004subspace}. In this paper, we do not strictly perform subspace clustering, but rather we allow the proposed approach to use different features in different combinations across the cluster space. 
}
\subsection{Related Work}

There has been a handful of work proposed that uses a graph-based representation in conjunction with EC for performing clustering. One notable example is the MOCK algorithm \citep{handl2007evolutionary}, which uses a GA with a locus-based graph approach to perform multi-objective clustering. Another GA method has also been proposed, which take inspiration from spectral clustering and uses either a label-based or medoid-based encoding to cluster the similarity graph \citep{menendez2014genetic}.


The use of GP for performing clustering is very sparse in the literature, with only about half a dozen pieces of work proposed.
One early work uses a grammar-based approach \citep{falco2005novel} where a grammar is evolved to assign instances to clusters based on how well they matched the evolved \textit{formulae}. Instances that do not match any formulae are assigned to the closest centroid. This assignment technique, and the fitness function used, means that the proposed method is biased towards hyper-spherical clustering.
Boric et al. \citep{boric2007genetic} proposed a multi-tree representation, where each tree in an individual corresponds to a single cluster. This method required the number of trees ($t$) to be set in advance, i.e.\ the number of clusters must be known \textit{a priori}, which may not be available in many cases. 
A single-tree approach has also been proposed \citep{ahn2011genetic}, which uses integer rounding to assign each instance to a cluster based on the output of the evolved tree. Such an approach is unlikely to work well on datasets with a relatively high $K$, and this method produces a difficult search space due to clusters having an implicit order.
Another proposed approach \citep{coelho2011multi} uses GP to automatically build consensus functions that combine the output of a range of clustering algorithms to produce a fusion partition. This is suggested to combine the benefits of each of the clustering algorithms, while avoiding their limitations. Each of the clustering algorithms use a fixed distance function to measure similarity between instances, and several of the algorithms require that $K$ is known in advance.
More recently, a GP approach has been proposed based on the idea of novelty search \citep{naredo2013searching}, where in lieu of an explicit fitness function, the \textit{uniqueness} (novelty) of a solution in the behavioural landscape is used to determine whether it is used in subsequent generations. This approach was only tested on problems with two clusters, and it is unclear how it would scale as $K$ increases, given that the behavioural landscape would become exponentially larger.

While there has been very little work utilising feature construction techniques for improving the performance of clustering, there has been a significant amount of study into using feature selection for clustering problems \citep{dy2004feature,alelyani2013feature}, with dimensionality reduction approaches such as Principal Component Analysis (PCA) \citep{jolliffe2011pca} being used with both EC \citep{kuo2012integration} and non-EC clustering methods. In addition, there has been some work using EC to perform simultaneous clustering and feature selection, with the aim of concurrently tailoring the features selected to the clusters produced. PSO, in particular, has been shown to be effective on this task \citep{sheng2008niching,lensen2017using}.

The clustering literature has an overwhelming focus on producing novel clustering algorithms which employ a wide range of techniques for modelling and searching the clustering problem space. However, there has been very little focus on new techniques for automatically creating more appropriate and more powerful similarity measures to accurately model the relationships between instances on a specific dataset. GP, with its intrinsic function-like solution structure, is a natural candidate for automatically evolving similarity functions tailored to the data it is trained on. GP, and EC methods in general, have been shown to be effective on large dataset sizes and dimensionality; GP has the potential to evolve smaller, more refined, and more interpretable similarity functions on very big datasets. This paper investigates the capability of GP for automatically constructing power similarity functions.

\section{Proposed Approaches}
\begin{figure}[!tp]
	\vspace{-1em}
	\centering
	{%
		\setlength{\fboxsep}{0pt}%
		\fbox{
			\includegraphics[width=0.7\linewidth]{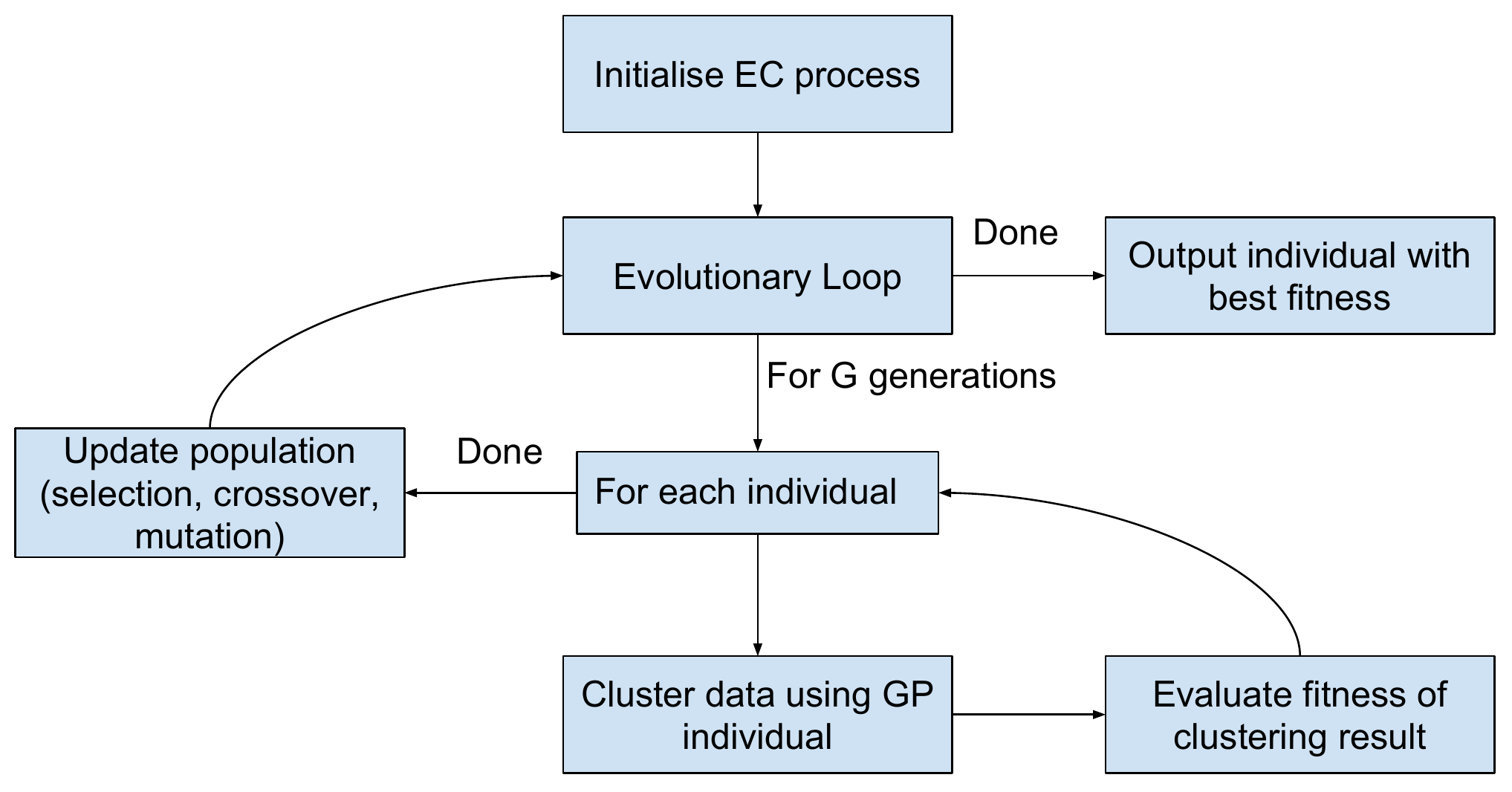}
		}
	}%
	
	\caption{The overall flow of the proposed GPGC algorithm. The clustering process is discussed in detail in Section \ref{clusteringAlgorithm}, and is shown in Algorithm \ref{proposedAlgorithm}.}
	\label{fig:gpgcjournalflow}
	\vspace{-1em}
\end{figure}

An overview of the proposed GPGC algorithm is shown in Fig.\ \ref{fig:gpgcjournalflow}. We discuss the different parts of this overall algorithm in the following subsections.

\subsection{GP Program Design}
\label{programDesign}
\begin{wrapfigure}{R}{0.4\textwidth}
	\vspace{-3.5em}
	\centering
	\includegraphics[height=10em]{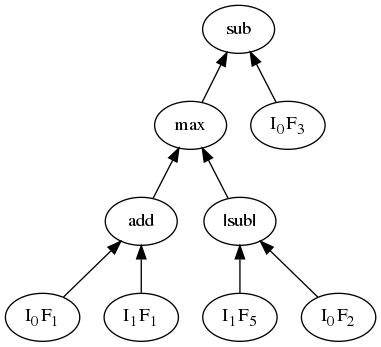}
	\centering
	\vspace{-.5em}
	\caption{An example of of a similarity function with the expression $sub\big(max(add(I_0F_1,I_1F_1),$ $|sub|(I_1F_5,I_0F_2)), I_0F_3\big)$.}
	\label{fig:exampleststructure}
	\vspace{-1em}
\end{wrapfigure}

To represent a similarity function, a GP tree must take two instances as input and produce a single floating-point output corresponding to the similarity of the two instances. \revisionOne{Therefore, we define the terminal set as all feature values of both instances}, such that there are $2m$ possible terminals for $m$ features ($I_0F_0$ and $I_1F_0$ through to $I_0F_{m-1}$ and $I_1F_{m-1}$), as well as a random floating-point value (for scaling purposes). The function set comprises of the normal arithmetic functions ($+$, $-$,$\times$,$\div$), two absolute arithmetic functions ($|+|$ and $|-|$), and the $max$, $min$ and $if$ operators. All of these functions asides from $if$ take two inputs and output a single value which is the result of applying that function. The $if$ function takes three inputs and outputs the second input if the first is positive, or the third input if it is not. We include the $if$, $max$, and $min$ functions to allow conditional behaviour within a program, in order to allow the creation of similarity functions which operate differently across the feature space. The $\div$ operator is protected division: if the divisor (the second input) is zero, the operator will return a value of one. An example of a similarity function using this GP program design is shown in Fig.\ \ref{fig:exampleststructure}.

\subsection{Clustering Process}
\label{clusteringAlgorithm}
As we are using a graph representation, every pair of instances which are deemed “close enough” by an evolved GP tree should be connected by an edge. As discussed before, we would like to refrain from using a fixed similarity threshold as varying thresholds may be required across a dataset due to varying cluster density. We therefore use the approach where each instance is connected to a number of its most similar neighbours (according to the evolved similarity function). 


To find the most similar neighbour of a given instance for an evolved similarity function requires comparing the instance to every other instance in the dataset. Normally, when using a distance metric, these pairwise similarities could be precomputed; however, in the proposed algorithm, these must be computed separately for every evolved similarity function, giving $O(n^2)$ comparisons for every GP individual on every generation of the training process, given $n$ instances. In order to reduce the computational cost, we use a heuristic whereby each instance is only compared to its $l$ nearest neighbours based on Euclidean distance. The set of nearest neighbours can be computed at the start of the EC training process, meaning only $O(nl)$ comparisons are required per GP individual. By using this approach, we balance the flexibility of allowing an instance to be connected to many different neighbours with the efficiency of using a subset of neighbours to compare to. As we use Euclidean distance only to give us the \textit{order} of neighbours, the problems associated with Euclidean distance at high dimensions should not occur. We found in practice that setting $l$ as $l = \lceil\sqrt[3]{n}\rceil$ gave a good neighbourhood size that is proportional to $n$, while ensuring $l$ is at least 2 \citep{lensen2017using}.

Algorithm \ref{proposedAlgorithm} shows the steps used to produce a cluster for a given GP individual, $X$. For each instance $I$ in the dataset, the nearest $l$ neighbours are found using the pre-computed Euclidean distance mappings. Each of these $l$ neighbours is then fed into the bottom of the tree ($X$) along with $I$. The tree is then evaluated, and produces an output corresponding to the similarity between $I$ and that neighbour. The neighbour with the highest similarity is chosen, and an edge is added between it and $I$. \revisionOne{As in \citep{lensen2017gpgc}, we tested adding edges to more than one nearest neighbour, but found that performance tended to drop.} Once this process has been completed for each $I \in Dataset$, the set of edges formed will give a set of graphs, where each graph represents a single cluster. These graphs can then be converted to a set of clusters by assigning all instances in each graph to the same cluster.


\begin{figure}[!t]
			\vspace{-1em}
\small 
	\removelatexerror
	\begin{algorithm}[H]

		\caption{Process to produce a cluster using a given GP individual ($X$) and the number of neighbours ($l$).}
		\label{proposedAlgorithm}
		
		
		
		
		$Edges = \{\}$\;
		\For(\emph{Choose edge}){$I \in Dataset$}{
			$Neighbours = nearestNeighbours(I,l)$\;
			$Neighbour_{Best} = \emptyset$\;
			$Similarity_{Best} = -\infty$\;
			\For(\emph{Test neighbour}){$Y \in Neighbours$}{
				$similarity = evaluate(X, I, Y)$\;
				\If{$similarity > Similarity_{Best}$}{$Neighbour_{Best}= Y$\;}
				
			}
			add edge from $I$ to $Neighbour_{Best}$ to $Edges$\;
		}
		$Cluster = graphToCluster(Edges)$\;
	\end{algorithm}
	\vspace{-1em}
\end{figure}

\subsection{Fitness Function}
\label{fitnessFunction}
The most common measures of cluster quality are cluster compactness and separability \citep{aggarwal2014data}. A good cluster partition should have distinct clusters which are very dense in terms of the instances they contain, and which are far away from other clusters. A third, somewhat less common measure, is the instance connectedness, which measures how well a given instance lies in the same cluster as its nearby neighbours \revisionTwo{\citep{handl2007evolutionary}}. The majority of the clustering literature measures performance in a way that implicitly encourages hyper-spherical clusters to be produced, by minimising each instance's distance to its cluster mean, and maximising the distance between different cluster means. Such an approach is problematic, as it introduces bias in the shape of clusters produced, meaning elliptical or other non-spherical clusters are unlikely to be found correctly.

As a graph representation is capable of modelling a variety of cluster shapes, we instead propose using a fitness function which balances these three measures of cluster quality in a way that gives minimal bias to the shape of clusters produced. We discuss each of these in turn below:

\paragraph{Compactness} To measure the \textit{compactness} of a cluster, we choose the instance in the cluster which is the furthest away from its nearest neighbour in the same cluster; that is, the instance which is the most isolated within the cluster. The distance between that instance and its nearest neighbour, called the \textit{sparsity} of the cluster, should be minimised. We define sparsity in Equation (\ref{eqn:sparsity}), where $C_i$ represents the $i^{th}$ cluster of $K$ clusters, $I_{a} \in C_{i}$ represents an instance in the $i^{th}$ cluster, and $d(I_a,I_b)$ is the Euclidean distance between two instances.
\begin{equation}
\small \text{Sparsity} = \max_{I_a \in C_i} \big\{\min_{I_b \in C_i} d(I_a,I_b) \big\vert I_a \neq I_b \big\}
\label{eqn:sparsity}
\end{equation}

\paragraph{Separability} To measure the separation of a cluster, we find the minimum distance from that cluster to any other cluster. This is equivalent to finding the minimum distance between the instances in the cluster and all other instances in the dataset which are not in the same cluster, as shown in Equation (\ref{separationEquation}). The separation of a cluster should be maximised to ensure that it is distinct from other clusters.
\begin{equation}
\small \text{Separation} = \min_{I_a \in C_i} \big\{\min_{I_b \notin C_i} d(I_a,I_b)\big\}
\label{separationEquation}
\end{equation}

\paragraph{Connectedness} An instance's \textit{connectedness} is measured by finding how many of its $c$ nearest neighbours are assigned to the same cluster as it, with higher weighting given to neighbours which are closer to the given instance, as shown in Equation (\ref{eqn:connectedness}). To prevent connectedness from encouraging spherical clusters, $c$ must be chosen to be adequately small --- otherwise, large cluster “blobs” will form. We found that setting $c=10$ provided a good balance between producing connected instances and allowing varying cluster shapes. The mean connectedness of a dataset should be maximised.
\begin{equation}
\small \text{Connectedness} = \frac{1}{K} \sum_{i=1}^{K} \frac{1}{|C_{i}|} \sum_{I_a\in C_{i},\ I_b \in N_{I_a}} d_{inv}(I_a,I_b) \big\vert I_b \cap C_{i}
\label{eqn:connectedness}
\end{equation}
where $N_{I_a}$ gives the $c$ nearest neighbours of $I_{a}$, \revisionTwo{$I_b \cap C_{i}$ indicates that $I_a$ and $I_b$ are (correctly) in the same cluster,} and
\begin{equation}
\small d_{inv}(I_a,I_b) = min \big[\frac{1}{d(I_a,I_b)},10\big]
\end{equation}
The inverse distance between two instances is capped at 10, to prevent very close instances from overly affecting the fitness measure. \revisionTwo{Inverse distance is used to weight closer neighbours more highly.}

Our proposed fitness function is a combination of these three measures (Equations (\ref{eqn:sparsity})--(\ref{eqn:connectedness})): we find each cluster's ratio of sparsity: separation (as they are competing objectives) as shown in Equation (\ref{eqn:spasep}), and then measure the partition's fitness by also considering the connectedness, as shown in Equation (\ref{fitnessFunctionEquation}). This fitness function should be maximised.

\begin{minipage}{.48\linewidth}
\begin{equation}
\small \text{Mean SpaSep} = \frac{1}{K} \sum_{i=1}^{K} \frac{\text{Sparsity}}{\text{Separation}}
\label{eqn:spasep}
\end{equation}	
\end{minipage}
\begin{minipage}{.48\linewidth}
\begin{equation}
\small \text{Fitness} = \frac{\text{Connectedness}}{\text{Mean SpaSep}}
\label{fitnessFunctionEquation}
\end{equation}	
\end{minipage}

\subsection{Using a Multi-Tree Approach}
\label{multiTree}
As previously discussed, using a single fixed similarity function means that every pair of instances across a dataset must be compared identically, i.e.\ with all features weighted equally regardless of the characteristics of the given instances. By using GP to automatically evolve similarity functions containing conditional nodes ($if$, $max$, and $min$), we are able to produce trees which will measure similarity dynamically. However, a tree is still limited in its flexibility, as there is an inherent trade-off between the number of conditional nodes used and the complexness of the constructed features in a tree --- more conditionals will tend to mean simpler constructed features with fewer operators (and vice versa), due to the limitations on tree depth and training time.

\begin{figure}[t!]
	\centering
	\subfloat[]{\includegraphics[height=7.5em]{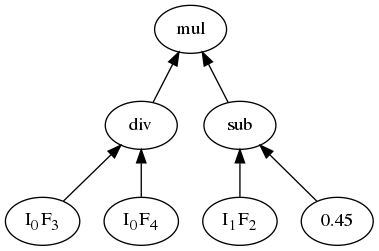}}\hfill
	\subfloat[]{\includegraphics[height=7.5em]{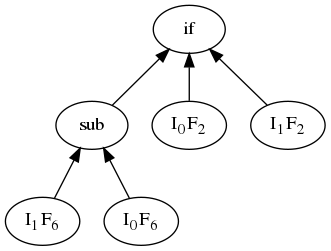}}\hfill
	\subfloat[]{\includegraphics[height=7.5em]{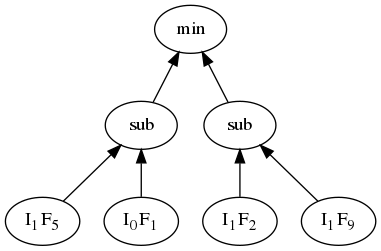}}\hfill
	\caption{An example of a multi-tree similarity function.}
	\label{fig:examplemtstructure}
		\vspace{-1em}
\end{figure}

To tackle these issues, while still maintaining reasonable tree depth and training time, we propose evolving a number of similarity functions concurrently. Using this approach, a pair of instances will be assigned a similarity score by each similarity function, which are then summed together to give a total measure of how similar the instances are\revisionTwo{\footnote{\revisionTwo{Of course, there are a variety of ways to combine similarity scores in the ensemble learning literature, such as taking the maximum output. We use the sum here to increase the stability of the joint similarity function, and to reduce the effect of outliers/edge cases. We hope to investigate this further in future work.}}}. In this regard, each similarity function provides a measure of its \textit{confidence} that two instances should lie in the same cluster, allowing different similarity functions to specialise on different parts of the dataset. This is implemented using GP with a multi-tree approach, where each GP individual contains not only one but multiple trees. An example of this structure is shown in Fig.\ \ref{fig:examplemtstructure} with the number of trees, $t=3$. As all $t$ similarity functions are evolved concurrently in a single individual, a set of cohesive functions will be evolved that work well together, but that are not expected to be good similarity functions independently. In this way, a GP individual can be thought of as a \textit{meta-function}. The core of the clustering process remains the same with this approach, with the only change being that the most similar neighbour for a given instance is \revisionOne{based on the sum of similarities} given by all trees in an individual. This change to the algorithm is shown in Algorithm \ref{multiTreeAlgorithm}.

\begin{figure}[!t]
\small 
	\removelatexerror
	\begin{algorithm}[H]
		\setcounter{AlgoLine}{5}
		
		\caption{Choosing the most similar neighbour to an instance ($I$) in the multi-tree approach for individual $X$.}
		\label{multiTreeAlgorithm}
		\For(\emph{Test neighbour}){$Y \in Neighbours$}{
			$similarity_{sum} = 0$\;
			\For(\emph{Each tree}){$T \in X$}{
				$similarity_{sum} \mathrel{+}= evaluate(T, I, Y)$\;
			}
			\If{$similarity_{sum} > Similarity_{Best}$}{$Neighbour_{Best}= Y$\;}
			
		}
		
	\end{algorithm}
	\vspace{-1em}
\end{figure}
There are several factors that must be considered when extending the proposed algorithm to use a multi-tree approach: how to perform crossover when there are multiple trees to crossover between, and how many trees to use. These two factors will be discussed in the following paragraphs. A third consideration is the maximum tree depth --- we use a smaller tree depth when multiple trees are used, as each tree is able to be more specialised and so does not require as many nodes to produce a good similarity function. Mutation is performed as normal, by randomly choosing a tree to mutate.

\subsubsection{Crossover Strategy}
In standard GP, crossover is performed by selecting two individuals, randomly selecting a subtree from each of these two individuals, and swapping the selected subtrees to produce new offspring. In multi-tree GP, a tree within each individual must also be selected. There are a number of possible methods for doing so \citep{haynes1997crossover,thomason2007novel}, as discussed below:

\paragraph{Random-index crossover} The most obvious method is to randomly select a tree from each individual, which we term \textit{random-index crossover} (RIC). This method may be problematic when applied to our proposed approach, as it reduces the ability of each tree to specialise, by exchanging information between trees which may have different ``niches''. 

\paragraph{Same-index crossover} An alternative method to avoid the limitations of RIC is to always pick two trees at the same index in each individual. For example, selecting the third tree in both individuals. This method, which we call \textit{same-index crossover} (SIC), allows an individual to better develop a number of distinct trees while still encouraging co-operation between individuals through the crossover of related trees. 

\paragraph{All-index crossover} The SIC method can be further extended by performing crossover between every pair of trees simultaneously, i.e.\ crossover between every $i^{th}$ tree in both individuals, where $ i \in [1,t]$ for $t$ trees. This approach, called \textit{all-index crossover} (AIC) allows information exchange to occur more aggressively between individuals, which should increase training efficiency. However, it introduces the requirement that the effect of performing all pairs of crossovers gives a net fitness increase which may limit the exploitation of individual solutions during the EC process. 

We will compare each of these crossover approaches to investigate which type of crossover is most appropriate.

\subsubsection{Number of Trees}
The number of trees used in a multi-tree approach must strike a balance between the performance benefit gained by using a large number of specialised trees and the difficulty in training many trees successfully. When using either the SIC or RIC crossover methods, increasing the number of trees used will reduce the chance proportionally that a given tree is chosen for crossover/mutation, thereby decreasing the rate at which each tree is refined. When the AIC method is used, a larger number of trees increases the probability that a crossover will not improve fitness, as the majority of the trees are unlikely to gain a performance boost when crossed over in the later stages of the training process when small “tweaks” to trees are required to optimise performance. We will investigate the effect of the number of trees used on the fitness obtained later in this paper.

\section{Experiment Design}
\subsection{Benchmark Techniques}
We compare our proposed single-tree approach (GPGC) to a variety of baseline clustering methods, which are listed below. We also compare the single- and multi-tree approaches, to investigate the effectiveness of using additional trees.
\begin{itemize}
	\item $k$-means++ \citep{arthur2007kmeansplusplus}, a commonly used partitional algorithm. Standard $k$-means++ cannot automatically discover the number of clusters, and so K is pre-fixed for this method. \revisionOne{We use this as an example of a relatively simple and  widely used method in the clustering literature.}
	\item OPTICS \citep{ankerst1999optics}, a well-known density-based algorithm.  OPTICS requires a contrast parameter, $\xi$, to be set in order to determine where in the dendrogram the cluster partition is extracted from; we test OPTICS with a range of $\xi$ values in $[0.001, 0.005, 0.01, 0.05, 0.1, 0.2, 0.3, 0.4, 0.5]$ and report the best result in terms of the Adjusted Rand Index (defined in Section \ref{evalMetrics}). 
	\item Two na\"{\i}ve graph-based approaches which connect every instance with an edge to its $n$-nearest neighbours \citep{luxburg2007spectral}. We test with both $n=2$ (called NG-2NN) and $n=3$ (called NG-3NN) in this work. \revisionOnePar{Note that the case where $n=1$ (NG-1NN) is similar to the clustering process used in Algorithm \ref{clusteringAlgorithm}; we exclude NG-1NN as it produces naive solutions with a fixed distance function.}
	\item \revisionOnePar{The Markov Clustering (MCL) algorithm \citep{van2001graph}, another clustering algorithm using a graph-based representation, which simulates random walks through the graph and keeps instances in the same cluster when they have a high number of paths between them.} 
	\item \revisionOnePar{The multi-objective clustering with automatic $k$-determination (MOCK) algorithm \citep{handl2007evolutionary} introduced earlier, as an example of a well-known high-quality EC clustering method.}
\end{itemize}

\subsection{Datasets}
We use a range of synthetic clustering datasets to evaluate the performance of our proposed approach, with varying cluster shapes, numbers of features ($m$), instances ($n$) and clusters ($K$). We avoid using real-world datasets with class labels as done in previous clustering studies, as there is no requirement that classes should correspond well to homogeneous clusters \citep{luxburg2011clustering} --- for example, clustering the well-known Iris dataset will often produce two clusters, as the versicolor and virginica classes overlap significantly in the feature space. The datasets were generated with the popular generators introduced by Handl et al.\ \citep{handl2007evolutionary}. The first generator uses a Gaussian distribution, which produces a range of clusters of varying shapes at low dimensions, but produces increasingly hyper-spherical clusters as $m$ increases. As such, we use this generator only at a small $m$, to produce the datasets shown in Table \ref{table:gaussianClusters}.  The second generator produces clusters using an elliptical distribution, which produces non-hyper-spherical clusters even at large dimensionality. A wide variety of datasets were generated with this distribution, with $m$ varying from 10 to 1000, and $K$ varying from 10 to 100, as shown in Table \ref{table:ellipsoidClusters}. Datasets with $K=10$ clusters have between $50$ and $500$ instances per cluster, whereas datasets with a higher $K$ have between $10$ and $100$ to limit the memory required. These datasets allow our proposed approach to be tested on high-dimensional problems. All datasets are scaled so that each feature is in $[0,1]$ to prevent feature range overly affecting the distance calculations used in the clustering process.  As a generator is used, the cluster that each instance is assigned to is known --- i.e.\ the datasets provide a \textit{gold standard} in the form of a ``cluster label'' for each instance. While this label is not used during training, it is useful for evaluating the clusters produced by the clustering methods.

\begin{table}[!t]
	\renewcommand{\arraystretch}{1.3}
	\footnotesize
	\centering
	\caption{Datasets generated using a Gaussian distribution \citep{handl2007evolutionary}.}
	\label{table:gaussianClusters}
	\begin{tabularx}{0.4\textwidth}{Xrrr}
		
		\toprule
		Name  & \begin{tabular}[c]{@{}c@{}}$m$\end{tabular}&\begin{tabular}[c]{@{}c@{}}$n$\end{tabular}& \begin{tabular}[c]{@{}c@{}}$K$\end{tabular} 
		\\ \midrule
		10d10cGaussian & 10 & 2730 & 10\\
		10d20cGaussian & 10 & 1014 & 20\\
		10d40cGaussian & 10 & 1938 & 40\\
		\bottomrule
	\end{tabularx}%
	\vspace{-1em}
\end{table}

\begin{table}[!t]
	\captionsetup{position=top}
	\renewcommand{\arraystretch}{1.3}
	\footnotesize
	\centering
	\caption{Datasets generated using an Elliptical distribution \citep{handl2007evolutionary}.}
	\label{table:ellipsoidClusters}
	\begin{tabularx}{0.7\textwidth}{@{}l@{}rrrXlrrr@{}}
		\toprule
		Name&\begin{tabular}[c]{@{}c@{}}$m$\end{tabular}&\begin{tabular}[c]{@{}c@{}}$n$\end{tabular}& \begin{tabular}[c]{@{}c@{}}$K$\end{tabular} && Name  & \begin{tabular}[c]{@{}c@{}}$m$\end{tabular}&\begin{tabular}[c]{@{}c@{}}$n$\end{tabular}& \begin{tabular}[c]{@{}c@{}}$K$\end{tabular}\\
		\cmidrule(r){1-4}  \cmidrule(l){6-9}
		10d10c & 10 & 2903 & 10 && 100d10c & 100 & 2893 & 10\\
		10d20c & 10 & 1030 & 20 && 100d20c & 100 & 1339 & 20\\
		10d40c & 10 & 2023 & 40 && 100d40c & 100 & 2212 & 40\\
		10d100c \hspace{4.4mm} & 10 & 5541 & 100 && 1000d10c & 1000 & 2753 & 10\\
		50d10c & 50 & 2699 & 10 && 1000d20c & 1000 & 1088 & 20\\
		50d20c & 50 & 1255 & 20 &&1000d40c & 1000 & 2349 & 40\\
		50d40c & 50 & 2335  & 40 && 1000d100c & 1000 & 6165 & 100\\
		\bottomrule
	\end{tabularx}%
	\captionsetup{position=bottom}
	\vspace{-1em}
\end{table}

\subsection{Parameter Settings}
\revisionOnePar{The non-deterministic methods ($k$-means++, GPGC, MOCK, MCL) were run 30 times, and the mean results were computed. $k$-means++, GPGC  and MOCK were run for 100 iterations, by which time $k$-means++ had achieved convergence. All benchmarks use Euclidean distance. The GP parameter settings for the single- and multi-tree GPGC methods, are based on standard parameters \citep{poli2008field}, and are shown in Table \ref{table:parameterSettings}; the multi-tree (MT) approach uses a smaller maximum tree depth than the single-tree (ST) approach due to having multiple more-specific trees. The MOCK experiments used the attainment score method to select the best solution from the multi-objective pareto front.}

\begin{table}[!t]
	\captionsetup{position=top}
	\renewcommand{\arraystretch}{1.3}
	
	\footnotesize
	\centering
	\caption{Common GP Parameter Settings}
	\label{table:parameterSettings}
	\begin{tabularx}{0.75\textwidth}{ll X ll}
		
		\toprule
		Parameter& Setting && Parameter & Setting\\
		\cmidrule(r){1-2}  \cmidrule(l){4-5}
		Generations & 100 && Population Size & 1024\\
		Mutation & 20\% && Crossover & 80\% \\
		Elitism & top-10 && Selection Type & Tournament\\
		Min. Tree Depth & 2 && Max. Tree Depth & 5 (MT), 7 (ST)\\
		Tournament Size & 7 && Pop. Initialisation & Half-and-half\\
		
		\bottomrule
	\end{tabularx}%
	\captionsetup{position=bottom}
	\vspace{-1em}
\end{table}

\subsection{Evaluation Metrics}
\label{evalMetrics}
To evaluate the performance of each of the clustering algorithms, we use the three measures defined previously (connectedness, sparsity, and separation), as well as \revisionOnePar{the Adjusted Rand Index (ARI), which compares the cluster partition produced by an algorithm to the gold standard provided by the cluster generators in an adjusted-for-chance manner \citep{nguyen2010information}.

Given a cluster partition $C$ produced by an algorithm and a gold standard cluster partition $G$, the ARI is calculated by first generating a contingency table where each entry $n_{ij}$ denotes the number of instances in common between $C_{i}$ and $G_{j}$, where $C_{i}$ is the $i$-th cluster in $C$, and $G_{j}$ is the $j$-th cluster in $G$. In addition, the sum of each row and column is computed, denoted as $a_i$ and $b_j$ respectively. As before, $n$ is the total number of instances. The ARI is then calculated according to Equation (\ref{eqn:ari}), which finds the frequency of occurrence of agreements between the two clusterings, while adjusting for the chance grouping of instances.

	\vspace{-.5em}
\begin{equation}
\text{ARI} = \frac{\textstyle  
\sum_{ij}\binom{n_{ij}}{2} - [\sum_{i}\binom{a_i}{2}\sum_{j}\binom{b_j}{2}]/\binom{n}{2}}
{\frac{1}{2}[\sum_{i}\binom{a_i}{2} + \sum_{j}\binom{b_j}{2}] - [\sum_{i}\binom{a_i}{2}\sum_{j}\binom{b_j}{2}]/\binom{n}{2}}
\label{eqn:ari}
\end{equation}
	\vspace{-.5em}
}

\section{Results and Discussion}
We provide and analyse the results of our experiments in this section. \revisionOne{We begin by comparing each of the proposed multi-tree approaches to the single-tree GPGC approach in order to decide which version of GPGC is the more effective (Section \ref{crossoverResults}). We then compare the best of these approaches, GPGC-AIC, to the benchmark methods to examine how well our proposed method performs relative to existing clustering methods (Section \ref{baselineResults}).} The effect of the number of trees on the performance of the multi-tree approach is analysed in Section \ref{numTreeResults}.

\subsection{GPGC using Multiple Trees}
\label{crossoverResults}

\begin{table}[!t]
	
	\captionsetup{position=top}
	\footnotesize
			\vspace{-1em}
	\caption{Crossover: Datasets using a Gaussian Distribution}
	\vspace{.25em}
	\label{table:crossover:gaussianResults}
\multitree{10d10cGaussian}{
	GPGC & 19.2333 & 21.5 & 41.9123 & 0.2932 & 0.1398 & 0.7501\\
	AIC & 23.7467\textsuperscript{$+$} & 8.7667 & 51.4233\textsuperscript{$+$} & 0.3241\textsuperscript{$-$} & 0.1542\textsuperscript{$+$} & 0.8795\textsuperscript{$+$}\\
	RIC & 24.4737\textsuperscript{$+$} & 8.1333 & 52.3573\textsuperscript{$+$} & 0.3242\textsuperscript{$-$} & 0.1562\textsuperscript{$+$} & 0.8593\textsuperscript{$+$}\\
	SIC & 24.661\textsuperscript{$+$} & 7.6 & 52.889\textsuperscript{$+$} & 0.326\textsuperscript{$-$} & 0.1571\textsuperscript{$+$} & 0.8331\textsuperscript{$+$}\\
}
\multitree{10d20cGaussian}{
	GPGC & 63.0047 & 19.7333 & 47.25 & 0.2684 & 0.3747 & 0.991\\
	AIC & 63.792 & 19.4667 & 47.2527 & 0.2682 & 0.3774 & 0.9803\textsuperscript{$-$}\\
	RIC & 63.2613 & 19.6667 & 47.255 & 0.2687 & 0.3757 & 0.9879\\
	SIC & 63.33 & 19.7333 & 47.254 & 0.2684 & 0.3755 & 0.9907\\
}
\multitree{10d40cGaussian}{
	GPGC & 57.8063 & 34.7667 & 48.5953 & 0.267 & 0.3312 & 0.9582\\
	AIC & 60.3727\textsuperscript{$+$} & 33.7333 & 48.9297\textsuperscript{$+$} & 0.2648\textsuperscript{$-$} & 0.3369\textsuperscript{$+$} & 0.943\textsuperscript{$-$}\\
	RIC & 60.0493\textsuperscript{$+$} & 34.0 & 48.8943\textsuperscript{$+$} & 0.2655\textsuperscript{$-$} & 0.3363\textsuperscript{$+$} & 0.9554\\
	SIC & 58.89 & 34.6 & 48.787 & 0.2665 & 0.3336 & 0.9581\\
}
	
	\vspace{-2em}
\end{table}

\begin{table}[!t]
	\captionsetup{position=top}
	\footnotesize
		\vspace{-2em}
	\caption{Crossover: Datasets using an Elliptical Distribution.}
	\vspace{.25em}
	\label{table:crossover:ellipsoidResults}
\multitree{10d10c}{
	GPGC & 19.1707 & 36.3667 & 61.1867 & 0.157 & 0.0605 & 0.7369\\
	AIC & 21.353\textsuperscript{$+$} & 24.0 & 71.719\textsuperscript{$+$} & 0.1659\textsuperscript{$-$} & 0.0592 & 0.8144\textsuperscript{$+$}\\
	RIC & 20.676\textsuperscript{$+$} & 23.0667 & 73.665\textsuperscript{$+$} & 0.1682\textsuperscript{$-$} & 0.0581\textsuperscript{$-$} & 0.799\textsuperscript{$+$}\\
	SIC & 21.1303\textsuperscript{$+$} & 24.7 & 72.8843\textsuperscript{$+$} & 0.1644\textsuperscript{$-$} & 0.0574 & 0.8058\textsuperscript{$+$}\\
}
\multitree{10d20c}{
	GPGC & 43.69 & 27.7667 & 73.3403 & 0.1504 & 0.0983 & 0.6629\\
	AIC & 49.114\textsuperscript{$+$} & 22.4667 & 77.4163\textsuperscript{$+$} & 0.1542 & 0.1059\textsuperscript{$+$} & 0.6658\\
	RIC & 49.5143\textsuperscript{$+$} & 21.9333 & 77.6527\textsuperscript{$+$} & 0.1544 & 0.106\textsuperscript{$+$} & 0.6564\\
	SIC & 50.2757\textsuperscript{$+$} & 21.8 & 78.344\textsuperscript{$+$} & 0.1551\textsuperscript{$-$} & 0.1062\textsuperscript{$+$} & 0.6773\\
}
\multitree{10d40c}{
	GPGC & 36.5683 & 55.5667 & 70.8793 & 0.1361 & 0.0787 & 0.5792\\
	AIC & 33.425\textsuperscript{$-$} & 49.2 & 76.6153\textsuperscript{$+$} & 0.1385 & 0.0724\textsuperscript{$-$} & 0.5215\\
	RIC & 31.9217\textsuperscript{$-$} & 53.1667 & 75.1063\textsuperscript{$+$} & 0.1375 & 0.0704\textsuperscript{$-$} & 0.4863\textsuperscript{$-$}\\
	SIC & 31.95\textsuperscript{$-$} & 51.1 & 75.9283\textsuperscript{$+$} & 0.1399 & 0.0706\textsuperscript{$-$} & 0.5386\\
}
\multitree{10d100c}{
	GPGC & 31.802 & 109.4667 & 73.9587 & 0.1312 & 0.0658 & 0.4243\\
	AIC & 32.4037 & 106.4 & 75.5567 & 0.1306 & 0.0658 & 0.421\\
	RIC & 30.835 & 134.9 & 72.567 & 0.1268 & 0.0639 & 0.4424\\
	SIC & 31.783 & 113.4333 & 74.1027 & 0.1297 & 0.0657 & 0.444\\
}
\multitree{50d10c}{
	GPGC & 31.662 & 12.5667 & 57.829 & 0.4453 & 0.2756 & 0.9623\\
	AIC & 42.492\textsuperscript{$+$} & 10.0 & 59.5203\textsuperscript{$+$} & 0.472\textsuperscript{$-$} & 0.3406\textsuperscript{$+$} & 0.9865\textsuperscript{$+$}\\
	RIC & 41.2063\textsuperscript{$+$} & 10.0333 & 60.1287\textsuperscript{$+$} & 0.4653\textsuperscript{$-$} & 0.3283\textsuperscript{$+$} & 0.977\textsuperscript{$+$}\\
	SIC & 40.211\textsuperscript{$+$} & 10.4 & 59.8547\textsuperscript{$+$} & 0.4592 & 0.3195\textsuperscript{$+$} & 0.9691\\
}
\multitree{50d20c}{
	GPGC & 30.3563 & 25.2333 & 50.6477 & 0.3498 & 0.2344 & 0.8068\\
	AIC & 36.783\textsuperscript{$+$} & 21.0667 & 51.821\textsuperscript{$+$} & 0.3585 & 0.2732\textsuperscript{$+$} & 0.8369\\
	RIC & 34.8187\textsuperscript{$+$} & 21.5 & 51.3273 & 0.3619\textsuperscript{$-$} & 0.2677\textsuperscript{$+$} & 0.8406\\
	SIC & 34.063\textsuperscript{$+$} & 22.1 & 51.3097 & 0.3578\textsuperscript{$-$} & 0.2605\textsuperscript{$+$} & 0.8476\\
}
\multitree{50d40c}{
	GPGC & 29.5833 & 49.8 & 54.5757 & 0.3044 & 0.1961 & 0.7263\\
	AIC & 34.358\textsuperscript{$+$} & 45.2333 & 56.0917\textsuperscript{$+$} & 0.308 & 0.2201\textsuperscript{$+$} & 0.8098\textsuperscript{$+$}\\
	RIC & 32.1273\textsuperscript{$+$} & 46.1667 & 55.7603\textsuperscript{$+$} & 0.3057 & 0.2089\textsuperscript{$+$} & 0.7212\\
	SIC & 32.2747\textsuperscript{$+$} & 46.3333 & 55.752\textsuperscript{$+$} & 0.3071 & 0.2078\textsuperscript{$+$} & 0.7756\\
}
\multitree{100d10c}{
	GPGC & 39.3957 & 10.4333 & 47.8547 & 0.6111 & 0.5448 & 0.9932\\
	AIC & 44.4043\textsuperscript{$+$} & 9.8 & 48.0887 & 0.6213 & 0.5802\textsuperscript{$+$} & 0.9976\\
	RIC & 44.8793\textsuperscript{$+$} & 9.6667 & 48.159 & 0.6219 & 0.584\textsuperscript{$+$} & 0.9965\\
	SIC & 42.8657 & 10.0333 & 48.1253 & 0.6172 & 0.5688 & 0.9958\\
}
\multitree{100d20c}{
	GPGC & 28.1833 & 22.2 & 38.3887 & 0.5273 & 0.4485 & 0.8828\\
	AIC & 31.995\textsuperscript{$+$} & 20.6 & 38.218 & 0.5353 & 0.4935\textsuperscript{$+$} & 0.9174\textsuperscript{$+$}\\
	RIC & 31.306\textsuperscript{$+$} & 20.6667 & 38.6293 & 0.5302 & 0.4813\textsuperscript{$+$} & 0.9052\\
	SIC & 31.5857\textsuperscript{$+$} & 20.5 & 38.3003 & 0.5376 & 0.4915\textsuperscript{$+$} & 0.9207\textsuperscript{$+$}\\
}
\multitree{100d40c}{
	GPGC & 21.596 & 50.2 & 39.6953 & 0.4355 & 0.2821 & 0.7238\\
	AIC & 25.018\textsuperscript{$+$} & 45.9333 & 40.64\textsuperscript{$+$} & 0.4399 & 0.3155\textsuperscript{$+$} & 0.7705\\
	RIC & 24.499\textsuperscript{$+$} & 47.5667 & 40.676\textsuperscript{$+$} & 0.4375 & 0.3112\textsuperscript{$+$} & 0.7768\textsuperscript{$+$}\\
	SIC & 23.6383\textsuperscript{$+$} & 49.2 & 39.8987 & 0.4376 & 0.3043\textsuperscript{$+$} & 0.792\textsuperscript{$+$}\\
}
\multitree{1000d10c}{
	GPGC & 11.6039 & 10.1 & 15.004 & 2.1322 & 1.7038 & 0.9801\\
	AIC & 12.597\textsuperscript{$+$} & 9.7333 & 14.917 & 2.1262 & 1.7839\textsuperscript{$+$} & 0.9871\textsuperscript{$+$}\\
	RIC & 12.5471\textsuperscript{$+$} & 9.6667 & 15.037 & 2.1218 & 1.7755\textsuperscript{$+$} & 0.9838\\
	SIC & 12.482\textsuperscript{$+$} & 9.6 & 15.1773 & 2.1152 & 1.7536\textsuperscript{$+$} & 0.9781\\
}
\multitree{1000d20c}{
	GPGC & 9.219 & 23.1333 & 11.9877 & 1.5389 & 1.325 & 0.8336\\
	AIC & 11.4834\textsuperscript{$+$} & 19.6333 & 12.4253\textsuperscript{$+$} & 1.5746 & 1.5114\textsuperscript{$+$} & 0.8095\\
	RIC & 11.3739\textsuperscript{$+$} & 19.5333 & 12.2193 & 1.5805\textsuperscript{$-$} & 1.5331\textsuperscript{$+$} & 0.7897\\
	SIC & 10.9564\textsuperscript{$+$} & 19.4 & 12.2603 & 1.5886\textsuperscript{$-$} & 1.4979\textsuperscript{$+$} & 0.8035\\
}
\multitree{1000d40c}{
	GPGC & 8.4841 & 47.5 & 13.9517 & 1.3761 & 1.0059 & 0.7966\\
	AIC & 10.1415\textsuperscript{$+$} & 42.5 & 14.152 & 1.3865 & 1.1301\textsuperscript{$+$} & 0.8035\\
	RIC & 10.0136\textsuperscript{$+$} & 42.5667 & 14.166 & 1.3845 & 1.1257\textsuperscript{$+$} & 0.8319\\
	SIC & 9.6639\textsuperscript{$+$} & 44.0667 & 14.173 & 1.3824 & 1.0861\textsuperscript{$+$} & 0.8279\\
}
\multitree{1000d100c}{
	GPGC & 7.9119 & 132.5333 & 15.7867 & 1.1722 & 0.7614 & 0.8387\\
	AIC & 9.9028\textsuperscript{$+$} & 117.2 & 15.9577 & 1.1889\textsuperscript{$-$} & 0.9013\textsuperscript{$+$} & 0.9155\textsuperscript{$+$}\\
	RIC & 9.0555\textsuperscript{$+$} & 119.8667 & 15.9577\textsuperscript{$+$} & 1.1864\textsuperscript{$-$} & 0.8389\textsuperscript{$+$} & 0.863\\
	SIC & 8.5584 & 124.6667 & 15.956\textsuperscript{$+$} & 1.1722 & 0.7966 & 0.8525\\
}
	
	\vspace{-2em}
	\captionsetup{position=bottom}
\end{table}


To further improve the performance of the proposed GPGC approach, we proposed an extension to use a multi-tree GP design in Section \ref{multiTree}. To analyse the effectiveness of this extension, and determine which type of multi-tree crossover is most effective, we evaluated the three crossover methods (RIC, SIC, AIC) against the single-tree GPGC approach. We used $t=7$ trees based on initial tests --- the effect of varying $t$ is discussed further in Section \ref{numTreeResults}. Tables \ref{table:crossover:gaussianResults} and \ref{table:crossover:ellipsoidResults} show the results of these experiments on the datasets generated using a Gaussian and elliptical distribution respectively. For each of the four methods, we provide the (mean) number of clusters ($K$), as well as four metrics of cluster quality: fitness achieved, connectedness (Conn), sparsity (Spar), separation (Sep), \revisionOne{and the ARI}. Connectedness, sparsity, and separation are defined in the same way as in the fitness function.  We performed a \revisionThree{two-tailed Mann Whitney U-Test at a 95\% confidence interval comparing each of the multi-tree approaches to the single-tree approach on each of the metrics. A ``$+$'' indicates a method is significantly better than the single-tree GPGC method, a ``$-$'' indicates it is significantly worse, and no symbol indicates no significant difference was found. For all metrics except for sparsity, a larger value indicates a better result.}

The most noticeable result of using a multi-tree approach is that the fitness achieved by the GP process is significantly improved across all datasets with the exception of the  \revisionThree{10d20cGaussian, }10d40c and 10d100c datasets, where the multi-tree approaches were significantly worse or had similar fitness to GPGC. On the datasets generated using a Gaussian distribution, the multi-tree approaches are able to find the number of clusters much accurately on 10d10cGaussian, and achieve a significantly higher ARI result. On the 10d40cGaussian dataset, both AIC and RIC achieved significantly better fitness, connectedness, sparsity, and separation than GPGC. \revisionThree{While AIC is significantly worse than GPGC on 10d20cGaussian and 10d40cGaussian, the decrease of \textapprox1.5\% ARI is not very meaningful given it gained 13\% ARI on 10d10cGaussian.}

The multi-tree approaches also tend to produce clusters that are both better connected and better separated than GPGC on the datasets generated with an elliptical distribution. It seems that using multiple trees allows the GP evolutionary process to better separate clusters, while still ensuring that similar instances are placed in the same cluster. Sparsity is either increased (i.e.\ made worse) or is similar compared to GPGC when a multi-tree approach is used --- this suggests that the single tree approach was overly favouring reducing sparsity at the expense of the overall fitness. Another interesting pattern is that the number of clusters ($K$) found by the multi-tree approaches was always lower than that found by GPGC; given that GPGC tended to over-estimate $K$, this can be seen as further evidence that using multiple trees improves clustering performance. Furthermore, a smaller $K$ is likely to directly improve connectedness as more instances will have neighbours in the same cluster, and separation since having fewer clusters increases the average distance between neighbouring clusters. 

In terms of the ARI, the multi-tree approaches were significantly better than GPGC on a number of elliptically-generated datasets, with the RIC, SIC, and AIC methods being significantly better on 3, 3, and 6 datasets respectively. Both the AIC and RIC methods have significantly better fitness than GPGC on \revisionThree{many of }these datasets, while SIC is not significantly better on \revisionThree{100d10c or} 1000d100c. 

\revisionThreePar{
To better understand which of the three multi-tree methods have the highest performance, we analysed the ARI results further as these give the best overall evaluation of how the multi-tree methods compare to the gold standard. We performed a Kruskal-Wallis rank sum test (at a 5\% significance level) followed by post-hoc pair-wise analysis using Dunn's test. The summary of this testing is shown in Table \ref{postHocResults}.

\begin{table}[]
	\footnotesize
	\centering
	\caption{\revisionThree{Summary of ARI post-hoc analysis findings. For each dataset, all results with a p-value below 0.05 (5\% significance level) are shown. ``AIC \textgreater GPGC'' indicates that AIC had a significantly better ARI than GPGC on the given dataset, with a given p-value.}}
	\label{postHocResults}
		\setlength{\tabcolsep}{.4em}
		\begin{tabularx}{.99\textwidth}{@{}Xrlrlrlrl@{}}
			
			\toprule
			Dataset & \multicolumn{1}{r}{Finding} & \multicolumn{1}{l}{p-value} & \multicolumn{1}{r}{Finding} & \multicolumn{1}{l}{p-value} & \multicolumn{1}{r}{Finding} & \multicolumn{1}{l}{p-value} & \multicolumn{1}{r}{Finding} & \multicolumn{1}{l}{p-value} \\ \midrule
			10d10cG & AIC \textgreater GPGC & 0.000 & AIC \textgreater SIC & 0.015 & RIC \textgreater GPGC & 0.002 & SIC \textgreater GPGC & 0.034 \\
			10d10c & AIC \textgreater GPGC & 0.002 & SIC \textgreater GPGC & 0.018 & RIC \textgreater GPGC & 0.028 &  &  \\
			10d40c & GPGC \textgreater AIC & 0.048 & GPGC \textgreater RIC & 0.001 & SIC \textgreater RIC & 0.047 &  &  \\
			50d10c & AIC \textgreater GPGC & 0.000 & AIC \textgreater RIC & 0.040 & AIC \textgreater SIC & 0.003 & RIC \textgreater GPGC & 0.016 \\
			50d40c & AIC \textgreater GPGC & 0.003 & AIC \textgreater RIC & 0.003 &  &  &  \\
			100d40c & AIC \textgreater GPGC & 0.013 & RIC \textgreater GPGC & 0.006 & SIC \textgreater GPGC & 0.004 &  &  \\
			1000d100c & AIC \textgreater GPGC & 0.000 & AIC \textgreater RIC & 0.037 & AIC \textgreater SIC & 0.004 &  &  \\ \bottomrule
		\end{tabularx}%
\end{table}

According to the post-hoc analysis, AIC outperformed GPGC 6 times, whereas RIC and SIC outperformed GPGC 4 and 3 times respectively. AIC outperformed RIC and SIC in 3 cases each as well. In one case, SIC outperformed SIC. Furthermore, AIC generally had a smaller p-value when it outperformed GPGC compared to SIC and RIC. Based on these findings, we conclude that AIC is the most effective of the three proposed multi-tree approaches. Based on this, we use GPGC-AIC in the next section to compare to other clustering methods.
}


\subsection{GPGC-AIC compared to the Benchmarks}

\label{baselineResults}

\begin{table}[t]
	\captionsetup{position=top}
	\footnotesize
			\vspace{-2em}
	\caption{Baselines: Datasets using a Gaussian distribution.}
		\vspace{.25em}
	\label{table:baseline:gaussianResults}

\baselines{10d10c}{
	AIC & 8.7667 & 51.4233 & 0.3241 & 0.1542 & 0.8795\\
	$k$-means++ & 10.0000 & 50.4367\textsuperscript{$-$} & 0.3408\textsuperscript{$-$} & 0.1327\textsuperscript{$-$} & 0.8481\textsuperscript{$-$}\\
	MCL & 8.0000 & 52.8400\textsuperscript{$+$} & 0.3233 & 0.1371\textsuperscript{$-$} & 0.9098\\
	MOCK & 13.5667 & 41.3773\textsuperscript{$-$} & 0.2905\textsuperscript{$-$} & 0.1667\textsuperscript{$+$} & 0.9631\textsuperscript{$+$}\\
	NG-2NN & 4.0000 & 45.3100\textsuperscript{$-$} & 0.3165\textsuperscript{$-$} & 0.1928\textsuperscript{$+$} & 0.3683\textsuperscript{$-$}\\
	NG-3NN & 1.0000 & 57.6200\textsuperscript{$+$} & 0.4277\textsuperscript{$-$} & 0.0000\textsuperscript{$-$} & 0.2477\textsuperscript{$-$}\\
	OPT-0.005 & 39.0000 & 27.9000\textsuperscript{$-$} & 0.2274\textsuperscript{$-$} & 0.1027\textsuperscript{$-$} & 0.5718\textsuperscript{$-$}\\
}
\baselines{10d20c}{
AIC & 19.4667 & 47.2527 & 0.2682 & 0.3774 & 0.9803\\
$k$-means++ & 20.0000 & 43.7780\textsuperscript{$-$} & 0.2727 & 0.2933\textsuperscript{$-$} & 0.8721\textsuperscript{$-$}\\
MCL & 20.0000 & 47.2400\textsuperscript{$-$} & 0.2686\textsuperscript{$-$} & 0.3733\textsuperscript{$-$} & 0.9981\textsuperscript{$+$}\\
MOCK & 20.7333 & 45.7043\textsuperscript{$-$} & 0.2652\textsuperscript{$-$} & 0.3590\textsuperscript{$-$} & 0.9900\\
NG-2NN & 19.0000 & 47.2700\textsuperscript{$+$} & 0.2682 & 0.3813\textsuperscript{$+$} & 0.9645\textsuperscript{$-$}\\
NG-3NN & 19.0000 & 47.2700\textsuperscript{$+$} & 0.2682 & 0.3813\textsuperscript{$+$} & 0.9645\textsuperscript{$-$}\\
OPT-0.001 & 26.0000 & 38.2400\textsuperscript{$-$} & 0.2866\textsuperscript{$-$} & 0.2180\textsuperscript{$-$} & 0.8954\textsuperscript{$-$}\\
}
\baselines{10d40c}{
AIC & 33.7333 & 48.9297 & 0.2648 & 0.3369 & 0.943\\
$k$-means++ & 40.0000 & 44.5080\textsuperscript{$-$} & 0.2702\textsuperscript{$-$} & 0.2604\textsuperscript{$-$} & 0.8947\textsuperscript{$-$}\\
MCL & 40.0000 & 47.2400\textsuperscript{$-$} & 0.2635\textsuperscript{$-$} & 0.3352\textsuperscript{$-$} & 0.9991\textsuperscript{$+$}\\
MOCK & 38.0000 & 46.3633\textsuperscript{$-$} & 0.2609\textsuperscript{$-$} & 0.3209\textsuperscript{$-$} & 0.9601\textsuperscript{$+$}\\
NG-2NN & 40.0000 & 46.3300\textsuperscript{$-$} & 0.2614\textsuperscript{$-$} & 0.3322\textsuperscript{$-$} & 0.9841\textsuperscript{$+$}\\
NG-3NN & 37.0000 & 47.5800\textsuperscript{$-$} & 0.2633\textsuperscript{$-$} & 0.3421\textsuperscript{$+$} & 0.9512\\
OPT-0.001 & 55.0000 & 36.3400\textsuperscript{$-$} & 0.2976\textsuperscript{$-$} & 0.1749\textsuperscript{$-$} & 0.8502\textsuperscript{$-$}\\
}

	\vspace{-2em}
	\captionsetup{position=bottom}
\end{table}

\begin{table}[!t]
	\captionsetup{position=top}
	\footnotesize
		\vspace{-2em}
	\caption{Baselines: Datasets using an Elliptical Distribution (Part 1).}
		\vspace{.25em}
	\label{table:baseline:ellipsoidResults}
\baselines{10d10cE}{
	AIC & 24.0 & 71.719 & 0.1659 & 0.0592 & 0.8144\\
	$k$-means++ & 10.0000 & 85.4903\textsuperscript{$+$} & 0.2118\textsuperscript{$-$} & 0.0377\textsuperscript{$-$} & 0.5524\textsuperscript{$-$}\\
	MCL & 15.0000 & 83.3000\textsuperscript{$+$} & 0.2053\textsuperscript{$-$} & 0.0496\textsuperscript{$-$} & 0.7034\textsuperscript{$-$}\\
	MOCK & 17.7667 & 86.0930\textsuperscript{$+$} & 0.1798\textsuperscript{$-$} & 0.0651 & 0.7932\\
	NG-2NN & 9.0000 & 75.0000 & 0.1757\textsuperscript{$-$} & 0.0705\textsuperscript{$+$} & 0.5100\textsuperscript{$-$}\\
	NG-3NN & 5.0000 & 82.3000\textsuperscript{$+$} & 0.1868\textsuperscript{$-$} & 0.0853\textsuperscript{$+$} & 0.3234\textsuperscript{$-$}\\
	OPT-0.05 & 32.0000 & 84.3700\textsuperscript{$+$} & 0.0633\textsuperscript{$-$} & 0.0250\textsuperscript{$-$} & 0.2387\textsuperscript{$-$}\\
}
\baselines{10d20cE}{
	AIC & 22.4667 & 77.4163 & 0.1542 & 0.1059 & 0.6658\\
	$k$-means++ & 20.0000 & 73.3217\textsuperscript{$-$} & 0.1898\textsuperscript{$-$} & 0.0870\textsuperscript{$-$} & 0.4592\textsuperscript{$-$}\\
	MCL & 28.0000 & 69.4300\textsuperscript{$-$} & 0.1734\textsuperscript{$-$} & 0.0946\textsuperscript{$-$} & 0.4508\textsuperscript{$-$}\\
	MOCK & 27.4667 & 79.2910\textsuperscript{$+$} & 0.1518 & 0.1128\textsuperscript{$+$} & 0.7518\textsuperscript{$+$}\\
	NG-2NN & 36.0000 & 69.4500\textsuperscript{$-$} & 0.1312\textsuperscript{$-$} & 0.0944\textsuperscript{$-$} & 0.5841\textsuperscript{$-$}\\
	NG-3NN & 16.0000 & 83.1200\textsuperscript{$+$} & 0.1452\textsuperscript{$-$} & 0.1101\textsuperscript{$+$} & 0.3548\textsuperscript{$-$}\\
	OPT-0.001 & 69.0000 & 45.9400\textsuperscript{$-$} & 0.2104\textsuperscript{$-$} & 0.0525\textsuperscript{$-$} & 0.3437\textsuperscript{$-$}\\
}
\baselines{10d40cE}{
	AIC & 49.2 & 76.6153 & 0.1385 & 0.0724 & 0.5215\\
	$k$-means++ & 40.0000 & 77.1587 & 0.1692\textsuperscript{$-$} & 0.0695 & 0.4173\textsuperscript{$-$}\\
	MCL & 54.0000 & 69.2000\textsuperscript{$-$} & 0.1493\textsuperscript{$-$} & 0.0942\textsuperscript{$+$} & 0.2877\textsuperscript{$-$}\\
	MOCK & 28.8000 & 84.2253\textsuperscript{$+$} & 0.1462\textsuperscript{$-$} & 0.0873\textsuperscript{$+$} & 0.2322\textsuperscript{$-$}\\
	NG-2NN & 43.0000 & 68.4300\textsuperscript{$-$} & 0.1097\textsuperscript{$-$} & 0.0849\textsuperscript{$+$} & 0.2559\textsuperscript{$-$}\\
	NG-3NN & 13.0000 & 73.6200\textsuperscript{$-$} & 0.1306\textsuperscript{$-$} & 0.1072\textsuperscript{$+$} & 0.0815\textsuperscript{$-$}\\
	OPT-0.005 & 93.0000 & 61.4200\textsuperscript{$-$} & 0.1526\textsuperscript{$-$} & 0.0481\textsuperscript{$-$} & 0.2989\textsuperscript{$-$}\\
}
\baselines{10d100cE}{
	AIC & 106.4 & 75.5567 & 0.1306 & 0.0658 & 0.421\\
	$k$-means++ & 100.0000 & 80.3510\textsuperscript{$+$} & 0.1532\textsuperscript{$-$} & 0.0556\textsuperscript{$-$} & 0.3976\\
	MCL & 98.0000 & 77.2500\textsuperscript{$+$} & 0.1360\textsuperscript{$-$} & 0.0703\textsuperscript{$+$} & 0.1253\textsuperscript{$-$}\\
	MOCK & 62.0000 & 86.2440\textsuperscript{$+$} & 0.1370\textsuperscript{$-$} & 0.0743\textsuperscript{$+$} & 0.0869\textsuperscript{$-$}\\
	NG-2NN & 91.0000 & 70.6000\textsuperscript{$-$} & 0.1075\textsuperscript{$-$} & 0.0800\textsuperscript{$+$} & 0.0490\textsuperscript{$-$}\\
	NG-3NN & 26.0000 & 73.4200 & 0.0957\textsuperscript{$-$} & 0.0838\textsuperscript{$+$} & 0.0298\textsuperscript{$-$}\\
	OPT-0.01 & 197.0000 & 69.5100\textsuperscript{$-$} & 0.0913\textsuperscript{$-$} & 0.0424\textsuperscript{$-$} & 0.0539\textsuperscript{$-$}\\
}
\baselines{50d10c}{
	AIC & 10.0 & 59.5203 & 0.472 & 0.3406 & 0.9865\\
	$k$-means++ & 10.0000 & 52.2093\textsuperscript{$-$} & 0.5553\textsuperscript{$-$} & 0.1023\textsuperscript{$-$} & 0.4847\textsuperscript{$-$}\\
	MCL & 12.0000 & 54.0900\textsuperscript{$-$} & 0.5189\textsuperscript{$-$} & 0.1022\textsuperscript{$-$} & 0.6042\textsuperscript{$-$}\\
	MOCK & 14.4333 & 56.6567\textsuperscript{$-$} & 0.4943\textsuperscript{$-$} & 0.2171\textsuperscript{$-$} & 0.8113\textsuperscript{$-$}\\
	NG-2NN & 18.0000 & 53.5700\textsuperscript{$-$} & 0.3720\textsuperscript{$-$} & 0.1676\textsuperscript{$-$} & 0.9673\textsuperscript{$-$}\\
	NG-3NN & 11.0000 & 58.7800\textsuperscript{$-$} & 0.4559\textsuperscript{$-$} & 0.3023\textsuperscript{$-$} & 0.9987\textsuperscript{$+$}\\
	OPT-0.05 & 28.0000 & 67.7200\textsuperscript{$+$} & 0.1963\textsuperscript{$-$} & 0.0805\textsuperscript{$-$} & 0.3685\textsuperscript{$-$}\\
}
\baselines{50d20c}{
	AIC & 21.0667 & 51.821 & 0.3585 & 0.2732 & 0.8369\\
	$k$-means++ & 20.0000 & 44.3167\textsuperscript{$-$} & 0.4287\textsuperscript{$-$} & 0.1678\textsuperscript{$-$} & 0.3528\textsuperscript{$-$}\\
	MCL & 26.0000 & 42.0300\textsuperscript{$-$} & 0.4146\textsuperscript{$-$} & 0.1745\textsuperscript{$-$} & 0.4823\textsuperscript{$-$}\\
	MOCK & 24.3333 & 50.2540\textsuperscript{$-$} & 0.3752\textsuperscript{$-$} & 0.2729 & 0.8837\\
	NG-2NN & 44.0000 & 40.9900\textsuperscript{$-$} & 0.3044\textsuperscript{$-$} & 0.2130\textsuperscript{$-$} & 0.8306\\
	NG-3NN & 23.0000 & 49.0000\textsuperscript{$-$} & 0.3692\textsuperscript{$-$} & 0.2860 & 0.8075\textsuperscript{$-$}\\
	OPT-0.005 & 73.0000 & 32.2000\textsuperscript{$-$} & 0.4550\textsuperscript{$-$} & 0.1287\textsuperscript{$-$} & 0.3856\textsuperscript{$-$}\\
}
\baselines{50d40c}{
	AIC & 45.2333 & 56.0917 & 0.308 & 0.2201 & 0.8098\\
	$k$-means++ & 40.0000 & 50.7367\textsuperscript{$-$} & 0.3519\textsuperscript{$-$} & 0.1397\textsuperscript{$-$} & 0.2538\textsuperscript{$-$}\\
	MCL & 48.0000 & 48.6100\textsuperscript{$-$} & 0.3287\textsuperscript{$-$} & 0.1612\textsuperscript{$-$} & 0.3514\textsuperscript{$-$}\\
	MOCK & 42.6667 & 56.6433\textsuperscript{$+$} & 0.3153\textsuperscript{$-$} & 0.2531\textsuperscript{$+$} & 0.8673\\
	NG-2NN & 86.0000 & 50.0500\textsuperscript{$-$} & 0.2487\textsuperscript{$-$} & 0.1700\textsuperscript{$-$} & 0.7618\textsuperscript{$-$}\\
	NG-3NN & 46.0000 & 56.2500 & 0.2788\textsuperscript{$-$} & 0.2170 & 0.7376\textsuperscript{$-$}\\
	OPT-0.05 & 73.0000 & 53.5300\textsuperscript{$-$} & 0.2391\textsuperscript{$-$} & 0.0975\textsuperscript{$-$} & 0.1632\textsuperscript{$-$}\\
}
\baselines{100d10c}{
	AIC & 9.8 & 48.0887 & 0.6213 & 0.5802 & 0.9976\\
	$k$-means++ & 10.0000 & 45.5490\textsuperscript{$-$} & 0.6945\textsuperscript{$-$} & 0.1306\textsuperscript{$-$} & 0.5617\textsuperscript{$-$}\\
	MCL & 16.0000 & 40.5100\textsuperscript{$-$} & 0.6768\textsuperscript{$-$} & 0.2072\textsuperscript{$-$} & 0.8766\textsuperscript{$-$}\\
	MOCK & 28.7667 & 45.6833\textsuperscript{$-$} & 0.5904\textsuperscript{$-$} & 0.1702\textsuperscript{$-$} & 0.5480\textsuperscript{$-$}\\
	NG-2NN & 16.0000 & 44.5700\textsuperscript{$-$} & 0.5517\textsuperscript{$-$} & 0.2870\textsuperscript{$-$} & 0.9341\textsuperscript{$-$}\\
	NG-3NN & 11.0000 & 45.7700\textsuperscript{$-$} & 0.6465\textsuperscript{$-$} & 0.5125\textsuperscript{$-$} & 0.9886\textsuperscript{$-$}\\
	OPT-0.001 & 92.0000 & 37.5900\textsuperscript{$-$} & 0.3721\textsuperscript{$-$} & 0.1132\textsuperscript{$-$} & 0.4552\textsuperscript{$-$}\\
}
\baselines{100d20c}{
	AIC & 20.6 & 38.218 & 0.5353 & 0.4935 & 0.9174\\
	$k$-means++ & 20.0000 & 34.1000\textsuperscript{$-$} & 0.5946\textsuperscript{$-$} & 0.2321\textsuperscript{$-$} & 0.3738\textsuperscript{$-$}\\
	MCL & 27.0000 & 29.4800\textsuperscript{$-$} & 0.5939\textsuperscript{$-$} & 0.2834\textsuperscript{$-$} & 0.5869\textsuperscript{$-$}\\
	MOCK & 24.6333 & 35.7030\textsuperscript{$-$} & 0.5725\textsuperscript{$-$} & 0.4873 & 0.8969\\
	NG-2NN & 41.0000 & 32.0600\textsuperscript{$-$} & 0.4497\textsuperscript{$-$} & 0.2907\textsuperscript{$-$} & 0.8194\textsuperscript{$-$}\\
	NG-3NN & 25.0000 & 34.6600\textsuperscript{$-$} & 0.5289 & 0.5198\textsuperscript{$+$} & 0.9653\textsuperscript{$+$}\\
	OPT-0.01 & 76.0000 & 28.0900\textsuperscript{$-$} & 0.5045\textsuperscript{$-$} & 0.1678\textsuperscript{$-$} & 0.3683\textsuperscript{$-$}\\
}
\baselines{100d40c}{
	AIC & 45.9333 & 40.64 & 0.4399 & 0.3155 & 0.7705\\
	$k$-means++ & 40.0000 & 35.3247\textsuperscript{$-$} & 0.4916\textsuperscript{$-$} & 0.2258\textsuperscript{$-$} & 0.2681\textsuperscript{$-$}\\
	MCL & 57.0000 & 32.9400\textsuperscript{$-$} & 0.4729\textsuperscript{$-$} & 0.2650\textsuperscript{$-$} & 0.4665\textsuperscript{$-$}\\
	MOCK & 41.8000 & 39.4233\textsuperscript{$-$} & 0.4763\textsuperscript{$-$} & 0.3714\textsuperscript{$+$} & 0.7838\\
	NG-2NN & 91.0000 & 34.6000\textsuperscript{$-$} & 0.3754\textsuperscript{$-$} & 0.2987\textsuperscript{$-$} & 0.7913\\
	NG-3NN & 49.0000 & 36.0800\textsuperscript{$-$} & 0.4309 & 0.3926\textsuperscript{$+$} & 0.7113\textsuperscript{$-$}\\
	OPT-0.001 & 140.0000 & 25.5800\textsuperscript{$-$} & 0.5925\textsuperscript{$-$} & 0.1499\textsuperscript{$-$} & 0.4303\textsuperscript{$-$}\\
}
\vspace{-2em}
\end{table}
\begin{table}[!tp]
	\captionsetup{position=top}
		\vspace{-2em}
	\footnotesize
	\ContinuedFloat
	\caption{Baselines: Datasets using an Elliptical Distribution (Part 2).}

\baselines{1000d10c}{
	AIC & 9.7333 & 14.917 & 2.1262 & 1.7839 & 0.9871\\
	$k$-means++ & 10.0000 & 13.8387\textsuperscript{$-$} & 2.3471\textsuperscript{$-$} & 0.3852\textsuperscript{$-$} & 0.4884\textsuperscript{$-$}\\
	MCL & 10.0000 & 12.1500\textsuperscript{$-$} & 2.4070\textsuperscript{$-$} & 0.4446\textsuperscript{$-$} & 0.4742\textsuperscript{$-$}\\
	MOCK & 16.0000 & 14.4683\textsuperscript{$-$} & 2.0785\textsuperscript{$-$} & 0.9947\textsuperscript{$-$} & 0.8001\textsuperscript{$-$}\\
	NG-2NN & 21.0000 & 14.9400\textsuperscript{$+$} & 1.6810\textsuperscript{$-$} & 0.6104\textsuperscript{$-$} & 0.9321\textsuperscript{$-$}\\
	NG-3NN & 10.0000 & 15.4800\textsuperscript{$+$} & 2.0780\textsuperscript{$-$} & 1.5410\textsuperscript{$-$} & 0.9472\textsuperscript{$-$}\\
	OPT-0.005 & 86.0000 & 13.1300\textsuperscript{$-$} & 1.1380\textsuperscript{$-$} & 0.4545\textsuperscript{$-$} & 0.3428\textsuperscript{$-$}\\
}
\baselines{1000d20c}{
	AIC & 19.6333 & 12.4253 & 1.5746 & 1.5114 & 0.8095\\
	$k$-means++ & 20.0000 & 9.6912\textsuperscript{$-$} & 1.8854\textsuperscript{$-$} & 0.8319\textsuperscript{$-$} & 0.3761\textsuperscript{$-$}\\
	MCL & 24.0000 & 9.3000\textsuperscript{$-$} & 1.8280\textsuperscript{$-$} & 0.7475\textsuperscript{$-$} & 0.3394\textsuperscript{$-$}\\
	MOCK & 25.5333 & 10.7653\textsuperscript{$-$} & 1.7060\textsuperscript{$-$} & 1.4230\textsuperscript{$-$} & 0.8961\\
	NG-2NN & 47.0000 & 9.7950\textsuperscript{$-$} & 1.4090\textsuperscript{$-$} & 1.0140\textsuperscript{$-$} & 0.7363\textsuperscript{$-$}\\
	NG-3NN & 26.0000 & 10.4500\textsuperscript{$-$} & 1.5310\textsuperscript{$-$} & 1.3790\textsuperscript{$-$} & 0.9447\textsuperscript{$+$}\\
	OPT-0.001 & 67.0000 & 6.9620\textsuperscript{$-$} & 2.1420\textsuperscript{$-$} & 0.5886\textsuperscript{$-$} & 0.4529\textsuperscript{$-$}\\
}
\baselines{1000d40c}{
	AIC & 42.5 & 14.152 & 1.3865 & 1.1301 & 0.8035\\
	$k$-means++ & 40.0000 & 11.7390\textsuperscript{$-$} & 1.5563\textsuperscript{$-$} & 0.6324\textsuperscript{$-$} & 0.2189\textsuperscript{$-$}\\
	MCL & 47.0000 & 10.7500\textsuperscript{$-$} & 1.5000\textsuperscript{$-$} & 0.6757\textsuperscript{$-$} & 0.1574\textsuperscript{$-$}\\
	MOCK & 41.7000 & 13.5683\textsuperscript{$-$} & 1.4878\textsuperscript{$-$} & 1.2313\textsuperscript{$+$} & 0.8872\textsuperscript{$+$}\\
	NG-2NN & 94.0000 & 12.1200\textsuperscript{$-$} & 1.2090\textsuperscript{$-$} & 0.8463\textsuperscript{$-$} & 0.7397\textsuperscript{$-$}\\
	NG-3NN & 52.0000 & 13.7200\textsuperscript{$-$} & 1.3460\textsuperscript{$-$} & 1.1300 & 0.8978\textsuperscript{$+$}\\
	OPT-0.001 & 132.0000 & 9.3390\textsuperscript{$-$} & 1.8410\textsuperscript{$-$} & 0.4818\textsuperscript{$-$} & 0.4223\textsuperscript{$-$}\\
}
\baselines{1000d100c}{
	AIC & 117.2 & 15.9577 & 1.1889 & 0.9013 & 0.9155\\
	$k$-means++ & 100.0000 & 14.4633\textsuperscript{$-$} & 1.2701\textsuperscript{$-$} & 0.5232\textsuperscript{$-$} & 0.1030\textsuperscript{$-$}\\
	MCL & 204.0000 & 11.4800\textsuperscript{$-$} & 1.2420\textsuperscript{$-$} & 0.4823\textsuperscript{$-$} & 0.1892\textsuperscript{$-$}\\
	MOCK & 64.3333 & 16.6880\textsuperscript{$+$} & 1.2648\textsuperscript{$-$} & 1.1456\textsuperscript{$+$} & 0.4340\textsuperscript{$-$}\\
	NG-2NN & 229.0000 & 14.6900\textsuperscript{$-$} & 1.0140\textsuperscript{$-$} & 0.6930\textsuperscript{$-$} & 0.7611\textsuperscript{$-$}\\
	NG-3NN & 132.0000 & 15.9500\textsuperscript{$-$} & 1.1070\textsuperscript{$-$} & 0.9329 & 0.8625\textsuperscript{$-$}\\
	OPT-0.05 & 182.0000 & 15.9700 & 0.9169\textsuperscript{$-$} & 0.4142\textsuperscript{$-$} & 0.1060\textsuperscript{$-$}\\
}
		\vspace{-2em}
	\captionsetup{position=bottom}
\end{table}

Tables \ref{table:baseline:gaussianResults} and \ref{table:baseline:ellipsoidResults} show how the proposed GPGC-AIC method compares to the six benchmarks across the datasets tested. For each of the seven methods, we provide the (mean) number of clusters ($K$), as well as the same four metrics of cluster quality as before. Note that $k$-means++ requires $K$ to be pre-defined, and so always obtains the correct $K$ value. We use the same \revisionThree{two-tailed Mann Whitney U-Test} as in Section \ref{crossoverResults}: \revisionTwo{a ``$+$'' indicates that a baseline method is significantly better than \revisionThree{the AIC} method, a ``$-$'' indicates it is significantly worse, and no symbol indicates no significant difference is found.}

\revisionOnePar{Table \ref{table:baseline:gaussianResults} shows the results on the datasets that were generated using a Gaussian distribution. In terms of the ARI, the AIC method is significantly worse than either the MCL or MOCK method across the three datasets, but generally outperforms all the other baselines. As these datasets were generated with a Gaussian distribution, they tend to contain very well-formed hyper-spherical clusters, and so methods such as MCL are very effective at clustering these correctly.}

\revisionOnePar{
On the results for the datasets generated using an elliptical distribution, shown in Table \ref{table:baseline:ellipsoidResults}, GPGC is significantly better than all baselines excluding MOCK across all datasets containing 10 features (10d*c). While the MOCK method is competitive (or better) on the datasets with 10 and 20 clusters, it achieves a very poor ARI on the more difficult datasets with 40 and 100 clusters, where the AIC method is clearly superior. The remaining baseline methods are nearly always significantly worse than AIC on these datasets. The NG baselines are particularly inconsistent, with the number of clusters and ARI values varying by up to three times depending on the number of nearest neighbours chosen. AIC can also automatically find the number of clusters much more accurately than OPTICS, and produces less sparse and more separated clusters than $k$-means++ across these datasets. In contrast to the previous datasets, the MCL method struggles significantly with these non-hyper-spherical datasets --- a pattern that is also true for the remaining datasets and which highlights a key weakness with the MCL method. Similar patterns are seen across the 50d*c datasets, with the exception being on 50d10c where NG-3NN achieves a near perfect result.
} 
\revisionOnePar{On the high-dimensionality datasets (100d*c, 1000d*c), the MOCK method has a high variance in accuracy, with ARI values ranging from 0.434 to 0.897. In contrast, the AIC method achieves consistently good performance, with the lowest ARI achieved being 0.771 and the highest being 0.998. While the MOCK method is superior \revisionThree{on 1000d40c}, its inconsistency \revisionThree{on the other datasets} makes it harder to use confidently in practice. On this set of datasets, the NG baselines achieve better results than previously in terms of the ARI, but GPGC is only ever significantly worse than one of NG-2NN and NG-3NN at most. GPGC also often has significantly better connectedness and separation than one or both of the NG baselines, suggesting it is a more consistent choice given that it is difficult to determine the number of nearest neighbours in advance (as the NG methods assume). All of the graph-based approaches are superior to $k$-means++ (due to the non-hyper-spherical cluster shape), and OPTICS across these datasets. The AIC method also appears to be the method which predicts $K$ most accurately overall across these datasets, especially where $K=100$.
}

\subsubsection{Summary}
\revisionOnePar{Table \ref{table:baseline:wins} shows the number of datasets for which each clustering method was the winner (i.e.\ highest mean ARI). The AIC method was the most successful, with six wins compared to four for the closest methods (NG-3NN and MOCK). This is consistent with our previous analysis, which showed AIC was the most consistent and best-performing method across datasets with high dimensionality, and was competitive with MOCK on the remaining datasets. The remaining baselines, with the exception of MCL, were almost always outperformed by at least one of AIC and MOCK. Given that MOCK is a multi-objective approach, we are hopeful that a future multi-objective variation of AIC would be able to improve the GPGC method even further and allow it to achieve a higher number of wins.}

\begin{table}[]
	\centering
	\caption{\revisionOnePar{Number of wins of each algorithm across the 17 datasets.}}
	\label{table:baseline:wins}
	\begin{tabular}{@{}lllllll@{}}
		\toprule
		AIC & $k$-means++ & MCL & MOCK & NG-2NN & NG-3NN & OPTICS \\ \midrule
	6 & 0 & 2 & 4 & 1 & 4 & 0\\ \bottomrule
	\end{tabular}
\vspace{-1em}
\end{table}

\subsection{Number of Trees: Effect on Fitness}
\label{numTreeResults}

\begin{figure*}[!tp] 
	\centering
	\mtGraph{a}{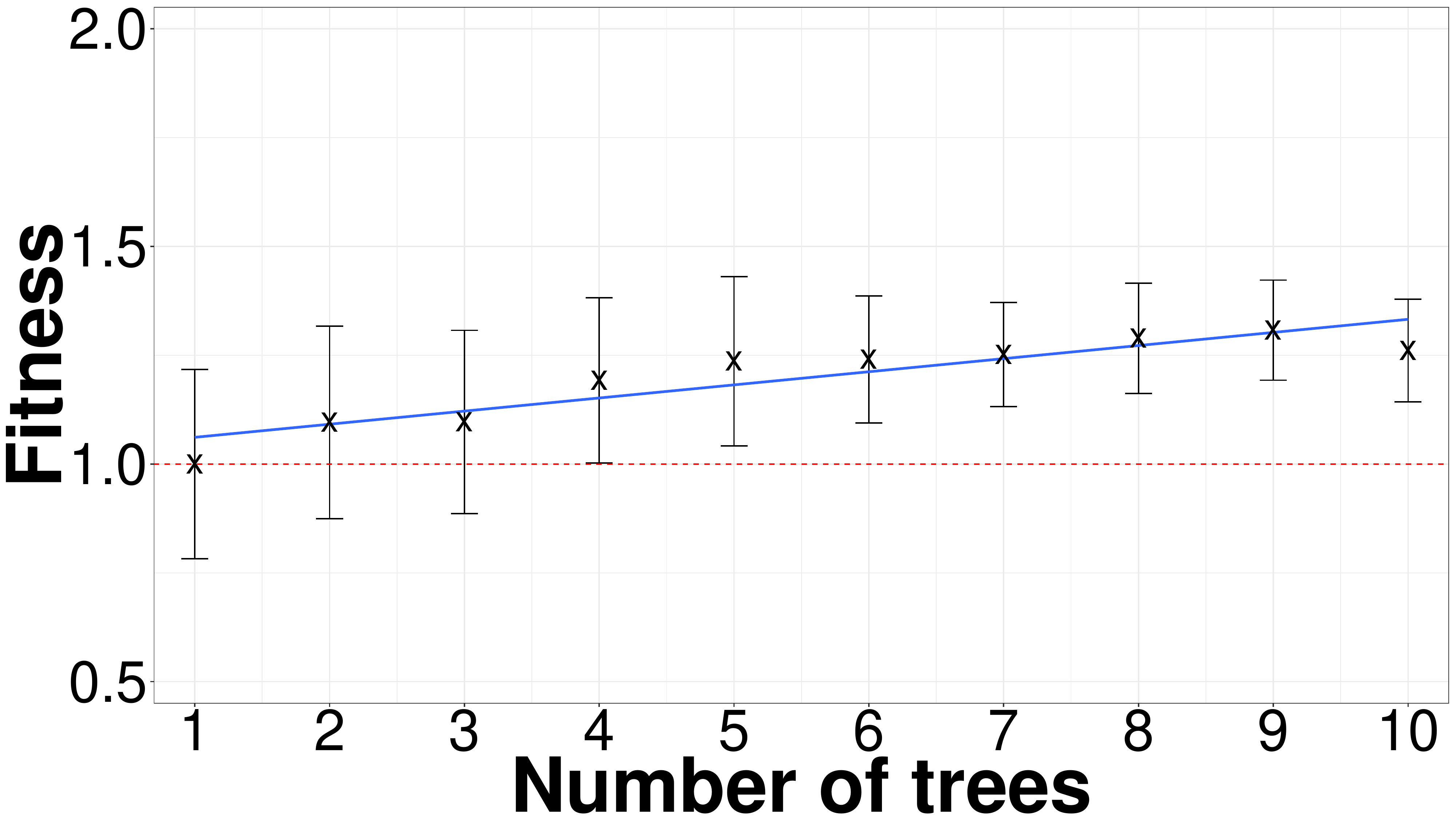}
	\mtGraph{b}{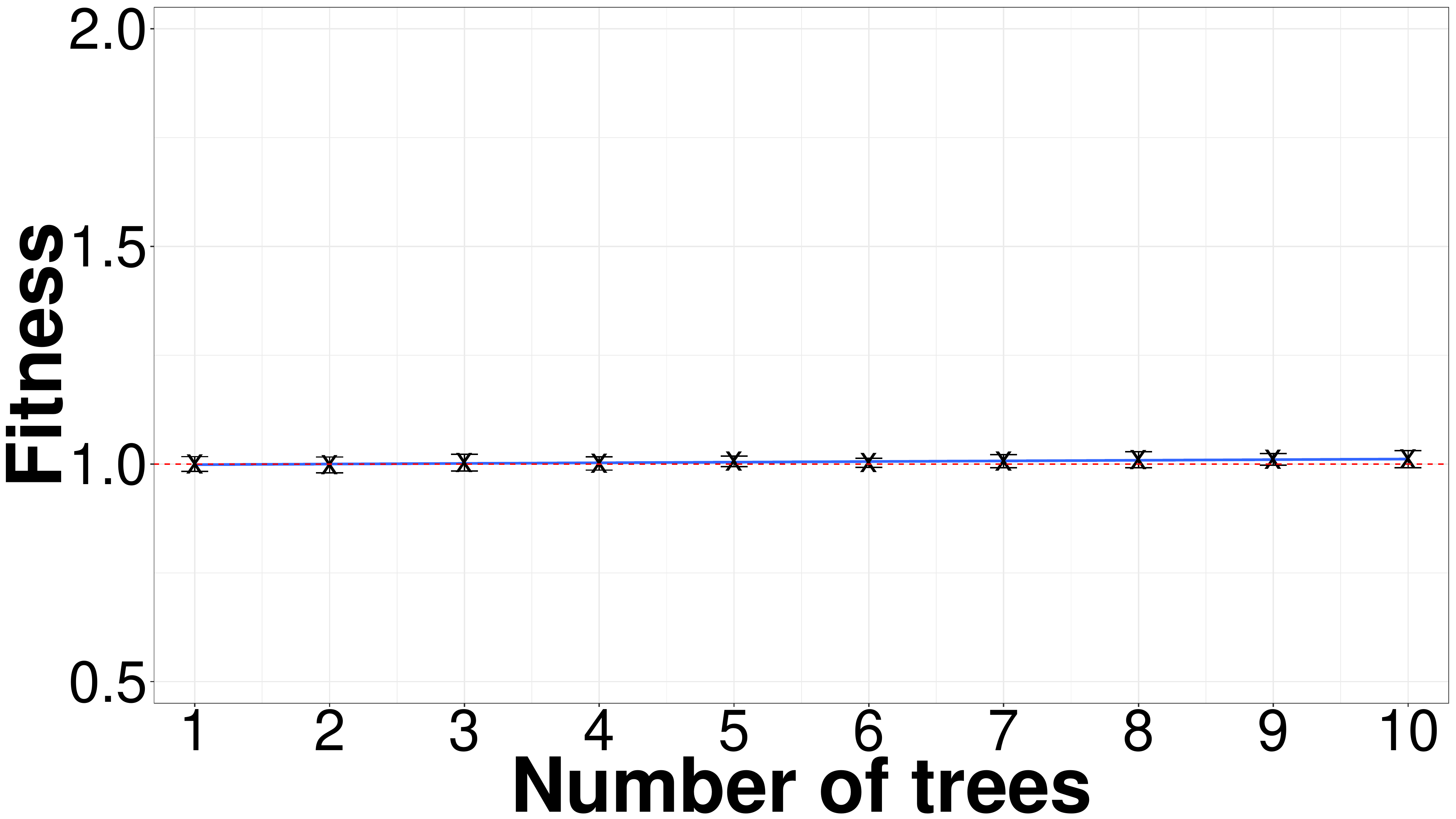}
	\mtGraph{c}{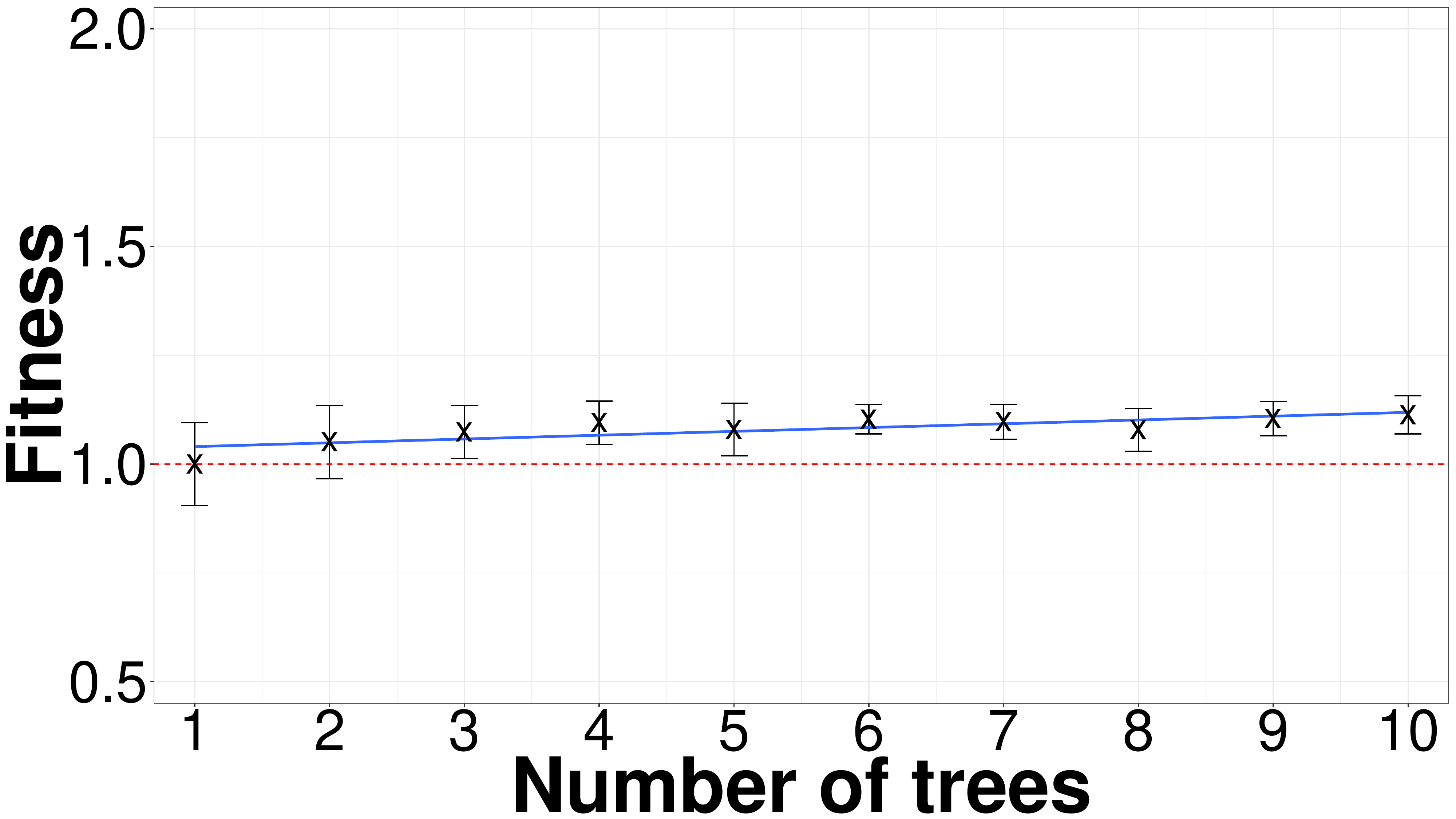}\\ 
	\mtGraph{d}{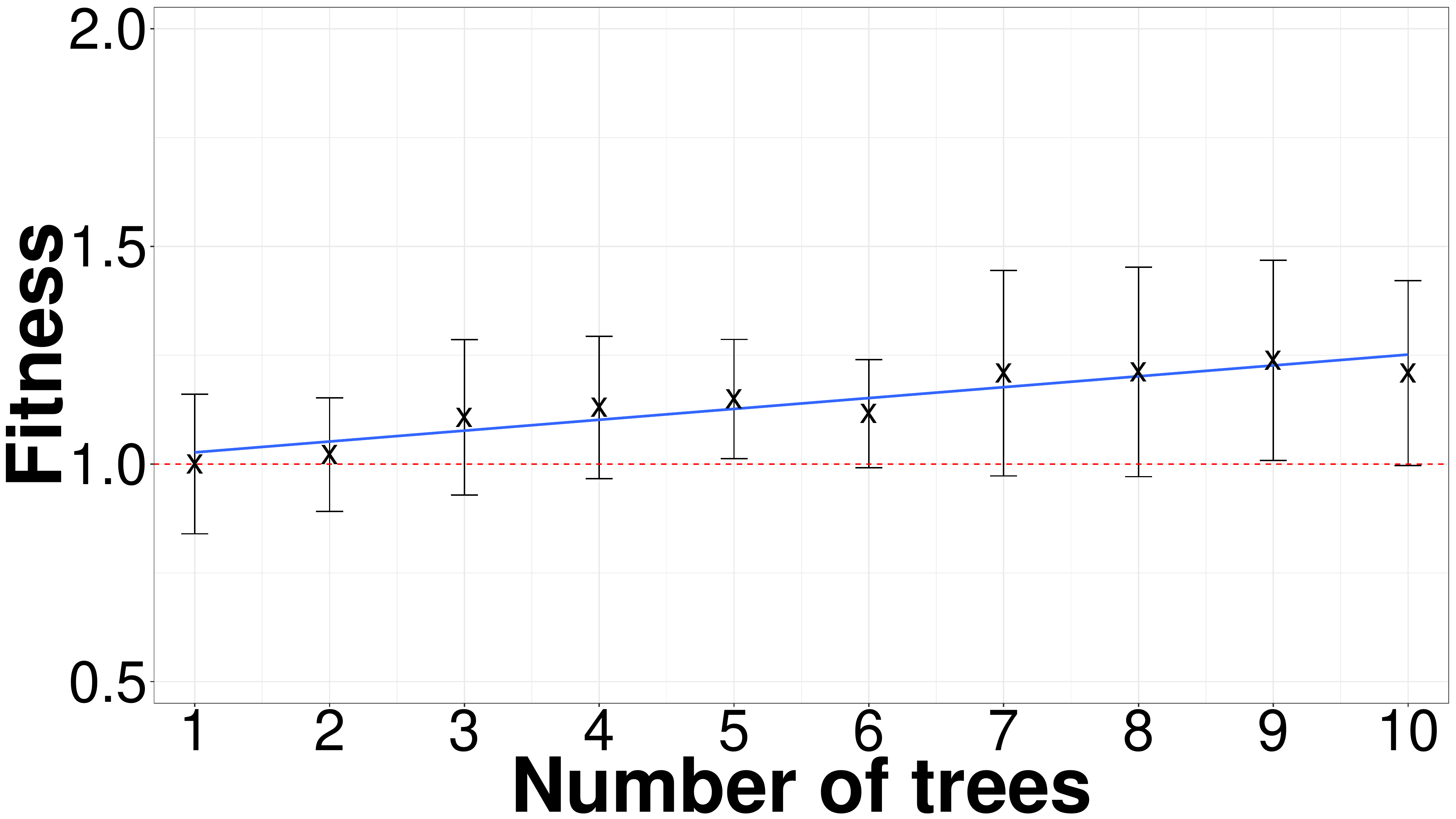}
	\mtGraph{e}{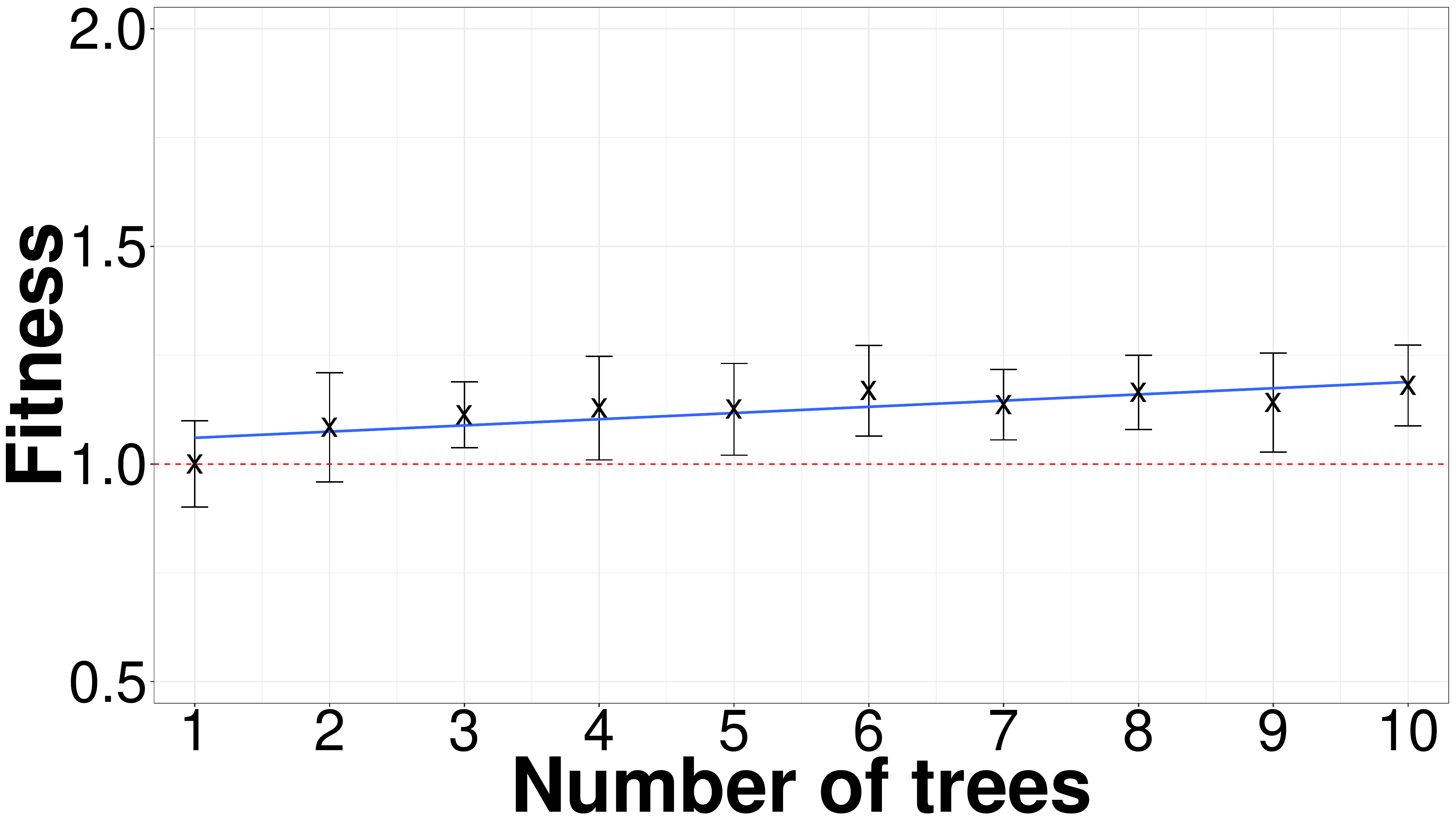}
	\mtGraph{f}{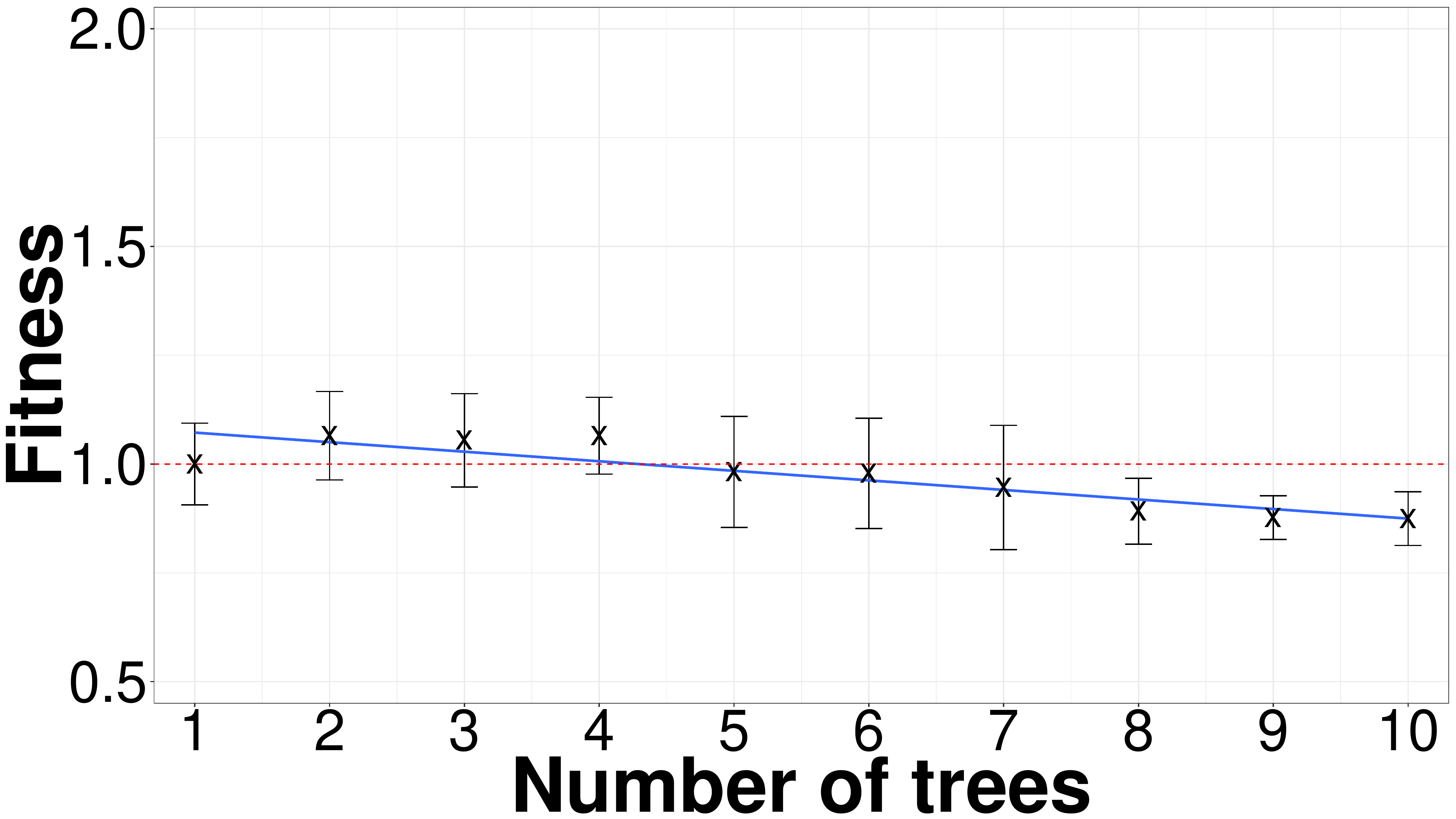}\\ 
	\mtGraph{g}{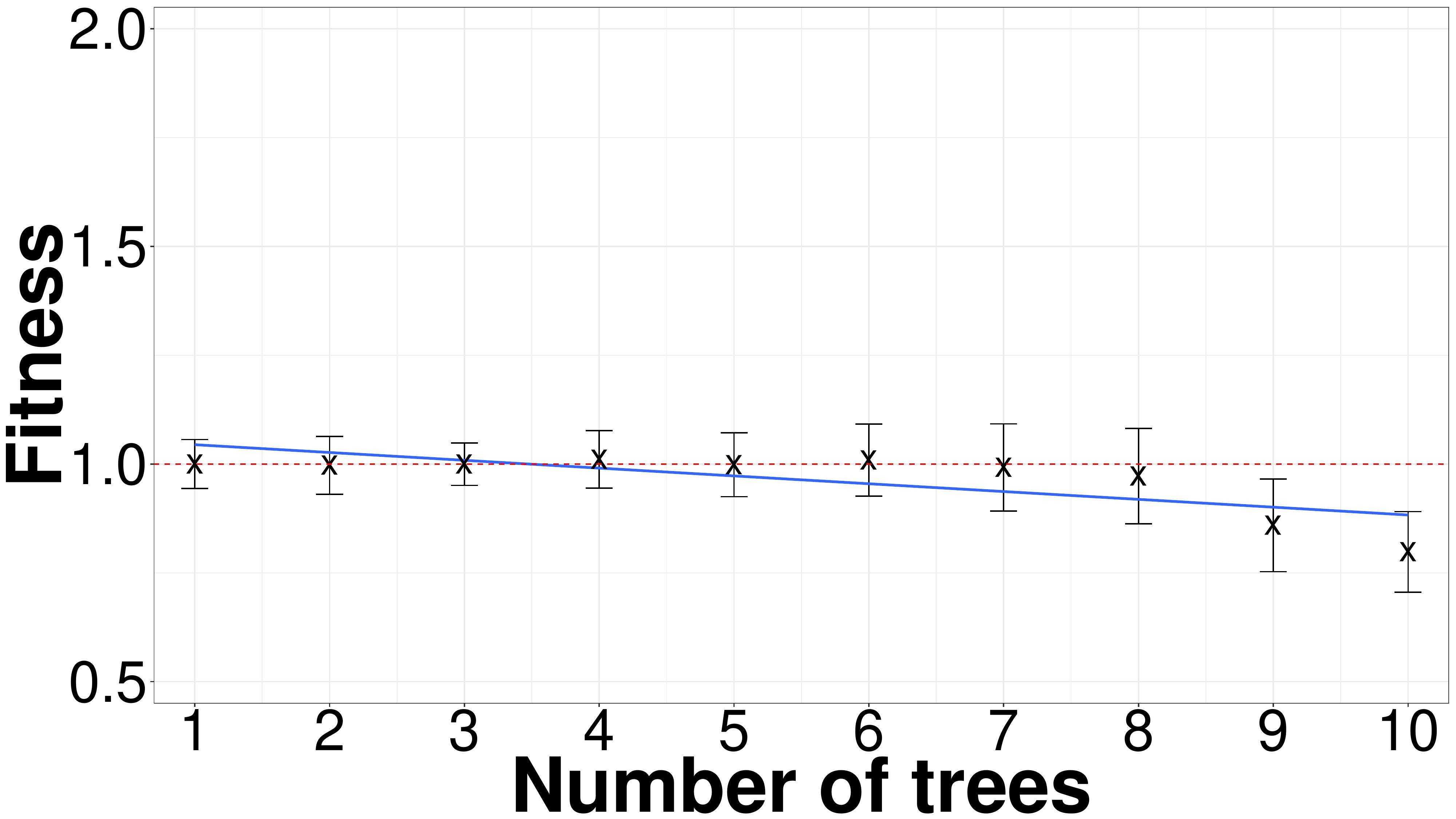}
	\mtGraph{h}{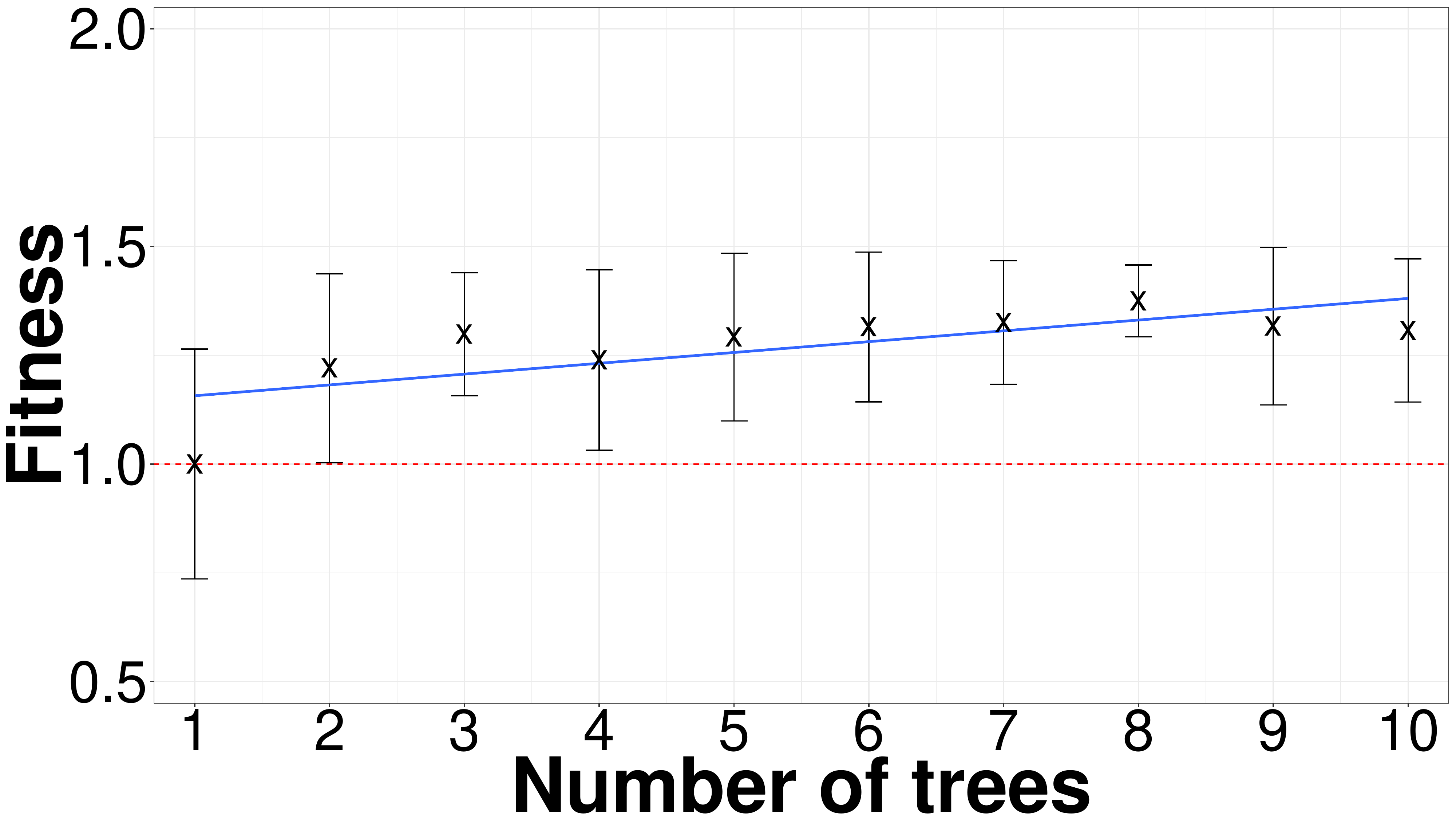}
	\mtGraph{i}{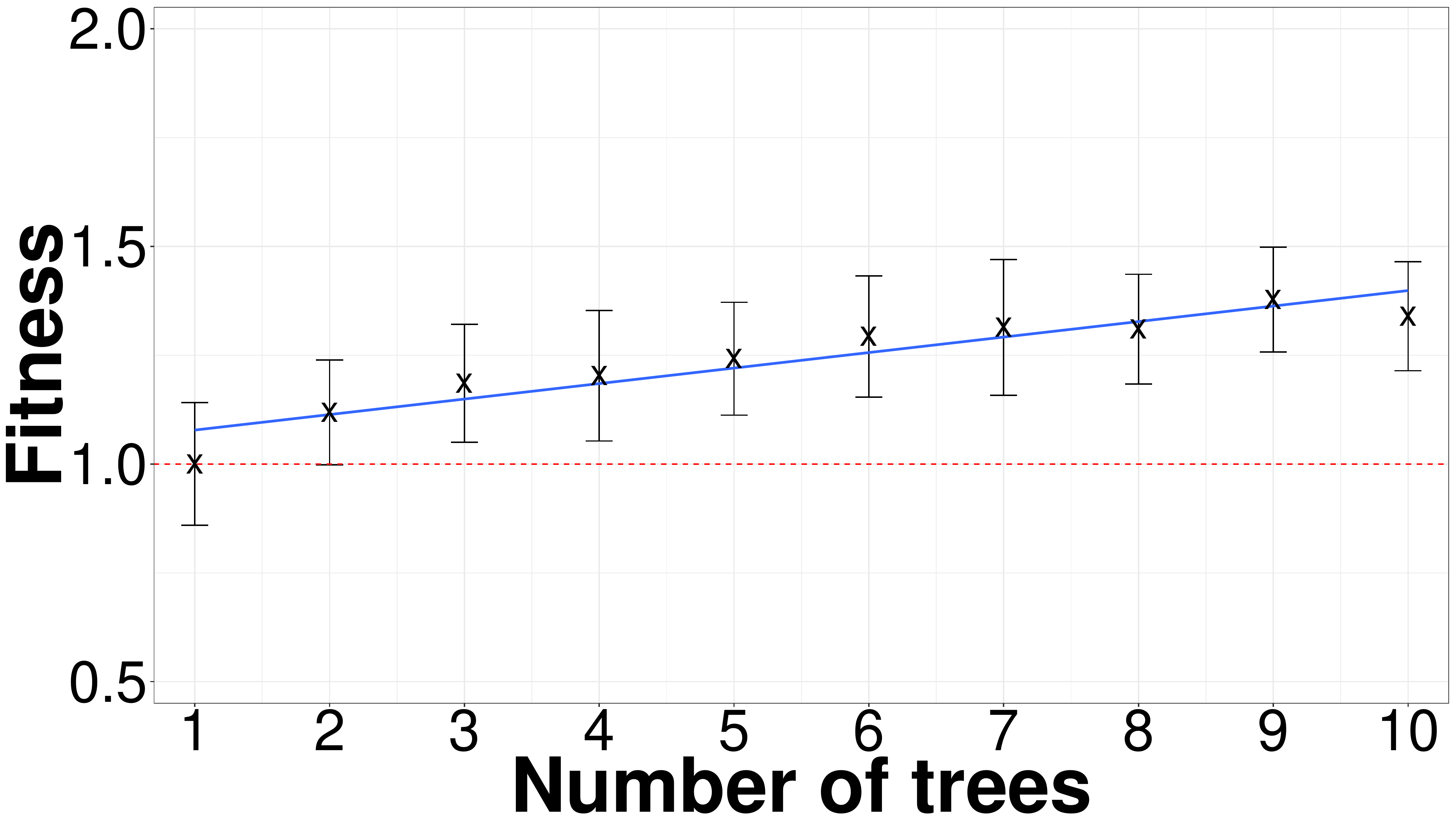}\\ 
	\mtGraph{j}{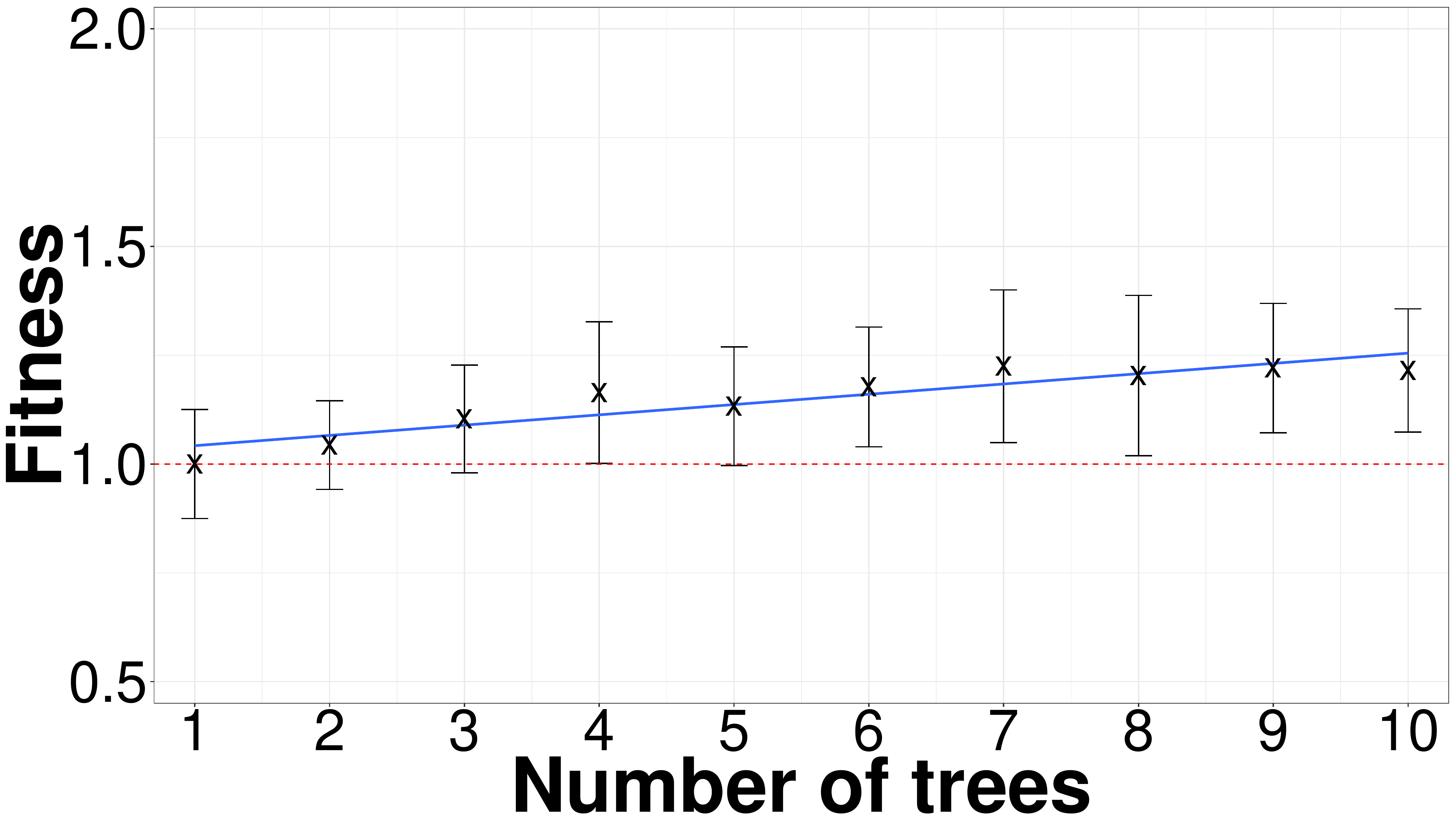}
	\mtGraph{k}{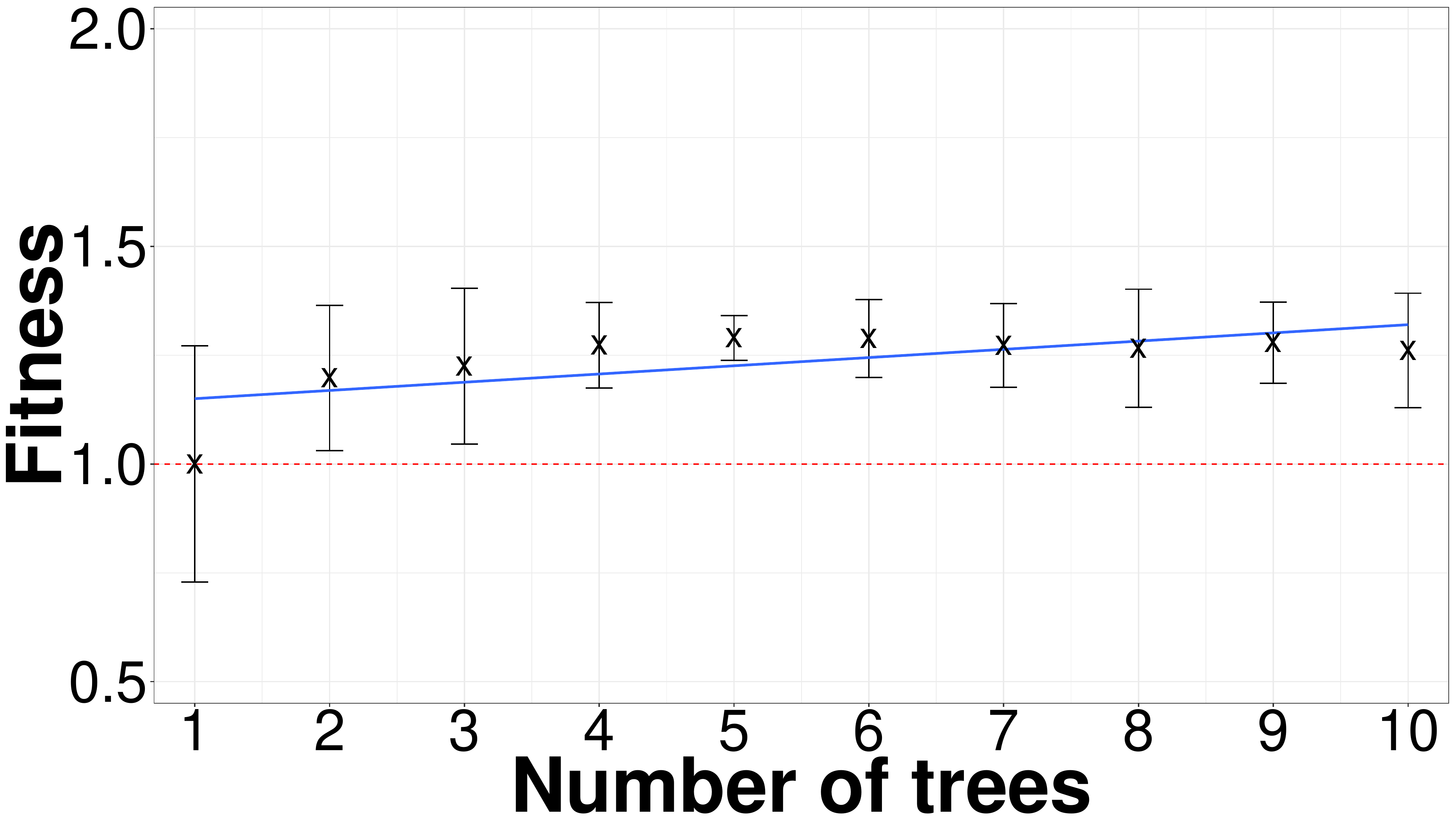}
	\mtGraph{l}{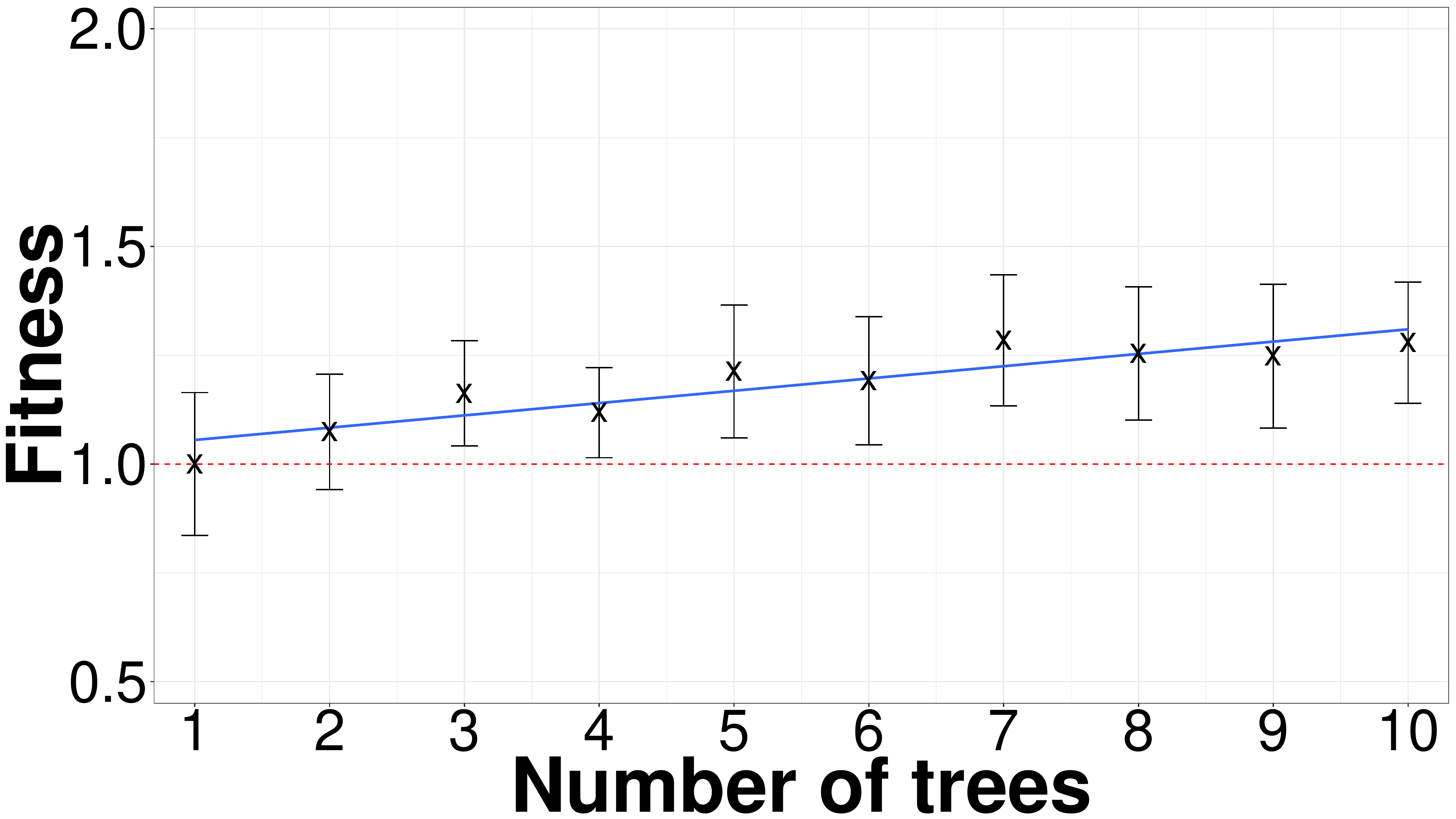}\\ 
	\mtGraph{m}{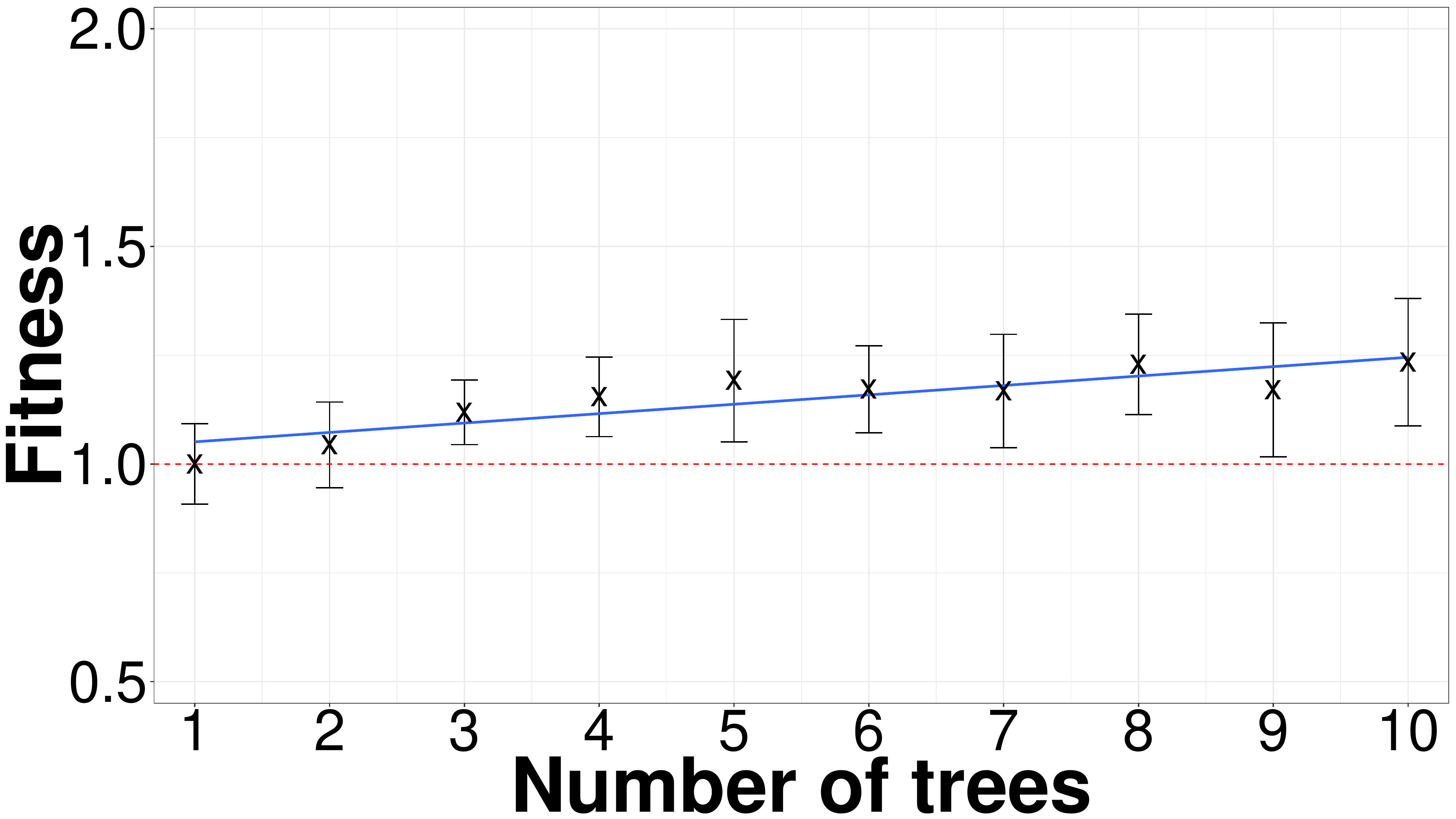}
	\mtGraph{n}{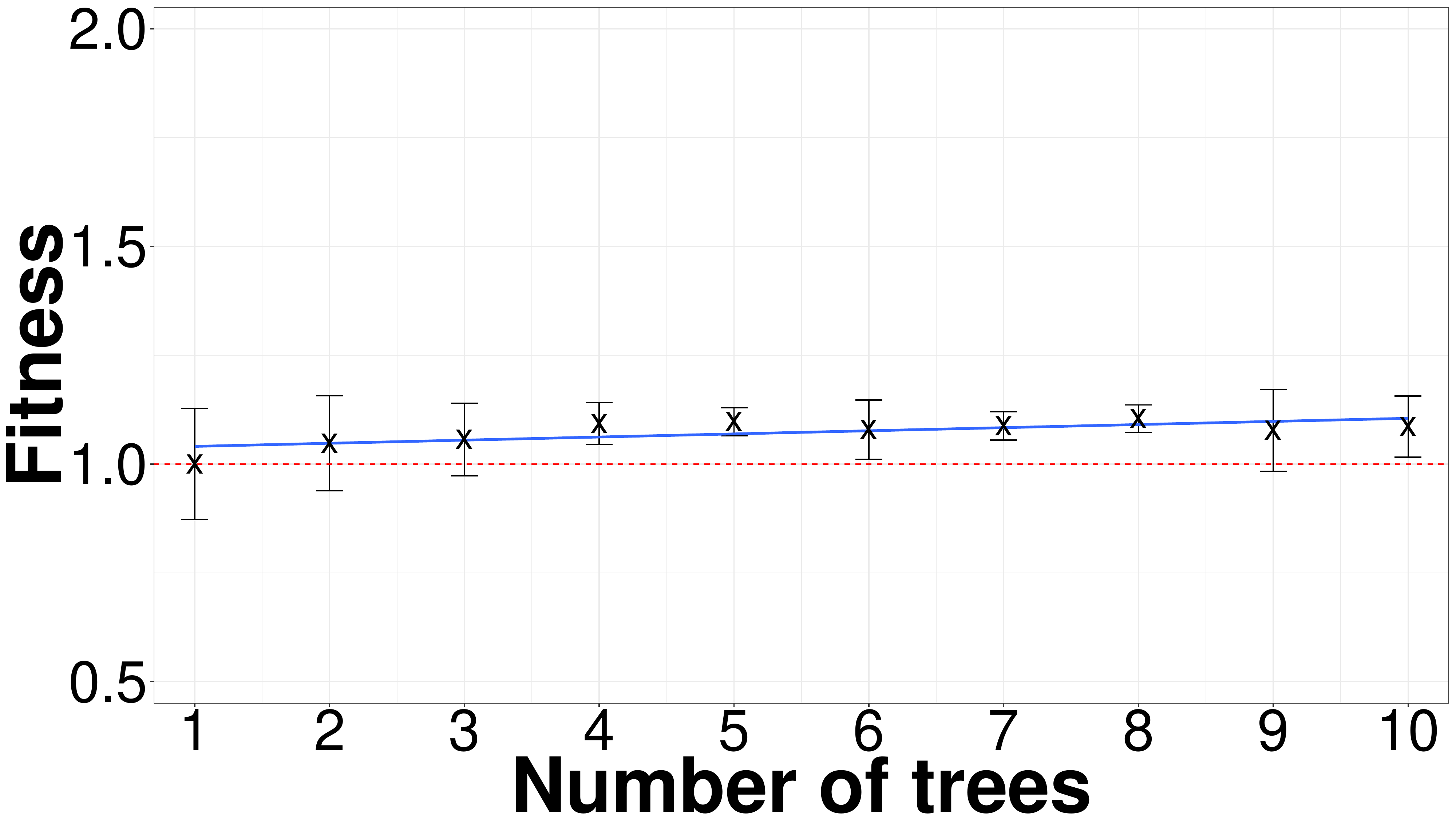}
	\mtGraph{o}{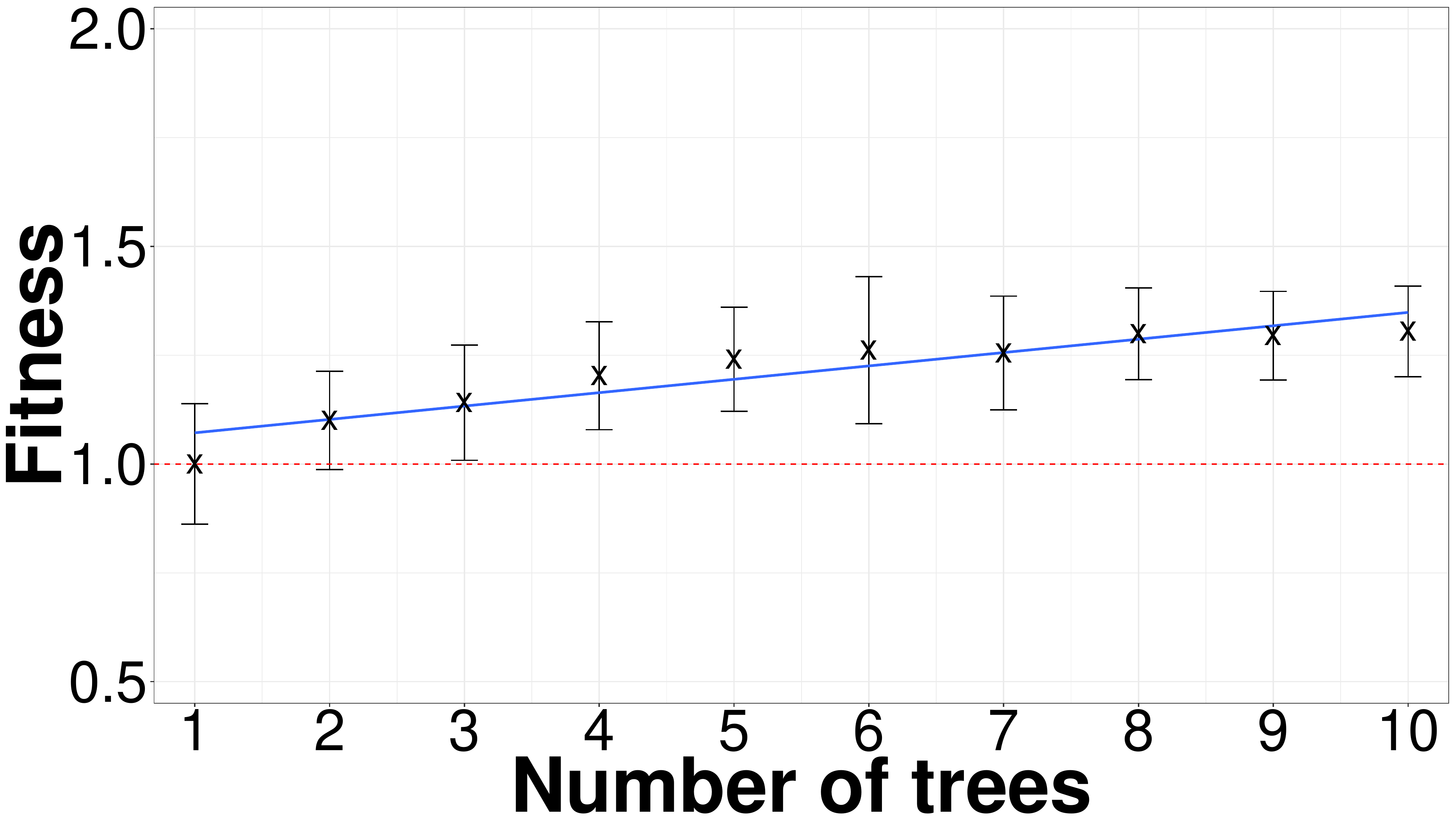}\\ 
	\mtGraph{p}{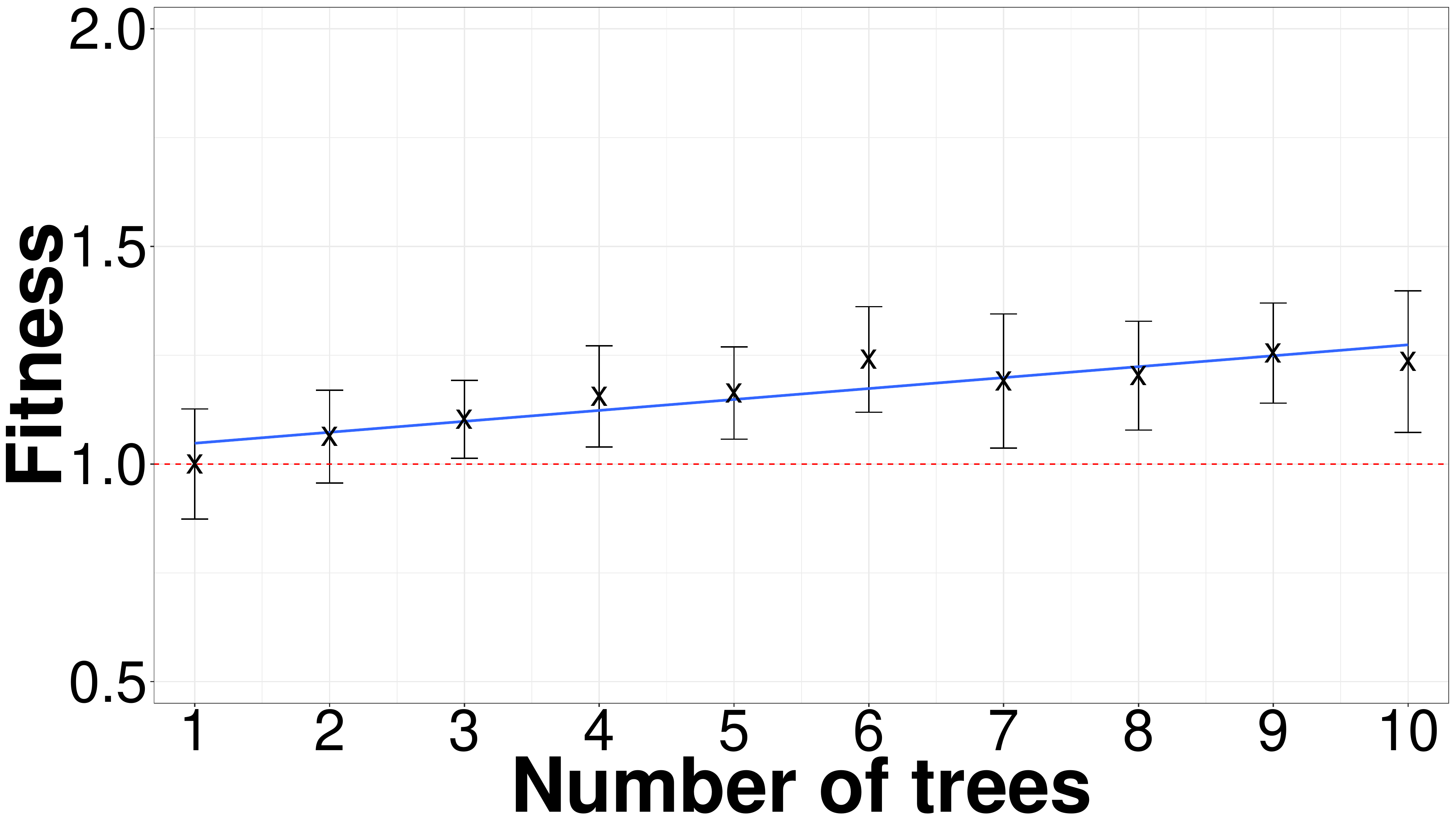}
	\mtGraph{q}{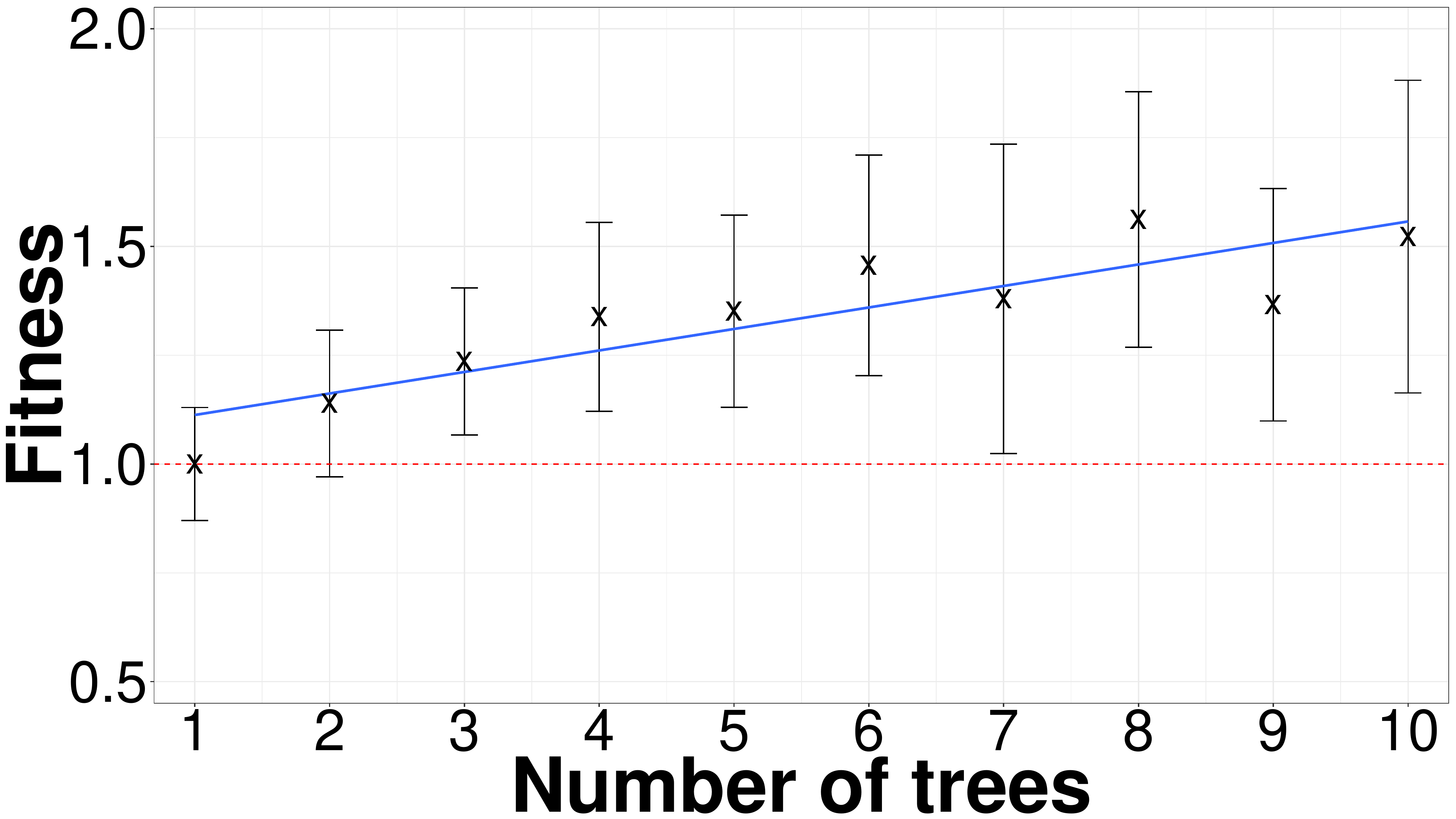}\\
	
	\caption{Effect on training performance as the number of trees is increased.}
	\label{numberOfTreesPlots} 
\end{figure*}

The number of trees to use in a multi-tree approach can be considered to be a form of parameter tuning. To investigate the effect of different numbers of trees on the training performance of the proposed multi-tree approach, we tested a range of trees for $t \in [1,10]$, using the AIC crossover method. We limit $t$ to a maximum of 10, as we found that multi-tree GP did not have improved performance, and trained more slowly at higher $t$ values. Note that the $t=1$ case is not equivalent to the single-tree GPGC method, due to the smaller maximum program depth of $5$. For each dataset, we calculated the mean fitness of 30 runs for each value of $t$. The results are plotted in Fig.\ \ref{numberOfTreesPlots}. Each plot corresponds to a single dataset, with a red dotted line indicating the baseline performance where $t=1$, and each point corresponds to the \textit{relative} fitness for a given value of $t$. The error bars show the standard deviation of each point, for the 30 runs performed for that $t$ value. The blue solid line is a trend-line fitted to the 10 points. 

On the majority of the plots (14 out of 17), increasing the number of trees causes an increase in the fitness obtained; the 10d20cGaussian plot has no noticeable improvement as $t$ is increased, while on the 10d40c and 10d100c plots, the fitness is actually reduced by using more trees. This is consistent with the results presented in Table \ref{table:crossover:ellipsoidResults}, where the AIC method had significantly worse fitness on the 10d40c dataset, and was not significantly different on the 10d100c dataset --- on the 10d100c plot, the fitness value for $t=7$ is very close to the baseline (i.e.\ a relative fitness of 1). For the 10d20cGaussian plot, we hypothesise that there is little room for fitness improvement as $t$ is increased, as shown by the small changes in fitness and ARI performance compared to GPGC in Table \ref{table:crossover:gaussianResults}. On the 14 plots where there was a positive association, the fitness improvement is between 10\% (on 1000d10c), and slightly over 50\% (on 1000d100c), with improvements of around 25\% on the majority of the datasets.

The optimal value of $t$ varies depending on the dataset that is used --- however, fitness tends to peak at a certain value of $t$ for each dataset, before remaining relatively constant or dipping slightly (with the exception of those where $t$ does not improve performance). The best value of $t$ for each dataset is hence the lowest value of $t$ for which performance is significantly better than all lower values of $t$, as this provides the best balance of maximising fitness while maintaining the interpretability and computational benefits of a lower $t$ value.  For most datasets, this value is between $t=5$ and $t=8$, with the exception of 50d20c and 1000d10c, where a $t$ value of 9 and 4 seem to be best, respectively. Hence, we suggest a value of $t=7$ or $t=8$ may be best for an unknown dataset in order to ensure good fitness is obtained.

\section{Further Analysis}

\subsection{Evolved GP Trees}
\label{section:evolvedTrees}
In addition to achieving good clustering performance, our proposed methods are also expected to automatically select a subset of features and construct new, more powerful high-level features due to the tree-based GP structure used. To evaluate the feature manipulation performance of our proposed methods, we analyse an example evolved individual for both the single- and multi-tree approaches in this subsection.

\begin{figure}[!t]
	\vspace{-1em}
	\centering
	\includegraphics[width=.8\textwidth]{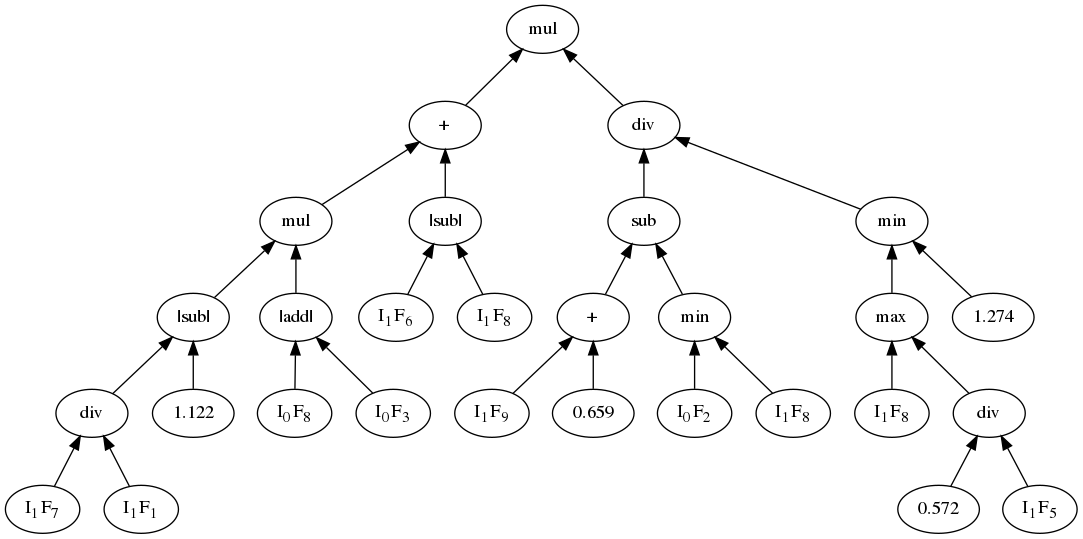}\hfill
	\caption{An evolved individual on the 10d10c dataset using the single-tree approach. Individual has a fitness of 21.37 and ARI of 0.9144 and produces $K=22$ clusters.}
	\label{fig:10d10cExample}
	\vspace{-1em}
\end{figure}

Fig.\ \ref{fig:10d10cExample} shows a single-tree individual evolved by GPGC on the 10d10c dataset, which has a very good ARI result of 0.9144. We can see that the tree produced is able to combine a number of different sub-trees to effectively construct a custom similarity function which can vary its behaviour across the dataset through the use of conditional \textit{max} and \textit{min} operators. A range of \textit{building blocks} are used to find the similarity of two instances, from simple feature weighting operations (e.g.\ $0.572 \div I_1F_5$, $ I_1F_9 + 0.659$) to more advanced feature comparisons (e.g.\ $min(I_0F_2,I_1F_8)$), with high-level features formed by combining these building blocks in a variety of ways. By evolving a similarity function tailored to this dataset, GPGC is able to outperform the benchmark methods, which use the inflexible Euclidean distance function. \revisionOnePar{This evolved function gives a different nearest neighbour to that of Euclidean distance for 97.04\% of the instances, and on average chooses the 5.4th nearest neighbour according to Euclidean distance ordering. Clearly, GPGC has produced a significantly different ordering which is more appropriate for this dataset than normal Euclidean distance ordering.}

\begin{figure}[t]

	\setlength{\hMp}{1.3em}
	\vspace{-2em}
	\centering
	\subfloat[]{\includegraphics[height=1.45\hMp]{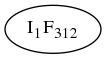}}\hfill
	\subfloat[]{\includegraphics[height=4\hMp]{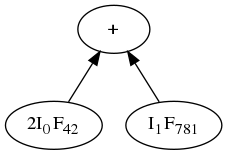}}\hfill
	\subfloat[]{\includegraphics[height=10.5\hMp]{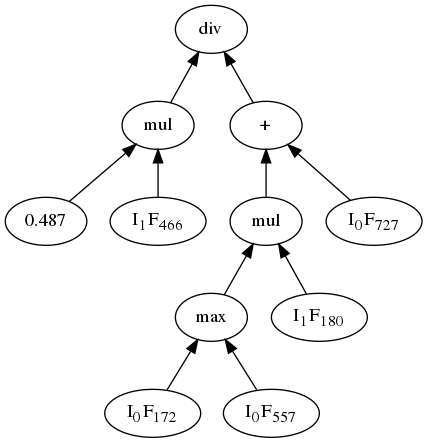}}\hfill
	\subfloat[]{\includegraphics[height=4\hMp]{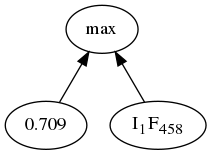}}\hfill
	\subfloat[]{\includegraphics[height=8.5\hMp]{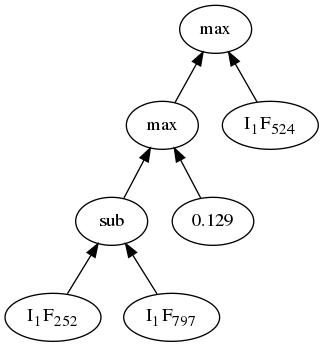}}\hfill	
	\subfloat[]{\includegraphics[height=6.5\hMp]{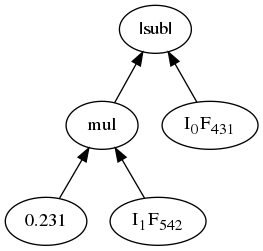}}\hfill
	\subfloat[]{\includegraphics[height=6.5\hMp]{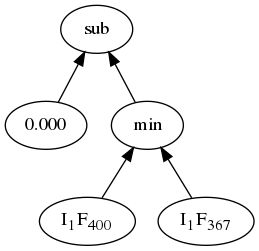}}\hfill
	\caption{An evolved individual on the 1000d20c dataset using AIC crossover and $t=7$. Individual has a fitness of 10.43 and ARI of 0.9616 and produces $K=22$ clusters.}
	\label{fig:1000d20cExample}
	\vspace{-1em}
\end{figure}

Fig.\ \ref{fig:1000d20cExample} shows an example of a multi-tree individual evolved on the 1000d20c dataset, which has a very good ARI result of 0.9616. We have simplified the trees where appropriate to aid interpretability by computing constants and removing dead branches. The seven evolved trees are generally quite simple, with only one tree (c) being the maximum depth of five, and one, two, two, and one trees having depths of four, three, two, and one, respectively. While it is difficult to understand why these trees perform well across each instance in the dataset, it is possible to gain insight by examining the general behaviour of each tree ((a)--(g)), as shown below:

\begin{enumerate}
\itemsep 1pt
	\item[(a)] simple feature selection of $I_{1}F_{312}$.
	\item[(b)] computing a weighted sum of two selected features.
	\item[(c)] constructing a more powerful high-level feature by weighting and constructing non-linear combinations of five original features.
	\item[(d)] thresholding $I_{1}F_{458}$ so that it has a minor impact on the total similarity.
	\item[(e)] finding the maximum of: a feature, a constant value, and the difference between two features. This gives varied behaviour based on the instances being considered.
	\item[(f)] finding the absolute difference between two features, with one feature scaled.
	\item[(g)] finds the maximum of two features, and then takes the result as a negative. In this way, the bigger the result, the less similar the two instances are said to be.
	
\end{enumerate}

Each of the seven trees evaluated above had distinctive and interesting behaviour, which gives insight into which features are useful in the dataset, and into what relationships between features can be used to gauge instances' similarities accurately. In contrast, a standard distance function cannot provide such insight, as it uses the \textit{full} feature set and performs only linear comparisons between instances' features. Of the 1000 features in the 1000d20c dataset, the example individual in Fig.\ \ref{fig:1000d20cExample} uses only 15 features to build its seven similarity functions. This means the clustering partition produced is much more interpretable than one produced by a standard nearest-neighbour graph-based clustering algorithm. \revisionOnePar{This evolved meta-similarity function chooses the same nearest neighbour as that of Euclidean distance on only 2.67\% of the instances in the dataset. On average, the 6.2th nearest neighbour is chosen as the first nearest neighbour: this is a similar trend to that of the previous example in that the neighbours have been significantly re-ordered to be tailored for this dataset.}

\subsection{Visualising the Clusters Found}
To further analyse the examples discussed in Section \ref{section:evolvedTrees}, we visualise the clusters produced compared to the ground truth clustering in this subsection using the commonly used t-{SNE} visualisation method \citep{maatenTSNE}, which minimises the probability distribution divergence between the two-dimensional visualisation and the original feature space.

\begin{figure}[!t]
	\vspace{-1em}
	\centering	
	\subfloat[GPGC (F-M: 0.9144)]{\includegraphics[width=.4\textwidth]{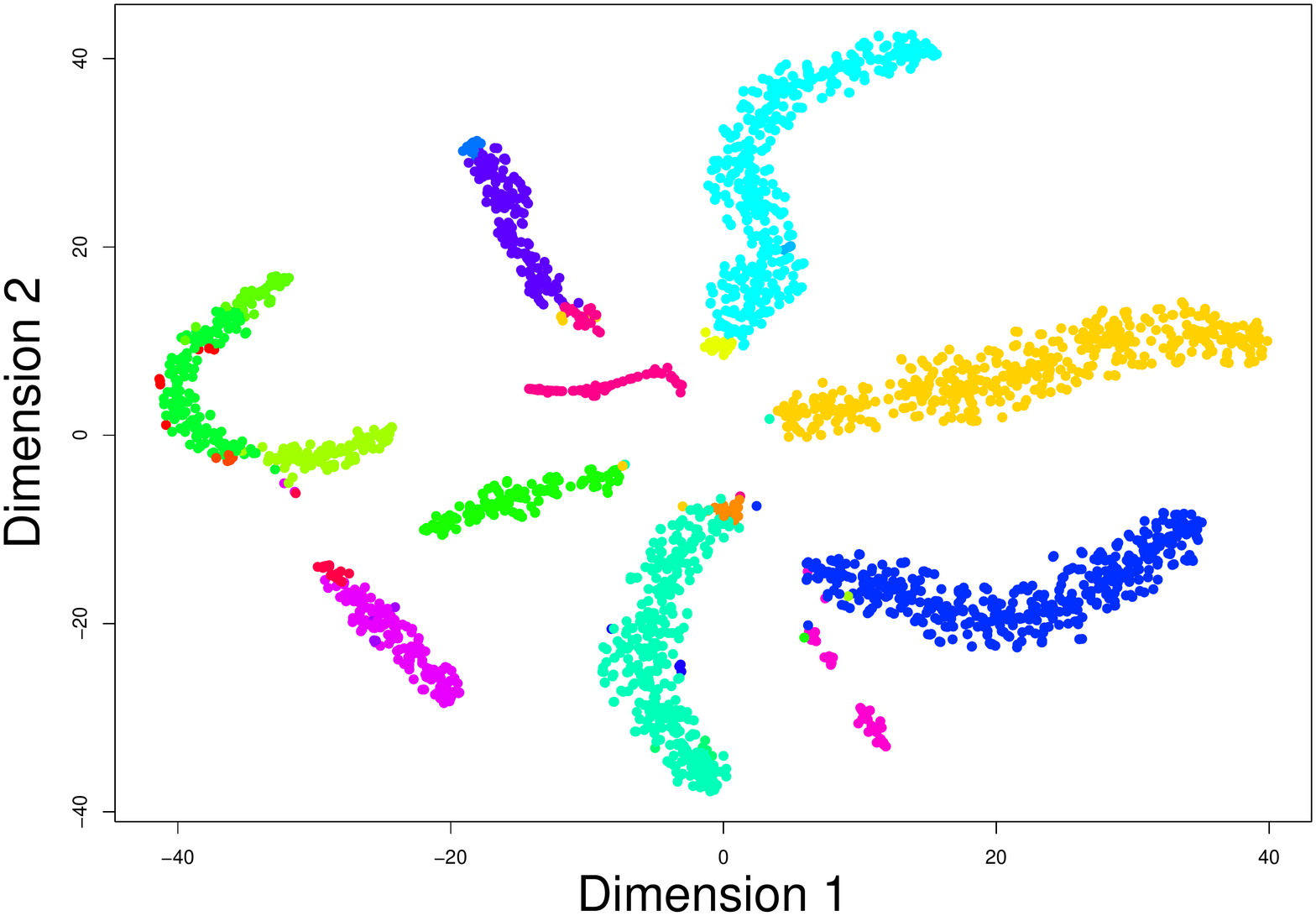}}\hfill
	\subfloat[Ground Truth]{\includegraphics[width=.4\textwidth]{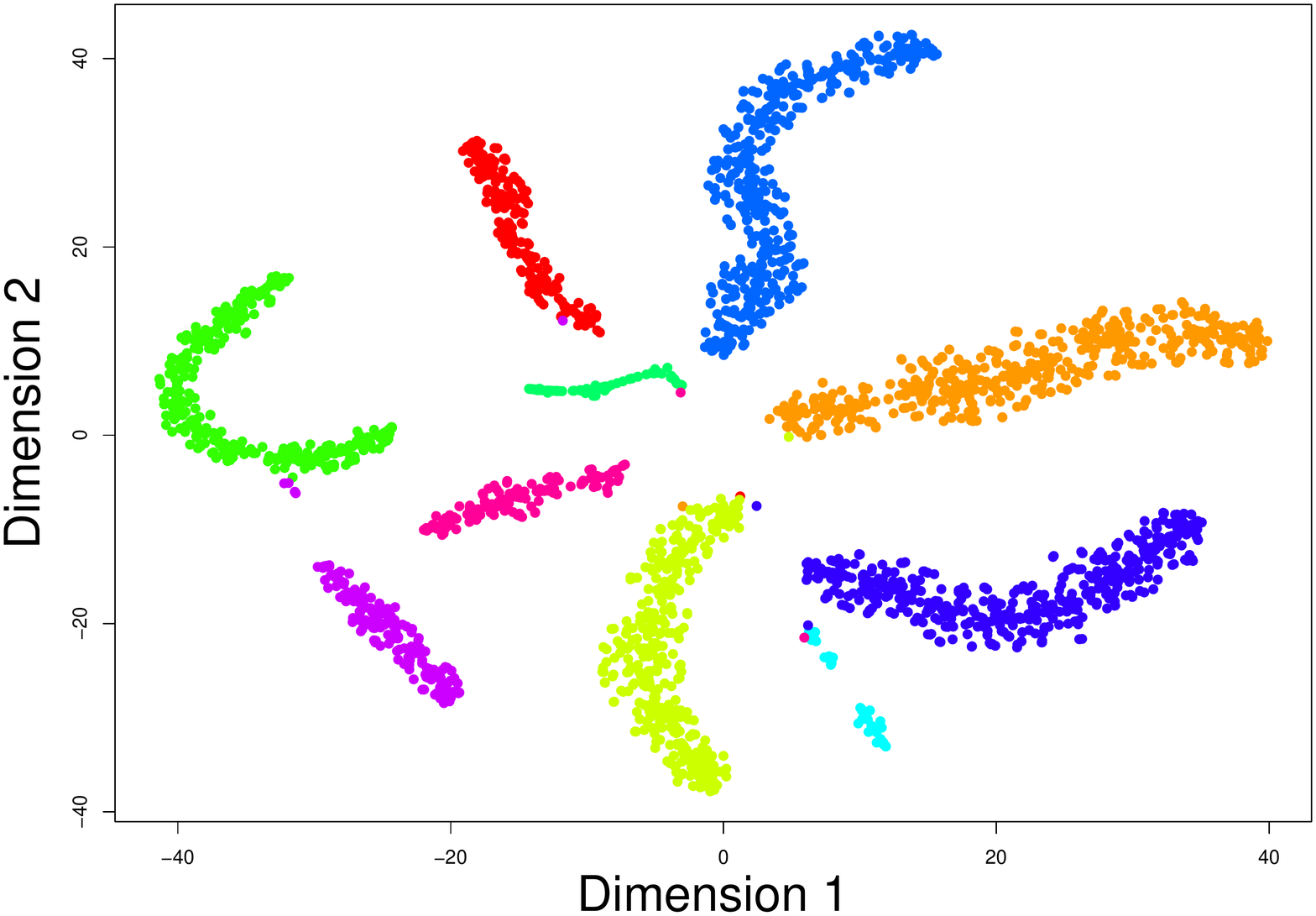}}\hfill \\
		\vspace{-.5em}
	\subfloat[$k$-means++ (F-M: 0.6134)]{\includegraphics[width=.4\textwidth]{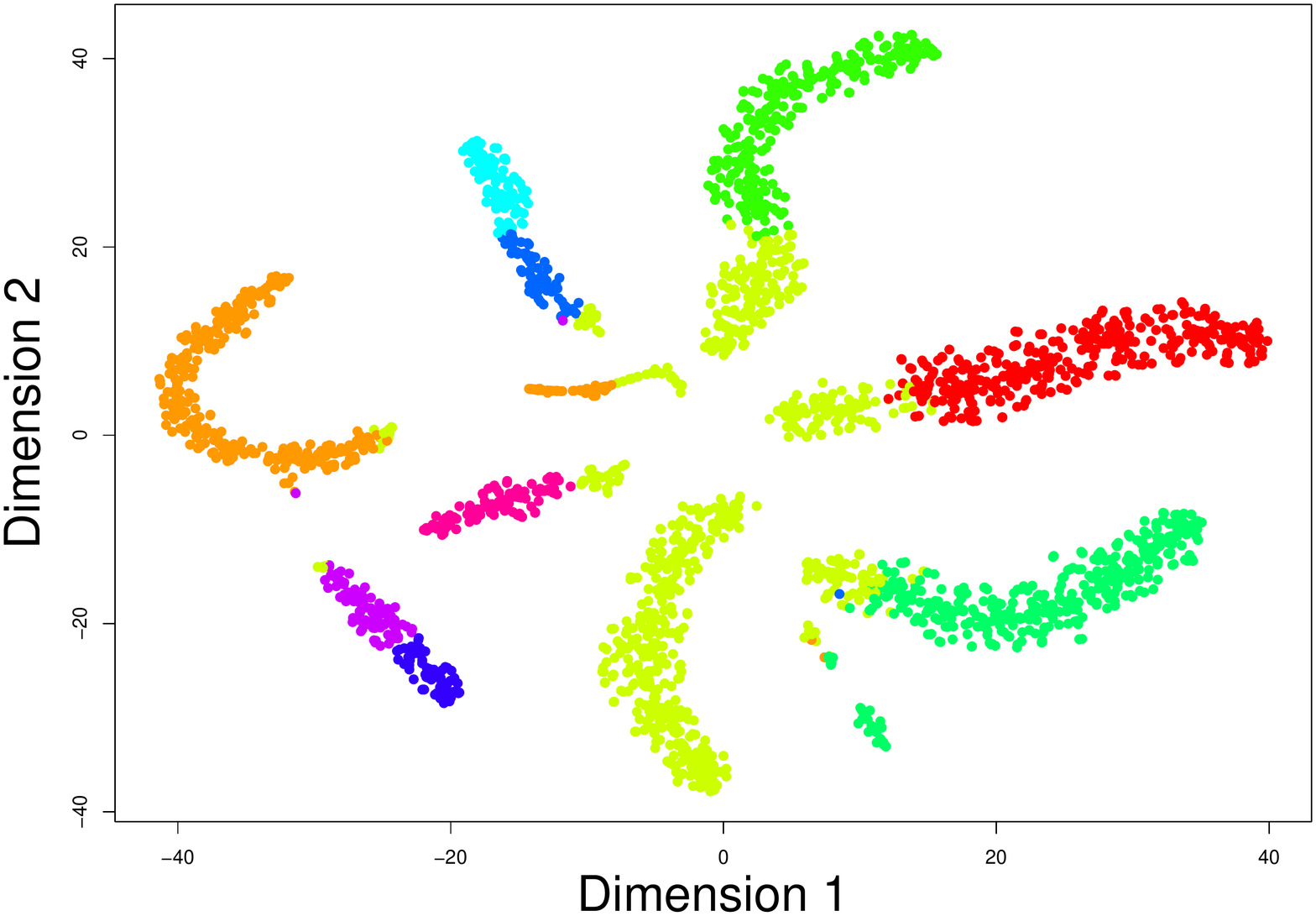}}\hfill
	\subfloat[NG-2NN (F-M: 0.5103)]{\includegraphics[width=.4\textwidth]{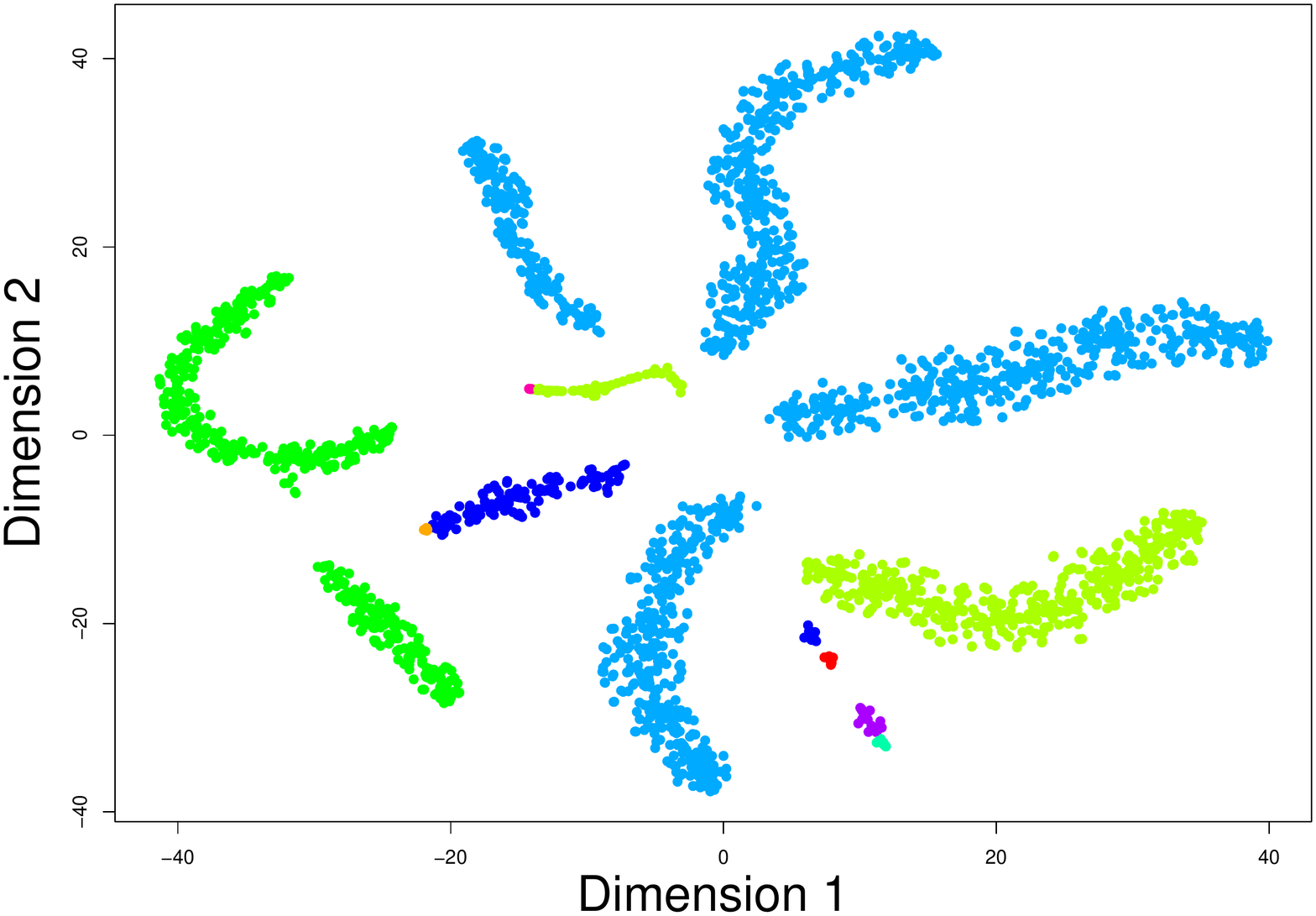}}\hfill \\
			\vspace{-.5em}

	\caption{Visualising the partitions chosen by a GP individual compared to \revisionTwo{a sample of the} baseline methods on the 10d10c dataset. t-SNE \citep{maatenTSNE} is used to reduce dimensionality to two dimensions. Each colour corresponds to a single cluster. }
	\label{fig:10d10cVisualise}
	\vspace{-1em}
\end{figure}

Fig.\ \ref{fig:10d10cVisualise} shows the clusters produced by GPGC, $k$-means++ and the NG-2NN methods on the 10d10c dataset. For GPGC, we use the same evolved individual as in Section \ref{section:evolvedTrees}. For $k$-means++, we chose the result with the highest ARI of the 30 runs. NG-2NN is deterministic, and so the single result is shown. In addition, the ground truth is shown in Fig.\ \ref{fig:10d10cVisualise} for reference. It is clear that GPGC is able to most accurately reproduce the ground truth, with the majority of the clusters mapping to the ground truth well, asides from a few instances in each cluster. The exception is on the horseshoe-shaped cluster on the left of the visualisation, where GPGC has over-clustered the data by splitting this cluster in two. The $k$-means++ method clearly performs very poorly, with only the horseshoe-shaped cluster being clustered nearly correctly; all other clusters have significant overlap. The NG-2NN method also produces clearly incorrect clusters, with many clusters being combined, including four distinct clusters combined into one single blue cluster. GPGC is clearly able to better find the natural clusters compared to these baseline methods.

\begin{figure}[!t]
	\vspace{-1em}
	\centering	
	\subfloat[All features: GPGC-MT]{\includegraphics[width=.4\textwidth]{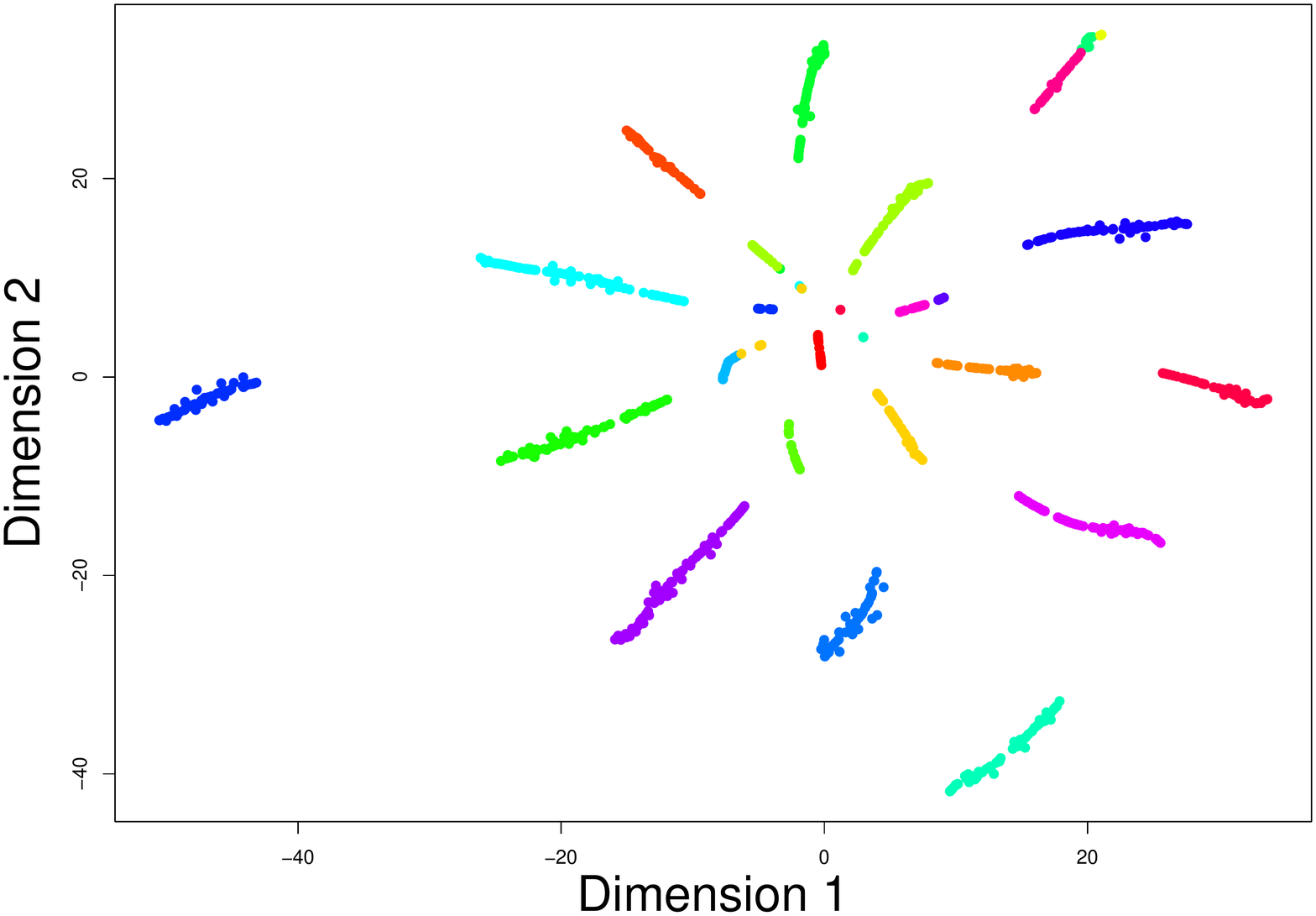}}\hfill
	\subfloat[All features: Ground Truth]{\includegraphics[width=.4\textwidth]{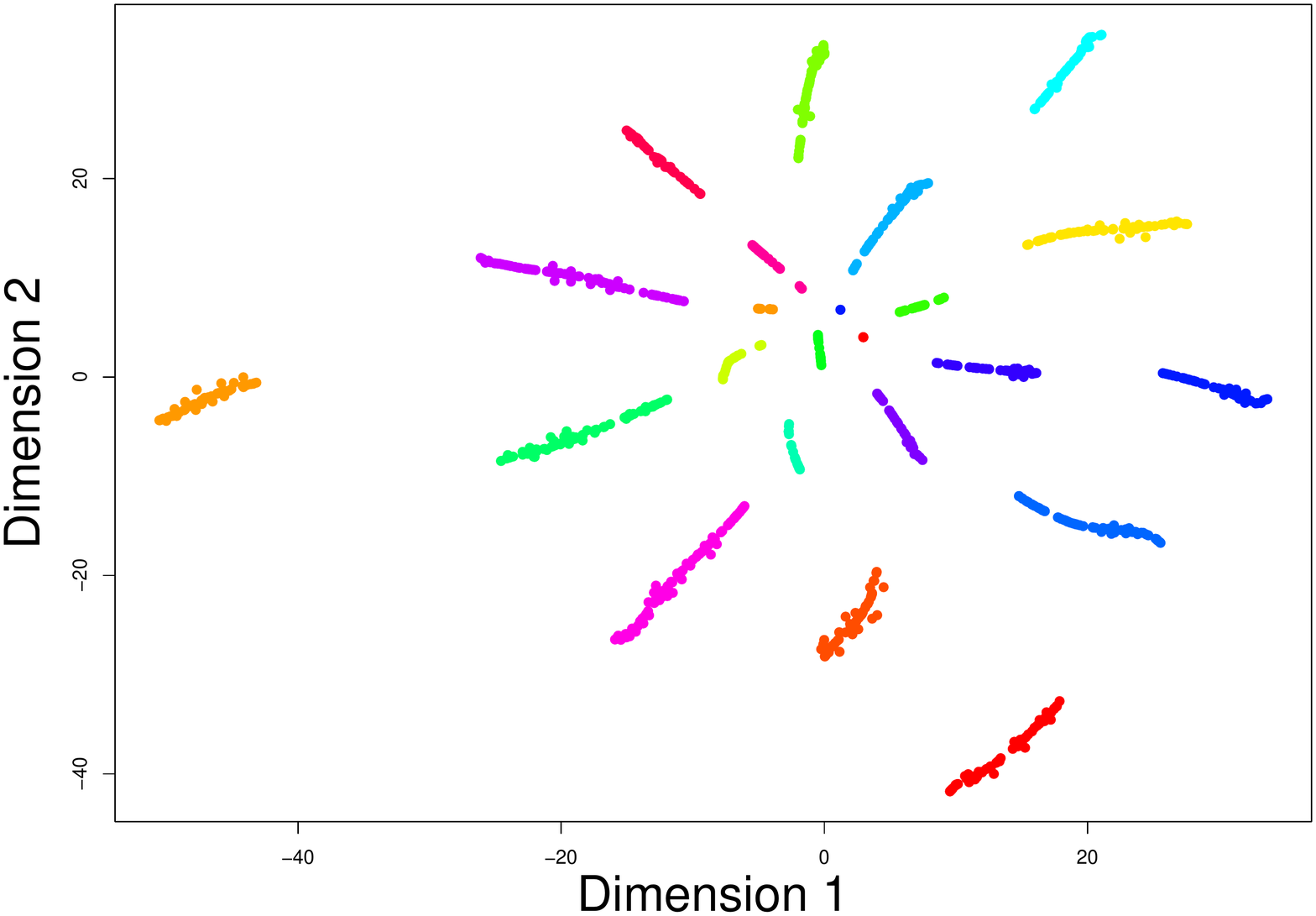}}\hfill \\
	\vspace{-.5em}
	\subfloat[Selected features: GPGC-MT]{\includegraphics[width=.4\textwidth]{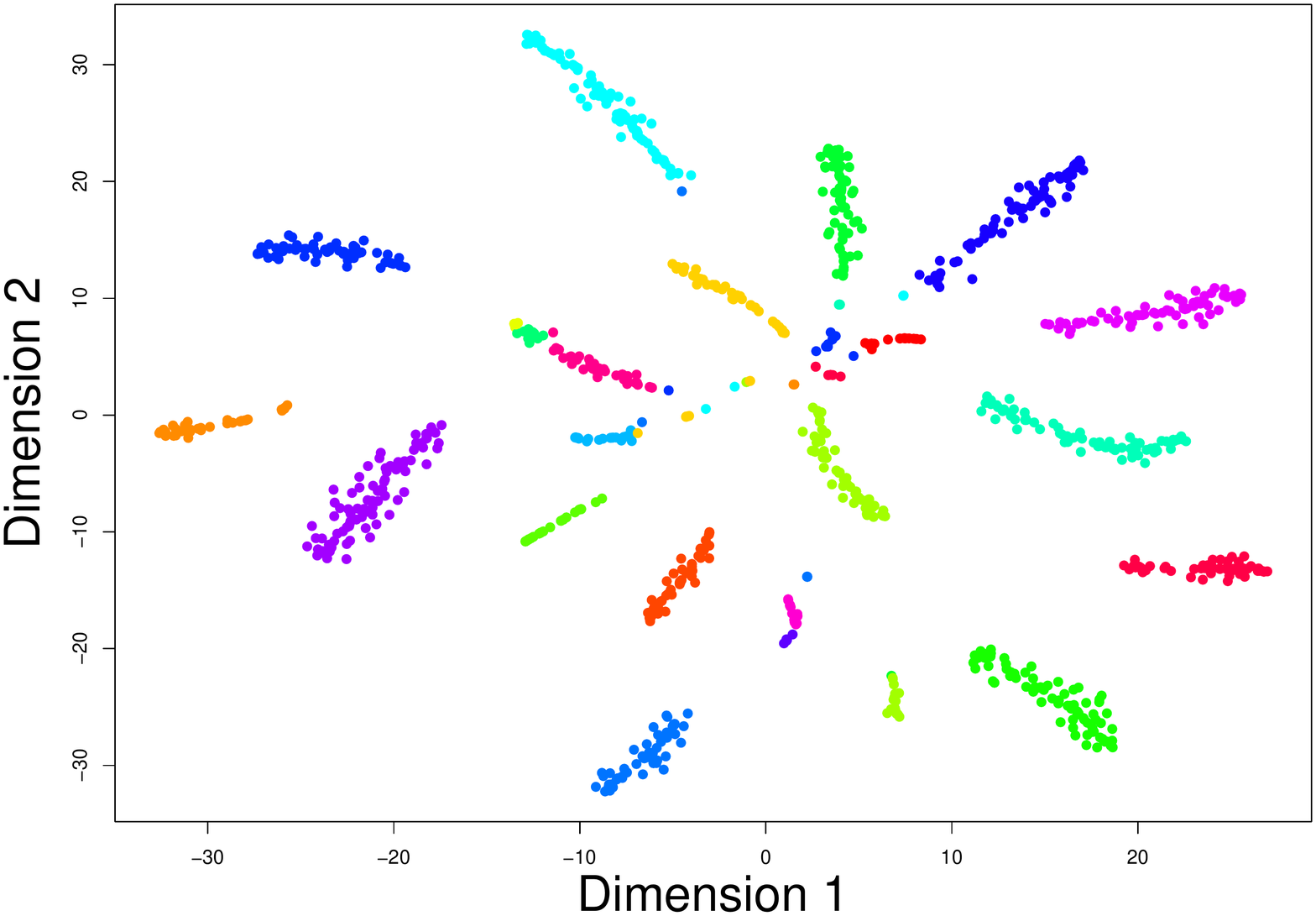}}\hfill
	\subfloat[Selected features: Ground Truth]{\includegraphics[width=.4\textwidth]{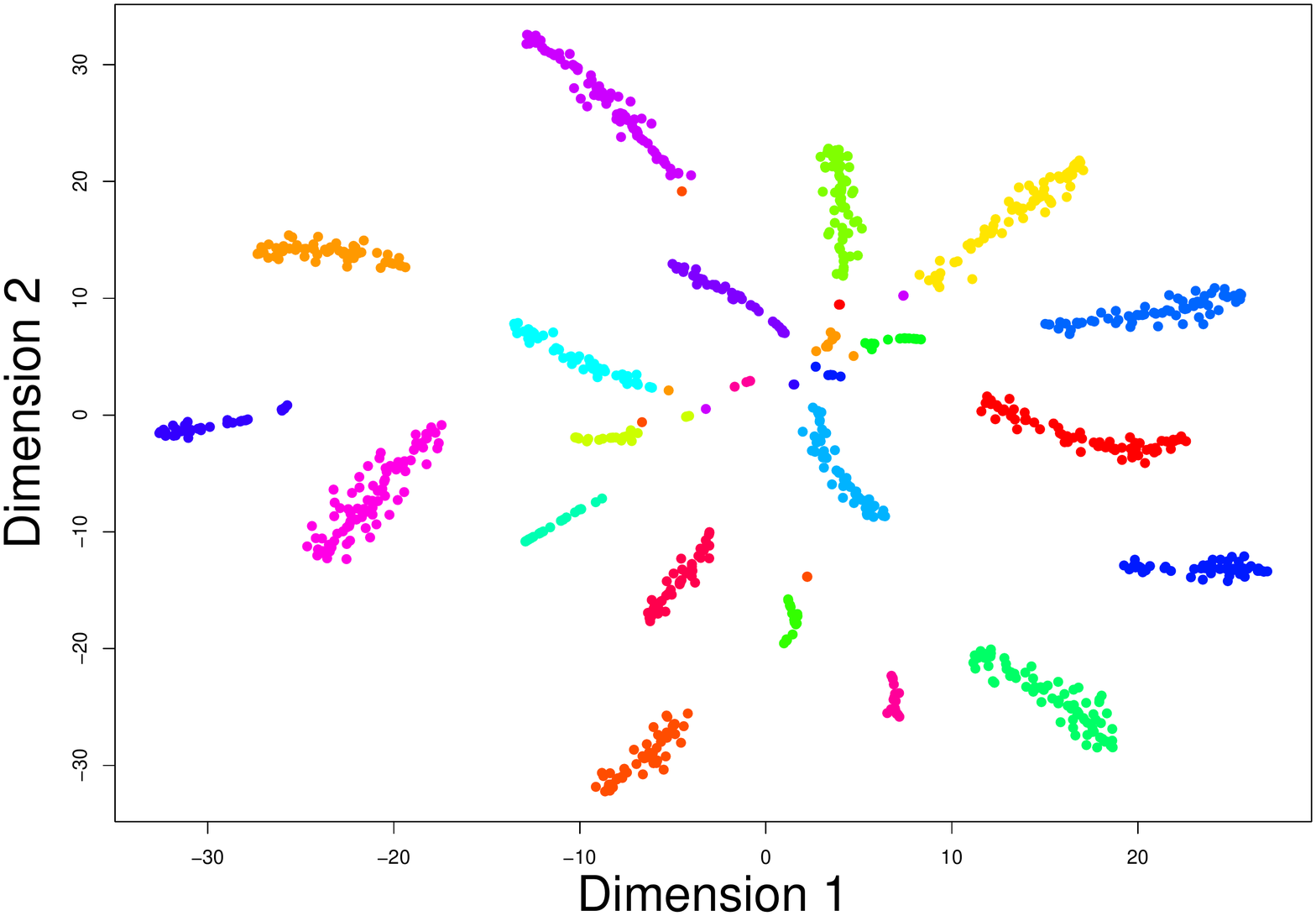}}\hfill \\
	\vspace{-.5em}
	
	\caption{Visualising the partitions chosen by a GP individual on the 1000d20c dataset. t-SNE \citep{maatenTSNE} is used to reduce dimensionality to two dimensions. Figures (a) and (b) show the visualisations formed when all features are used by t-SNE, whereas (c) and (d) show the visualisations using only the features used in the GP tree. Each colour corresponds to a single cluster.}
	\label{fig:1000d20cVisualise}
	\vspace{-1em}
\end{figure}

Fig.\ \ref{fig:1000d20cVisualise} (a) and (b) show the clusters produced by the GPGC-MT method (using the evolved tree discussed in Section \ref{section:evolvedTrees}) and the ground truth respectively on the 1000d20c dataset. GPGC-MT reproduces the ground truth accurately, with only a small amount of over-clustering. The GPGC-MT method uses only a subset of features in the evolved trees; in this case, only 15 of the 1000 features are used. To analyse whether using so few features would reduce the interpretability of the clusters produced, we performed another set of visualisations which used only the 15 selected features as input, as shown in Fig.\ \ref{fig:1000d20cVisualise} (c) and (d). The clusters shown in these visualisations are still very distinct and well-separated, which suggests that the GPGC-MT method was able to successfully perform feature selection implicitly in the evolved similarity functions. While t-{SNE} is able to reliably project the feature space into two dimensions, it does so at the cost of interpretability -- the two dimensions produced cannot be easily mapped back to the original feature set, and so it is very difficult to analyse why a cluster contains certain instances. In contrast, GPGC-MT uses only a small subset of the feature set, and explicitly combines features in an interpretable manner in the evolved trees.

\subsection{Evolutionary Process}
\begin{figure*}[!t]
	\centering
	\null\hfill
	\subfloat[10d10cGaussian]{\includegraphics[width=.49\textwidth]{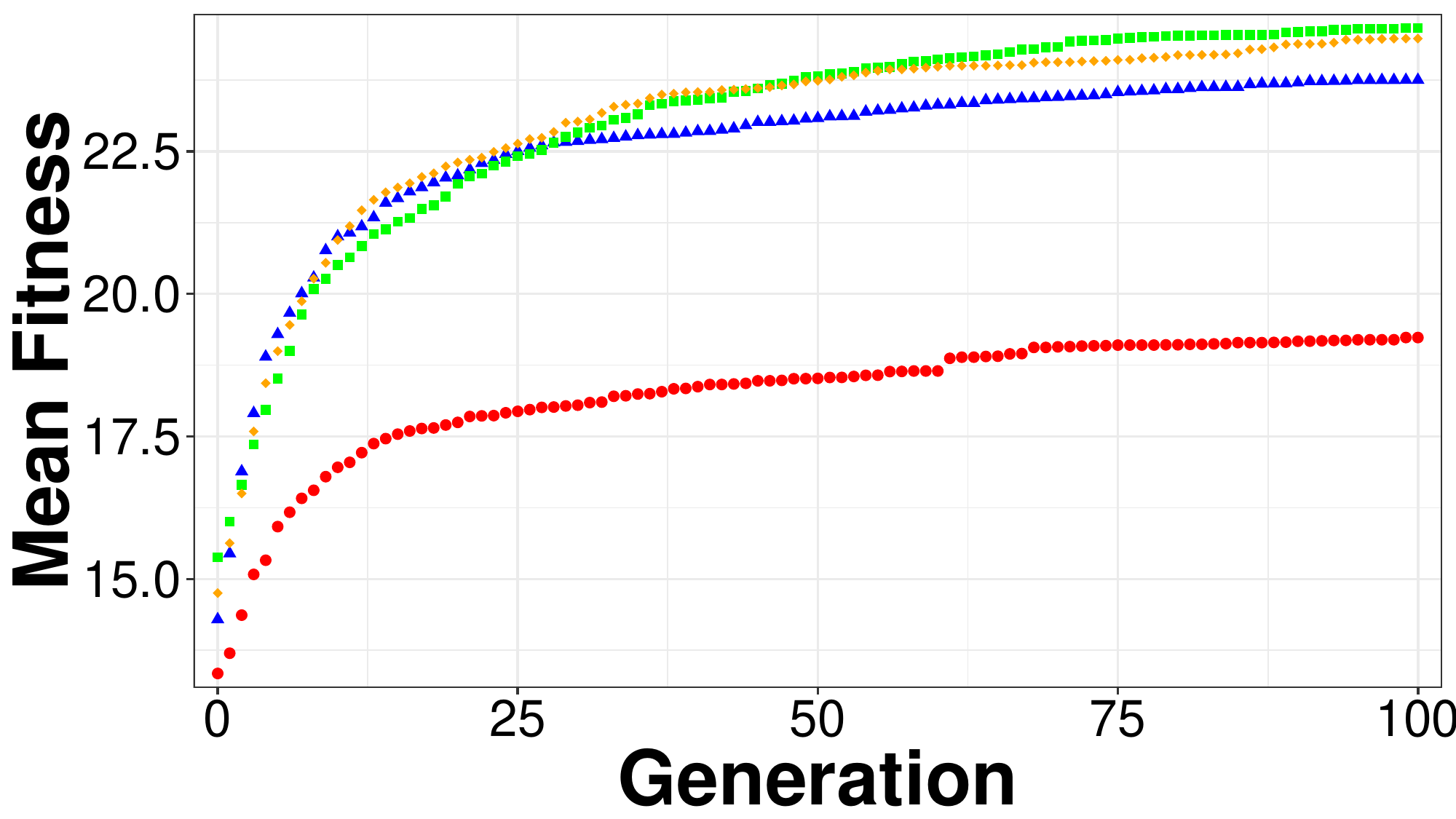}}
	\hfill
	\subfloat[1000d100c]{\includegraphics[width=.49\textwidth]{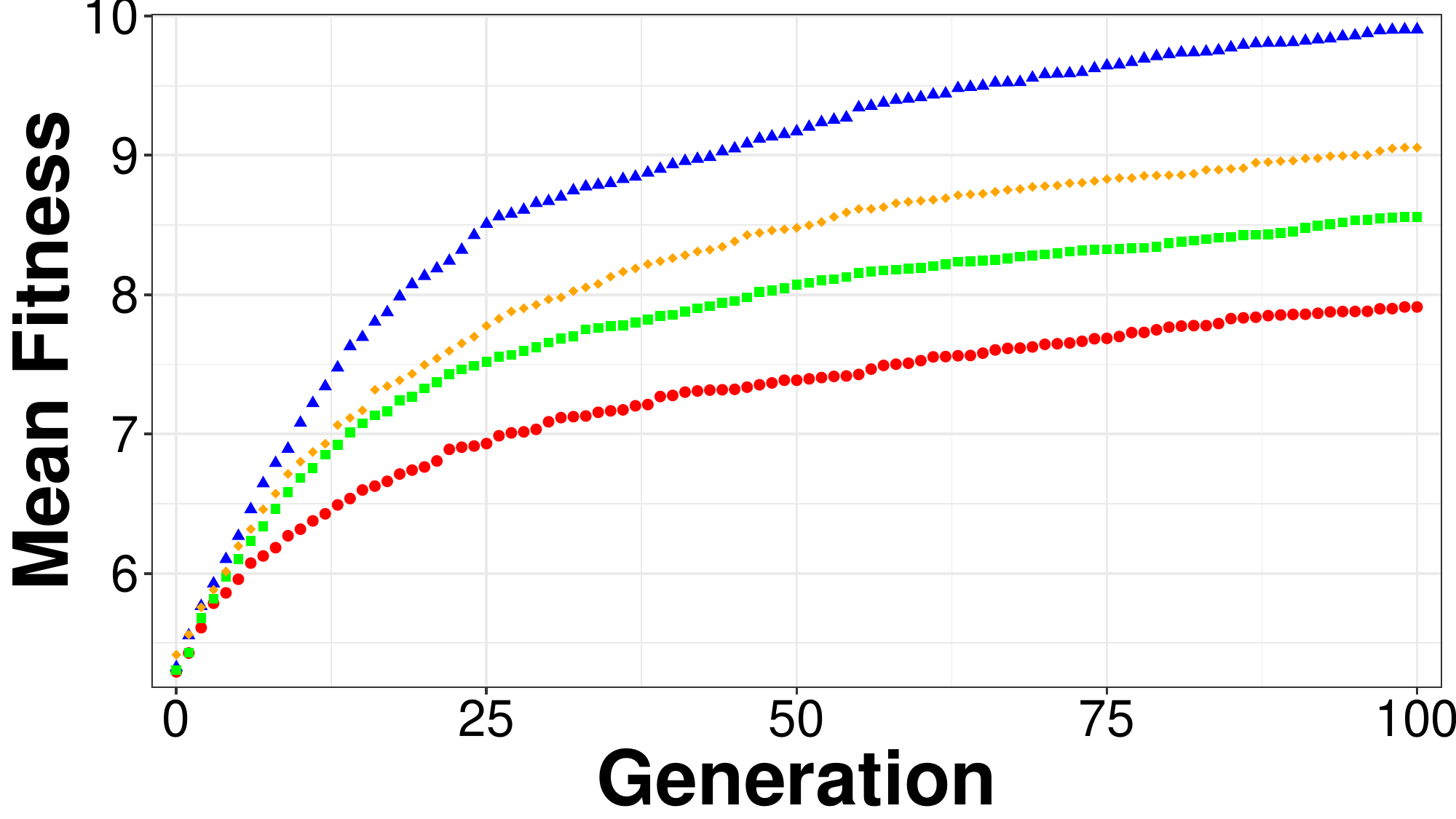}} 
	\hfill\null
	
	\caption{\revisionOne{Fitness of the two proposed methods over the evolutionary process. The red dots correspond to GPGC, the blue triangles to GPGC-AIC, the green squares to GPGC-SIC, and the orange diamonds to GPGC-RIC respectively.}}
	\label{fig:evolutionary}
	\vspace{-1em}
\end{figure*}

To further analyse the learning effectiveness of \revisionOnePar{the proposed methods (GPGC and the three multi-tree crossover approaches)}, we plot the fitness over the evolutionary process for the 10d10cGaussian and 1000d100c datasets, as shown in Fig\ \ref{fig:evolutionary} (a) and (b) respectively. For each dataset, we plot the mean fitness of the best individual at each generation, taken across the 30 independent runs. These datasets were selected as they represent the datasets with the lowest and highest $m$ and $K$ values, and are from each of the two different generators used. \revisionOne{Both datasets show the same pattern for the single-tree compared with the multi-tree approaches: while all methods begin at similar fitnesses, the multi-tree methods increase in mean fitness at a significantly faster rate, and reaches a much higher mean fitness overall. Indeed, the final fitness of GPGC at the 100th generation is achieved by each of the multi-tree approaches by generation 25 in both datasets. It is clear that the multi-tree approaches can train more efficiently (i.e.\ with a steeper initial slope), and effectively, by reaching a higher final fitness over the same number of generations. While the GPGC-AIC method is slightly outperformed by the other two crossover approaches on the 10d10cGaussian dataset, it is clearly the best method on the more difficult 1000d100c dataset, which reinforces our view that this is the best of the proposed approaches.} While fitness appears to have levelled off by generation 100 on the 10d10cGaussian dataset, it appears that additional generations could improve the performance on the 1000d100c dataset even further.

\section{Conclusions and Future Work}
In this work, we proposed a novel approach to performing clustering, whereby GP was used to automatically evolve similarity functions in place of the commonly used inflexible distance metrics. The results of our experiments showed that the automatically generated similarity functions could improve the performance and consistency of clustering algorithms using a graph representation, while producing more interpretable similarity metrics, which \revisionTwo{have the potential to be understood by a domain expert as they select only the most important features in a dataset.} We also showed that a multi-tree GP approach could be utilised to further improve the performance by automatically evolving several highly-specific similarity functions, which are able to specialise on different components of the overall clustering problem. 

\revisionOnePar{While the investigation in this paper was focused on graph-based clustering, due to its ability to model a range of cluster shapes, we also hope our proposed approaches can be applied to nearly any clustering method which uses a similarity function to perform clustering. By replacing the graph-based clustering approach with another given clustering algorithm, the evolved similarity functions will need to be optimised to work with that algorithm instead.} We would also like to test our proposed approaches on real-world data which has ``gold-standard'' labels, but all real-world datasets we have found provide \textbf{class} labels only, which are not suitable for measuring cluster quality. \revisionOnePar{This paper focused on using a scalar fitness function so as to constrain the scope of this work, allowing us to directly evaluate the quality of the proposed GP representation.}  In future work, we would like to extend our proposed fitness function by using an evolutionary multi-objective optimisation (EMO) approach --- the three key measures of cluster quality (compactness, separability, connectedness) partially conflict with each other, and so using an EMO approach may allow better and more varied solutions to be generated. \revisionOnePar{Initial experiments (see Appendix A) showed that GPGC had promise for subspace clustering, but that better performance could likely be achieved in the future by developing a new fitness function and designing new genetic operators to be specific to subspace clustering tasks.} There is also scope for refining the GP program design used: the terminals and functions could be further tailored to the clustering domain through the use of other feature comparison operators.

\small

\setlength{\bibsep}{2.5pt}
\bibliographystyle{apalike}
\bibliography{gpgcECJ}

\begin{thebibliography}{}

\bibitem[Aggarwal and Reddy, 2014]{aggarwal2014data}
Aggarwal, C.~C. and Reddy, C.~K., editors (2014).
\newblock {\em Data Clustering: Algorithms and Applications}.
\newblock {CRC} Press.

\bibitem[Ahn et~al., 2011]{ahn2011genetic}
Ahn, C.~W., Oh, S., and Oh, M. (2011).
\newblock A genetic programming approach to data clustering.
\newblock In {\em Proceedings of the International Conference on Multimedia,
  Computer Graphics and Broadcasting (MulGraB), Part {II}}, pages 123--132.

\bibitem[Alelyani et~al., 2013]{alelyani2013feature}
Alelyani, S., Tang, J., and Liu, H. (2013).
\newblock Feature selection for clustering: {A} review.
\newblock In {\em Data Clustering: Algorithms and Applications}, pages 29--60.
  {CRC} Press.

\bibitem[Ankerst et~al., 1999]{ankerst1999optics}
Ankerst, M., Breunig, M.~M., Kriegel, H., and Sander, J. (1999).
\newblock {OPTICS:} ordering points to identify the clustering structure.
\newblock In {\em Proceedings of the International Conference on Management of
  Data}, pages 49--60.

\bibitem[Arthur and Vassilvitskii, 2007]{arthur2007kmeansplusplus}
Arthur, D. and Vassilvitskii, S. (2007).
\newblock k-means++: the advantages of careful seeding.
\newblock In {\em Proceedings of the Eighteenth Annual {ACM-SIAM} Symposium on
  Discrete Algorithms (SODA)}, pages 1027--1035.

\bibitem[Boric and Est{\'{e}}vez, 2007]{boric2007genetic}
Boric, N. and Est{\'{e}}vez, P.~A. (2007).
\newblock Genetic programming-based clustering using an information theoretic
  fitness measure.
\newblock In {\em Proceedings of the {IEEE} Congress on Evolutionary
  Computation (CEC)}, pages 31--38.

\bibitem[Coelho et~al., 2011]{coelho2011multi}
Coelho, A. L.~V., Fernandes, E., and Faceli, K. (2011).
\newblock Multi-objective design of hierarchical consensus functions for
  clustering ensembles via genetic programming.
\newblock {\em Decision Support Systems}, 51(4):794--809.

\bibitem[Dy and Brodley, 2004]{dy2004feature}
Dy, J.~G. and Brodley, C.~E. (2004).
\newblock Feature selection for unsupervised learning.
\newblock {\em Journal of Machine Learning Research}, 5:845--889.

\bibitem[Eiben and Smith, 2015]{eiben2015introduction}
Eiben, A.~E. and Smith, J.~E. (2015).
\newblock {\em Introduction to Evolutionary Computing}.
\newblock Natural Computing Series. Springer.

\bibitem[Espejo et~al., 2010]{espejo2010survey}
Espejo, P.~G., Ventura, S., and Herrera, F. (2010).
\newblock A survey on the application of genetic programming to classification.
\newblock {\em {IEEE} Trans. Systems, Man, and Cybernetics, Part {C}},
  40(2):121--144.

\bibitem[Ester et~al., 1996]{ester1996density}
Ester, M., Kriegel, H., Sander, J., and Xu, X. (1996).
\newblock A density-based algorithm for discovering clusters in large spatial
  databases with noise.
\newblock In {\em Proceedings of the Second International Conference on
  Knowledge Discovery and Data Mining (KDD-96), Portland, Oregon, {USA}}, pages
  226--231.

\bibitem[Falco et~al., 2005]{falco2005novel}
Falco, I.~D., Tarantino, E., Cioppa, A.~D., and Gagliardi, F. (2005).
\newblock A novel grammar-based genetic programming approach to clustering.
\newblock In {\em Proceedings of the {ACM} Symposium on Applied Computing
  (SAC)}, pages 928--932.

\bibitem[Fayyad et~al., 1996]{fayyad1996data}
Fayyad, U., Piatetsky-Shapiro, G., and Smyth, P. (1996).
\newblock From data mining to knowledge discovery in databases.
\newblock {\em AI Magazine}, 17(3):37--54.

\bibitem[Garc{\'{\i}}a and G{\'{o}}mez{-}Flores, 2016]{garcia2016automatic}
Garc{\'{\i}}a, A.~J. and G{\'{o}}mez{-}Flores, W. (2016).
\newblock Automatic clustering using nature-inspired metaheuristics: {A}
  survey.
\newblock {\em Appl. Soft Comput.}, 41:192--213.

\bibitem[Garc{\'{\i}}a{-}Pedrajas et~al., 2014]{pedrajas2014scalable}
Garc{\'{\i}}a{-}Pedrajas, N., de~Haro{-}Garc{\'{\i}}a, A., and
  P{\'{e}}rez{-}Rodr{\'{\i}}guez, J. (2014).
\newblock A scalable memetic algorithm for simultaneous instance and feature
  selection.
\newblock {\em Evolutionary Computation}, 22(1):1--45.

\bibitem[Handl and Knowles, 2007]{handl2007evolutionary}
Handl, J. and Knowles, J.~D. (2007).
\newblock An evolutionary approach to multiobjective clustering.
\newblock {\em {IEEE} Trans. Evolutionary Computation}, 11(1):56--76.

\bibitem[Hartuv and Shamir, 2000]{shamir2000clustering}
Hartuv, E. and Shamir, R. (2000).
\newblock A clustering algorithm based on graph connectivity.
\newblock {\em Inf. Process. Lett.}, 76(4-6):175--181.

\bibitem[Haynes and Sen, 1997]{haynes1997crossover}
Haynes, T. and Sen, S. (1997).
\newblock Crossover operators for evolving a team.
\newblock In Koza, J.~R., Deb, K., Dorigo, M., Fogel, D.~B., Garzon, M., Iba,
  H., and Riolo, R.~L., editors, {\em Genetic Programming 1997: Proceedings of
  the Second Annual Conference}, pages 162--167, CA, USA. Morgan Kaufmann.

\bibitem[J.~A.~Hartigan, 1979]{hartigan1979}
J.~A.~Hartigan, M. A.~W. (1979).
\newblock Algorithm {AS} 136: A k-means clustering algorithm.
\newblock {\em Journal of the Royal Statistical Society. Series C (Applied
  Statistics)}, 28(1):100--108.

\bibitem[Jain, 2010]{jain2010data}
Jain, A.~K. (2010).
\newblock Data clustering: 50 years beyond k-means.
\newblock {\em Pattern Recognition Letters}, 31(8):651--666.

\bibitem[Jolliffe, 2011]{jolliffe2011pca}
Jolliffe, I.~T. (2011).
\newblock Principal component analysis.
\newblock In {\em International Encyclopedia of Statistical Science}, pages
  1094--1096. Springer.

\bibitem[Koza, 1992]{koza1992genetic}
Koza, J.~R. (1992).
\newblock {\em Genetic programming: on the programming of computers by means of
  natural selection}, volume~1.
\newblock MIT press.

\bibitem[Kuo et~al., 2012]{kuo2012integration}
Kuo, R.~J., Syu, Y.~J., Chen, Z., and Tien, F. (2012).
\newblock Integration of particle swarm optimization and genetic algorithm for
  dynamic clustering.
\newblock {\em Inf. Sci.}, 195:124--140.

\bibitem[Lensen et~al., 2017a]{lensen2017gpgc}
Lensen, A., Xue, B., and Zhang, M. (2017a).
\newblock {GPGC:} genetic programming for automatic clustering using a flexible
  non-hyper-spherical graph-based approach.
\newblock In {\em Proceedings of the Genetic and Evolutionary Computation
  Conference, {GECCO}.}, pages 449--456. {ACM}.

\bibitem[Lensen et~al., 2017b]{lensen2017using}
Lensen, A., Xue, B., and Zhang, M. (2017b).
\newblock Using particle swarm optimisation and the silhouette metric to
  estimate the number of clusters, select features, and perform clustering.
\newblock In {\em Proceedings of the 20th European Conference on the
  Applications of Evolutionary Computation (EvoApplications), Part {I}}, volume
  10199 of {\em Lecture Notes in Computer Science}, pages 538--554. Springer.

\bibitem[Liu and Motoda, 2012]{liu2012feature}
Liu, H. and Motoda, H. (2012).
\newblock {\em Feature selection for knowledge discovery and data mining},
  volume 454.
\newblock Springer Science \& Business Media.

\bibitem[Liu and Yu, 2005]{liu2005toward}
Liu, H. and Yu, L. (2005).
\newblock Toward integrating feature selection algorithms for classification
  and clustering.
\newblock {\em {IEEE} Trans. Knowl. Data Eng.}, 17(4):491--502.

\bibitem[Lorena and Furtado, 2001]{lorena2001constructive}
Lorena, L. A.~N. and Furtado, J.~C. (2001).
\newblock Constructive genetic algorithm for clustering problems.
\newblock {\em Evolutionary Computation}, 9(3):309--328.

\bibitem[Men{\'{e}}ndez et~al., 2014]{menendez2014genetic}
Men{\'{e}}ndez, H.~D., Barrero, D.~F., and Camacho, D. (2014).
\newblock A genetic graph-based approach for partitional clustering.
\newblock {\em International Journal of Neural Systems}, 24(3).

\bibitem[M{\"{u}}ller et~al., 2009]{muller2009evaluating}
M{\"{u}}ller, E., G{\"{u}}nnemann, S., Assent, I., and Seidl, T. (2009).
\newblock Evaluating clustering in subspace projections of high dimensional
  data.
\newblock In {\em Proceedings of the 35th International Conference on Very
  Large Data Bases ({VLDB})}, pages 1270--1281.

\bibitem[Nanda and Panda, 2014]{nanda2014survey}
Nanda, S.~J. and Panda, G. (2014).
\newblock A survey on nature inspired metaheuristic algorithms for partitional
  clustering.
\newblock {\em Swarm and Evolutionary Computation}, 16:1--18.

\bibitem[Naredo and Trujillo, 2013]{naredo2013searching}
Naredo, E. and Trujillo, L. (2013).
\newblock Searching for novel clustering programs.
\newblock In {\em Genetic and Evolutionary Computation Conference, {GECCO} '13,
  Amsterdam, The Netherlands, July 6-10, 2013}, pages 1093--1100.

\bibitem[Neshatian et~al., 2012]{neshatian2012filter}
Neshatian, K., Zhang, M., and Andreae, P. (2012).
\newblock A filter approach to multiple feature construction for symbolic
  learning classifiers using genetic programming.
\newblock {\em {IEEE} Trans. Evolutionary Computation}, 16(5):645--661.

\bibitem[Parsons et~al., 2004]{parsons2004subspace}
Parsons, L., Haque, E., and Liu, H. (2004).
\newblock Subspace clustering for high dimensional data: a review.
\newblock {\em {SIGKDD} Explorations}, 6(1):90--105.

\bibitem[Peignier et~al., 2015]{peignier2015subspace}
Peignier, S., Rigotti, C., and Beslon, G. (2015).
\newblock Subspace clustering using evolvable genome structure.
\newblock In {\em Proceedings of the Genetic and Evolutionary Computation
  Conference, {GECCO}}, pages 575--582.

\bibitem[Picarougne et~al., 2007]{picarougne2007new}
Picarougne, F., Azzag, H., Venturini, G., and Guinot, C. (2007).
\newblock A new approach of data clustering using a flock of agents.
\newblock {\em Evolutionary Computation}, 15(3):345--367.

\bibitem[Poli et~al., 2008]{poli2008field}
Poli, R., Langdon, W.~B., and McPhee, N.~F. (2008).
\newblock {\em A Field Guide to Genetic Programming}.
\newblock lulu.com.

\bibitem[Sheng et~al., 2016]{sheng2016adaptive}
Sheng, W., Chen, S., Sheng, M., Xiao, G., Mao, J., and Zheng, Y. (2016).
\newblock Adaptive multisubpopulation competition and multiniche crowding-based
  memetic algorithm for automatic data clustering.
\newblock {\em {IEEE} Trans. Evolutionary Computation}, 20(6):838--858.

\bibitem[Sheng et~al., 2008]{sheng2008niching}
Sheng, W., Liu, X., and Fairhurst, M.~C. (2008).
\newblock A niching memetic algorithm for simultaneous clustering and feature
  selection.
\newblock {\em {IEEE} Trans. Knowl. Data Eng.}, 20(7):868--879.

\bibitem[Tang et~al., 2014]{tang2014feature}
Tang, J., Alelyani, S., and Liu, H. (2014).
\newblock Feature selection for classification: {A} review.
\newblock In {\em Data Classification: Algorithms and Applications}, pages
  37--64. {CRC} Press.

\bibitem[Thomason and Soule, 2007]{thomason2007novel}
Thomason, R. and Soule, T. (2007).
\newblock Novel ways of improving cooperation and performance in ensemble
  classifiers.
\newblock In {\em Proceedings of the Genetic and Evolutionary Computation
  Conference, ({GECCO})}, pages 1708--1715.

\bibitem[Vahdat and Heywood, 2014]{vahdat2014on}
Vahdat, A. and Heywood, M.~I. (2014).
\newblock On evolutionary subspace clustering with symbiosis.
\newblock {\em Evolutionary Intelligence}, 6(4):229--256.

\bibitem[van~der Maaten and Hinton, 2008]{maatenTSNE}
van~der Maaten, L. and Hinton, G.~E. (2008).
\newblock Visualizing high-dimensional data using t-{SNE}.
\newblock {\em Journal of Machine Learning Research}, 9:2579--2605.

\bibitem[Van~Dongen, 2000]{van2001graph}
Van~Dongen, S.~M. (2000).
\newblock {\em Graph clustering by flow simulation}.
\newblock PhD thesis, University of Utrecht.

\bibitem[Vinh et~al., 2010]{nguyen2010information}
Vinh, N.~X., Epps, J., and Bailey, J. (2010).
\newblock Information theoretic measures for clusterings comparison: Variants,
  properties, normalization and correction for chance.
\newblock {\em Journal of Machine Learning Research}, 11:2837--2854.

\bibitem[von Luxburg, 2007]{luxburg2007spectral}
von Luxburg, U. (2007).
\newblock A tutorial on spectral clustering.
\newblock {\em Statistics and Computing}, 17(4):395--416.

\bibitem[von Luxburg et~al., 2012]{luxburg2011clustering}
von Luxburg, U., Williamson, R.~C., and Guyon, I. (2012).
\newblock Clustering: Science or art?
\newblock In {\em Proceedings of the Unsupervised and Transfer Learning
  Workshop held at {ICML} 2011}, pages 65--80.

\bibitem[Xu and II, 2005]{xu2005survey}
Xu, R. and II, D. C.~W. (2005).
\newblock Survey of clustering algorithms.
\newblock {\em {IEEE} Trans. Neural Networks}, 16(3):645--678.

\bibitem[Xue et~al., 2016]{xue2015survey}
Xue, B., Zhang, M., Browne, W.~N., and Yao, X. (2016).
\newblock A survey on evolutionary computation approaches to feature selection.
\newblock {\em {IEEE} Trans. Evolutionary Computation}, 20(4):606--626.

\end{thebibliography}
\pagebreak
\begin{appendices}
	\vspace{-1em}
	\normalsize
	\section{Subspace Clustering Results}
		\begin{figure*}[t]
		\centering
		\null\hfill
		\subfloat[Varying the number of dimensions]{\includegraphics[width=.49\textwidth]{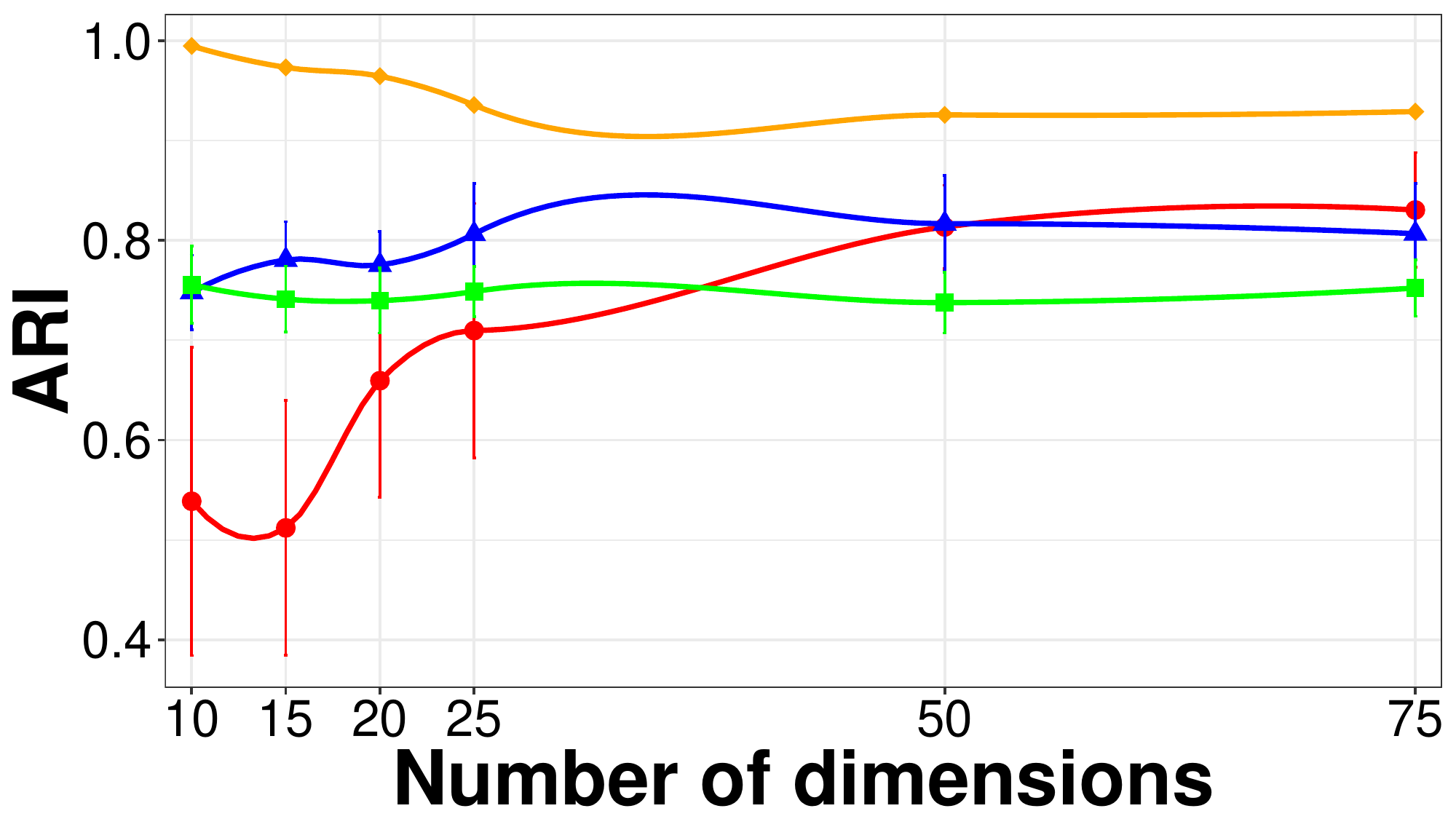}}
		\hfill
		\subfloat[Varying the number of instances]{\includegraphics[width=.49\textwidth]{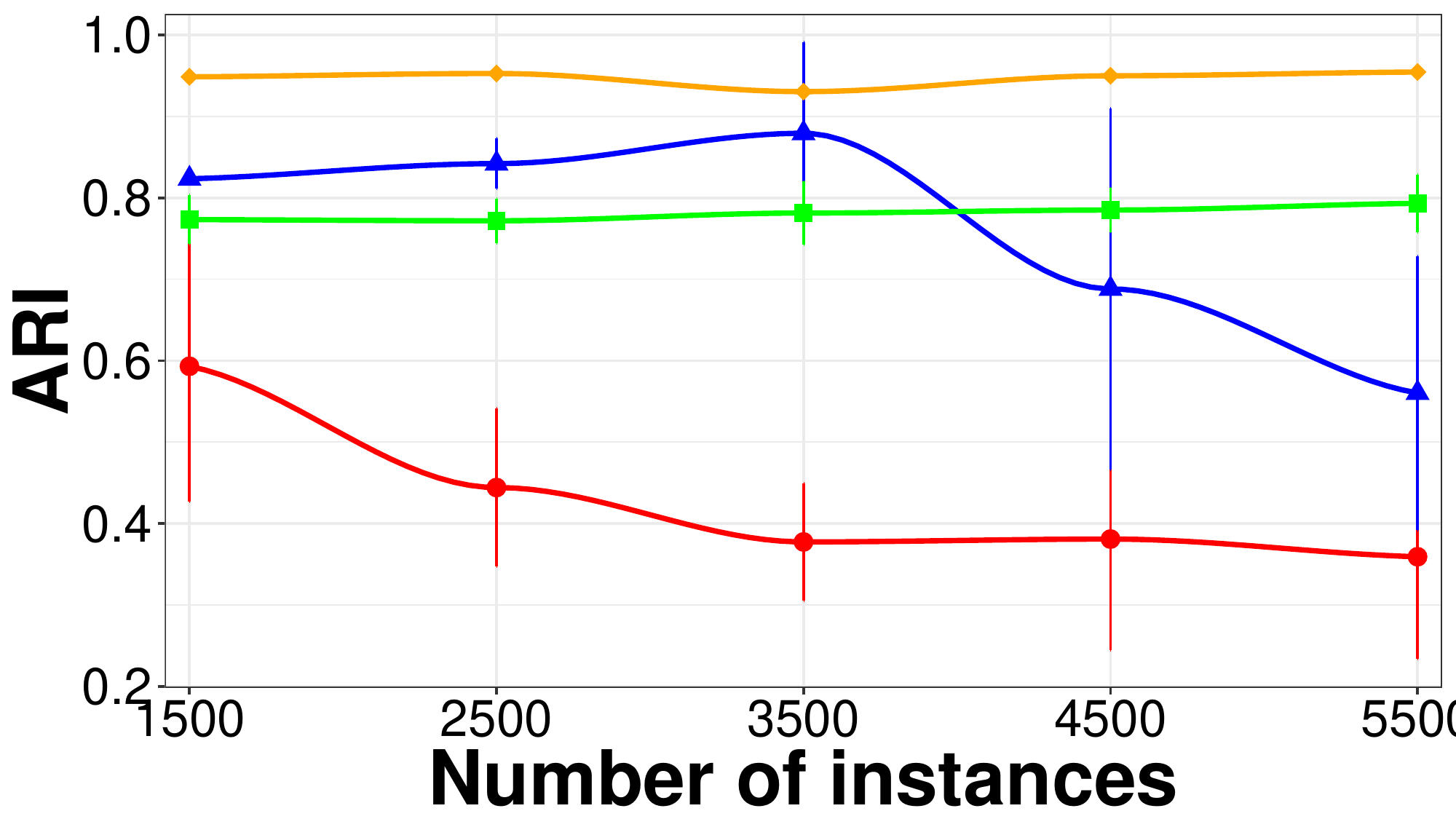}} 
		\hfill\null
		\caption{\revisionOne{ARI achieved by each clustering method on the OpenSubspace clustering datasets, where either the number of dimensions, or number of instances are varied.The red dots correspond to GPGC, the blue triangles to GPGC-AIC, the green squares to PROCLUS, and the orange diamonds to DOC respectively.}}
		\label{fig:subspace}
		\vspace{-1em}
	\end{figure*}
	\revisionOnePar{The fitness function proposed in this paper is designed to use all features in the feature space when calculating distances, as our primary goal is to reduce dimensionality and produce interpretable similarity functions, while maintaining good cluster quality. However, a natural extension of our proposed approach is to apply it to subspace clustering problems, as our GP representation has the potential to use different feature subsets in different clusters. To investigate the plausibility of this extension, we applied GPGC and GPGC-AIC to datasets from OpenSubspace \citep{muller2009evaluating}, a collection of popular subspace benchmarking datasets. We chose to use the PROCLUS and DOC algorithms for comparison as examples of commonly used cell-based and clustering-oriented subspace clustering algorithms respectively. We chose PROCLUS and DOC as they have been shown to have superior (or similar) performance to other subspace algorithms in their paradigm \citep{muller2009evaluating}. We do not compare to a clustering algorithm from the third paradigm -- density-based subspace clustering -- as these methods all produced overlapping (i.e.\ non-crisp) clusters, which is not the focus of this study.
	
	Fig.\ \ref{fig:subspace} shows the performance of each of the four methods (GPGC, GPGC-AIC, PROCLUS, and DOC) as the number of dimensions is varied (Fig.\ \ref{fig:subspace}a) and the number of instances (Fig.\ \ref{fig:subspace}b) is varied respectively. For each dataset, we plot the ARI achieved by each method, as well as the standard deviation across 30 runs (the vertical bars). The two proposed methods are competitive with PROCLUS as the dimensionality is increased, but PROCLUS clearly outperforms GPGC and is slightly better than GPGC-AIC as the number of instances is increased. The DOC method is clearly the best of all the methods on both categories of datasets. However, GPGC-AIC in particular shows some promise given it has not been optimised for this task, but can still achieve competitive results often with one common subspace clustering method. Furthermore, GPGC-AIC is clearly superior to vanilla GPGC, which further reinforces the findings throughout this paper. We hope to further improve GPGC-AIC and make it a competitive subspace clustering algorithm in the future by developing a new fitness function and designing new genetic operators.
}
\end{appendices}

\end{document}